\documentclass[pdflatex,sn-mathphys-num]{sn-jnl}

\usepackage{gensymb}            
\usepackage{tabularx}           

\usepackage{graphicx}%
\usepackage{multirow}%
\usepackage{amsmath,amssymb,amsfonts}%
\usepackage{amsthm}%
\usepackage{mathrsfs}%
\usepackage[title]{appendix}%
\usepackage{xcolor}%
\usepackage{textcomp}%
\usepackage{manyfoot}%
\usepackage{booktabs}%
\usepackage{algorithm}%
\usepackage{algorithmicx}%
\usepackage{algpseudocode}%
\usepackage{listings}%

\newcommand{\ba}{\textbf{\emph{a}}}
\newcommand{\be}{\textbf{\emph{e}}}
\newcommand{\bm}{\textbf{\emph{m}}}
\newcommand{\bp}{\textbf{\emph{p}}}
\newcommand{\bq}{\textbf{\emph{q}}}
\newcommand{\br}{\textbf{\emph{r}}}
\newcommand{\bu}{\textbf{\emph{u}}}
\newcommand{\bv}{\textbf{\emph{v}}}
\newcommand{\bx}{\textbf{\emph{x}}}
\newcommand{\by}{\textbf{\emph{y}}}
\newcommand{\bI}{\textbf{\emph{I}}}
\newcommand{\bK}{\textbf{\emph{K}}}
\newcommand{\bP}{\textbf{\emph{P}}}
\newcommand{\bR}{\textbf{\emph{R}}}
\newcommand{\bS}{\textbf{\emph{S}}}
\newcommand{\bT}{\textbf{\emph{T}}}
\newcommand{\PP}{\mathbb{P}}
\newcommand{\bbeta}{\boldsymbol{\beta}}
\newcommand{\bell}{\boldsymbol{\ell}}
\newcommand{\bPhi}{\boldsymbol{\Phi}}

\newcolumntype{Y}{>{\centering\arraybackslash}X}
\newcolumntype{P}[1]{>{\centering\arraybackslash}p{#1}}

\theoremstyle{thmstyleone}%
\theoremstyle{thmstyletwo}%

\theoremstyle{thmstylethree}%

\begin{document}

\title[Lunar Flashlight OPNAV]{LONEStar: The Lunar Flashlight Optical Navigation Experiment }


\author[1]{\fnm{Michael} \sur{Krause}}
\equalcont{These authors contributed equally to this work.}

\author[1]{\fnm{Ava} \sur{Thrasher}}

\author[1]{\fnm{Priyal} \sur{Soni}}

\author[1]{\fnm{Liam} \sur{Smego}}

\author[1]{\fnm{Reuben} \sur{Isaac}}

\author[1]{\fnm{Jennifer} \sur{Nolan}}

\author[1]{\fnm{Micah} \sur{Pledger}}

\author[1]{\fnm{E. Glenn} \sur{Lightsey}}

\author[2]{\fnm{W. Jud} \sur{Ready}}

\author*[1]{\fnm{John} \sur{Christian}}\email{john.a.christian@gatech.edu}
\equalcont{These authors contributed equally to this work.}

\affil[1]{\orgdiv{Guggenheim School of Aerospace Engineering}, \orgname{Georgia Institute of Technology}, \orgaddress{\city{Atlanta}, \postcode{30332}, \state{GA}, \country{USA}}}

\affil[2]{ \orgname{Georgia Tech Research Institute}, \orgaddress{\city{Atlanta}, \postcode{30318}, \state{GA}, \country{USA}}}


\abstract{This paper documents the results from the highly successful Lunar flashlight Optical Navigation Experiment with a Star tracker (LONEStar). Launched in December 2022, Lunar Flashlight (LF) was a NASA-funded technology demonstration mission. After a propulsion system anomaly prevented capture in lunar orbit, LF was ejected from the Earth-Moon system and into heliocentric space. NASA subsequently transferred ownership of LF to Georgia Tech to conduct an unfunded extended mission to demonstrate further advanced technology objectives, including LONEStar. From August-December 2023, the LONEStar team performed on-orbit calibration of the optical instrument and a number of different OPNAV experiments. This campaign included the processing of nearly 400 images of star fields, Earth and Moon, and four other planets (Mercury, Mars, Jupiter, and Saturn). LONEStar provided the first on-orbit demonstrations of heliocentric navigation using only optical observations of planets. Of special note is the successful in-flight demonstration of (1) instantaneous triangulation with simultaneous sightings of two planets with the LOST algorithm and (2) dynamic triangulation with sequential sightings of multiple planets. }




\maketitle

\section{Introduction}\label{Sec:Intro}
Lunar Flashlight (LF) was a NASA technology demonstration mission \cite{Cheek:2022} launched in December 2022 that had a secondary science goal of searching for water ice on the Moon \cite{Cohen:2020}. LF successfully completed all of its primary objectives as a technology demonstration mission. However, following a propulsion system anomaly, the spacecraft was unable to reach the originally planned lunar orbit and the secondary (non-mandatory) science goal was lost. The vehicle instead executed an Earth flyby in May 2023 before being ejected from the Earth-Moon system and into its present heliocentric orbit. Fortunately, despite some of the challenges encountered with the innovative propulsion system, the vehicle was otherwise fully operational. Consequently, as part of an extended mission through the Georgia Institute of Technology (GT), the Lunar Flashlight team identified optical navigation as a compelling technology demonstration experiment that could be completed along the vehicle’s Earth-Moon departure trajectory. This investigation was designated the LF Optical Navigation Experiment with a Star tracker (LONEStar) and became a primary objective of the GT-led extended mission of the Lunar Flashlight spacecraft. 

The art of optical navigation (OPNAV) has a colorful history \cite{Owen:2008,Butrica:2014} and has played a pivotal role in the success of many planetary exploration missions---including Voyager \cite{Synnott:1986,Riedel:1990}, Cassini \cite{Gillam:2007,Hollenberg:2019}, New Horizons \cite{Williams:2015}, Artemis I \cite{Inman:2024}, and many others. With the exception of Deep Space 1 (DS1) \cite{Bhaskaran:1998,Bhaskaran:2000}, however, few missions have explored the efficacy of autonomous OPNAV in a heliocentric orbit. Similar ideas were considered for ESA’s Smart-1 mission, but never actually demonstrated in flight \cite{Polle:2006}. At the time this paper was written, DS1 and Lunar Flashlight are the only two missions to have successfully demonstrated OPNAV-only orbit determination in heliocentric space.

The DS1 AutoNav and LONEStar experiments both navigated using line-of-sight (LOS) observations to unresolved celestial bodies. However, the results of these two projects differ in three important ways. First, DS1 AutoNav processed images and performed state estimation entirely onboard the spacecraft (the first of its kind), whereas LONEStar was conducted by human analysts on Earth. Second, the DS1 team chose to collect images of nearby asteroids (instead of distant planets) against star field backgrounds to achieve better absolute navigation performance. For LONEStar, however, no nearby asteroids were visible during the investigation and the LF team instead focused on collecting images of planets. As a consequence, LONEStar became the first flight demonstration of an OPNAV-only orbit solution using exclusively planet observations in heliocentric space. Third, DS1 AutoNav directly processed LOS measurements in a navigation filter. While this was also done for LONEStar (see Section~\ref{Sec:BatchFilter}), many of the LONEStar experiments focused on explicit triangulation for instantaneous localization or initial orbit determination. Thus, LONEStar is the first demonstration of heliocentric spacecraft localization by (1) absolute triangulation using the simultaneous observation of multiple planets (Mercury and Mars, see Section~\ref{Sec:InstantTriMercuryMars}), (2) absolute triangulation with sequential images of two planets (Jupiter and Saturn, see Section~\ref{Sec:SequentialTriJupSat}), and (3) dynamic triangulation using sequential observations of multiple planets over a long period of time (Jupiter and Saturn, see Section~\ref{Sec:DynamicTri}). 

Although LONEStar is the first flight demonstration of planet-based celestial triangulation, this fundamental idea has been proposed since at least the 1950s \cite{Atkinson:1950,Stuhlinger:1957,Bock:1959,Bock:1961}. After a period of stagnation, the idea of OPNAV by explicit triangulation gained significant traction in the last decade \cite{Karimi:2015,Broschart:2019,Franzese:2022,Andreis:2022,Henry:2023a,Henry:2023b}. LONEStar builds on this legacy of innovation, while also providing an important experimental verification of the Linear Optimal Sine Triangulation (LOST) method recently theorized by Henry and Christian \cite{Henry:2023a,Henry:2023b}. 

\section{The Lunar Flashlight Mission}

\begin{figure}[b!]
    \centering
    \includegraphics[width=1\linewidth]{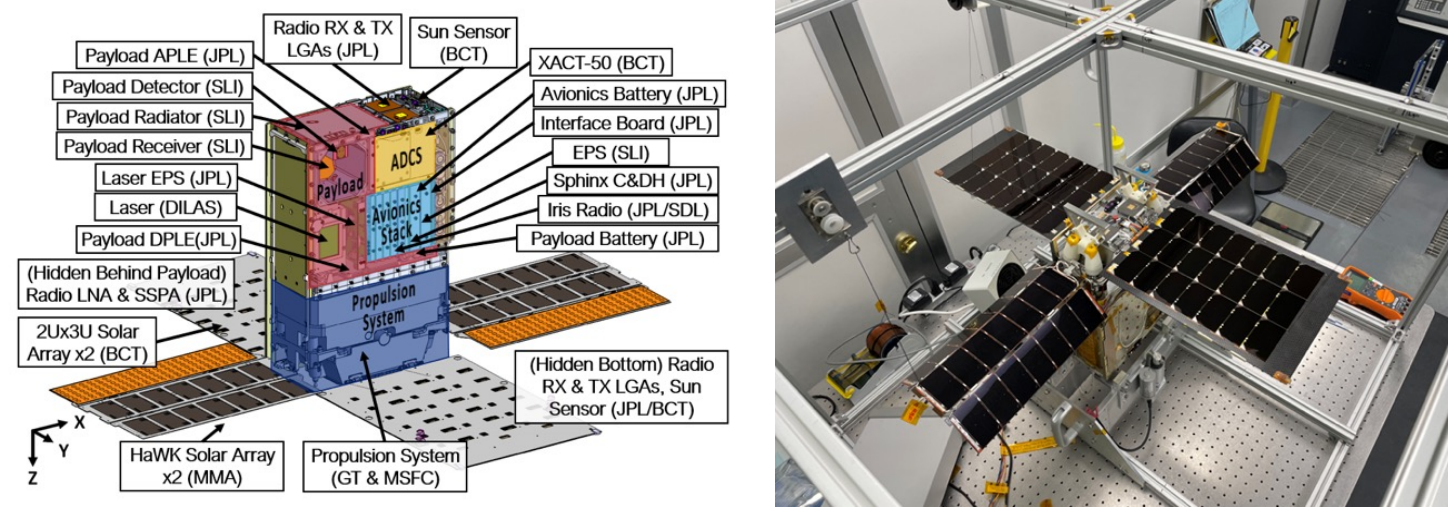}
    \caption{The Lunar Flashlight spacecraft. Shown here are the major spacecraft subsystems and components \cite{Smith:2023} (left) and the spacecraft during integration and testing at Georgia Tech (right).}
    \label{fig:LFoverview}
\end{figure}

Lunar Flashlight was conceived as a small satellite technology demonstration mission developed by the Jet Propulsion Laboratory (JPL) and funded by NASA’s Space Technology Mission Directorate (STMD) \cite{Cohen:2020,Paige:2016}. Lunar Flashlight’s primary mission objectives were to demonstrate several advanced small satellite technologies on a cislunar mission. Innovative components of the 6U CubeSat (see Fig.~\ref{fig:LFoverview}) included: (1) a radiation tolerant microprocessor, (2) a Deep Space Network (DSN) compatible transceiver, (3) a relatively high-power infrared laser, and (4) an experimental monopropellant propulsion system. The mission had an additional science goal to survey the lunar south pole region for evidence of surface ice in permanently shadowed regions using the onboard infrared laser. This latter science goal required orbit insertion into a highly elliptic lunar orbit.

The Lunar Flashlight team was a partnership among several different institutions. JPL managed the project and developed several of the key subsystems, including the radiation tolerant microprocessor, radio transceiver, and laser instrument. NASA’s Marshall Space Flight Center and Georgia Tech designed and built the experimental propulsion system \cite{Huggins:2021,Cheek:2021}. Additionally, spacecraft integration, testing, and mission operations were conducted at Georgia Tech \cite{Cheek:2022}.  

\begin{figure}[t!]
    \centering
    \includegraphics[width=1\linewidth]{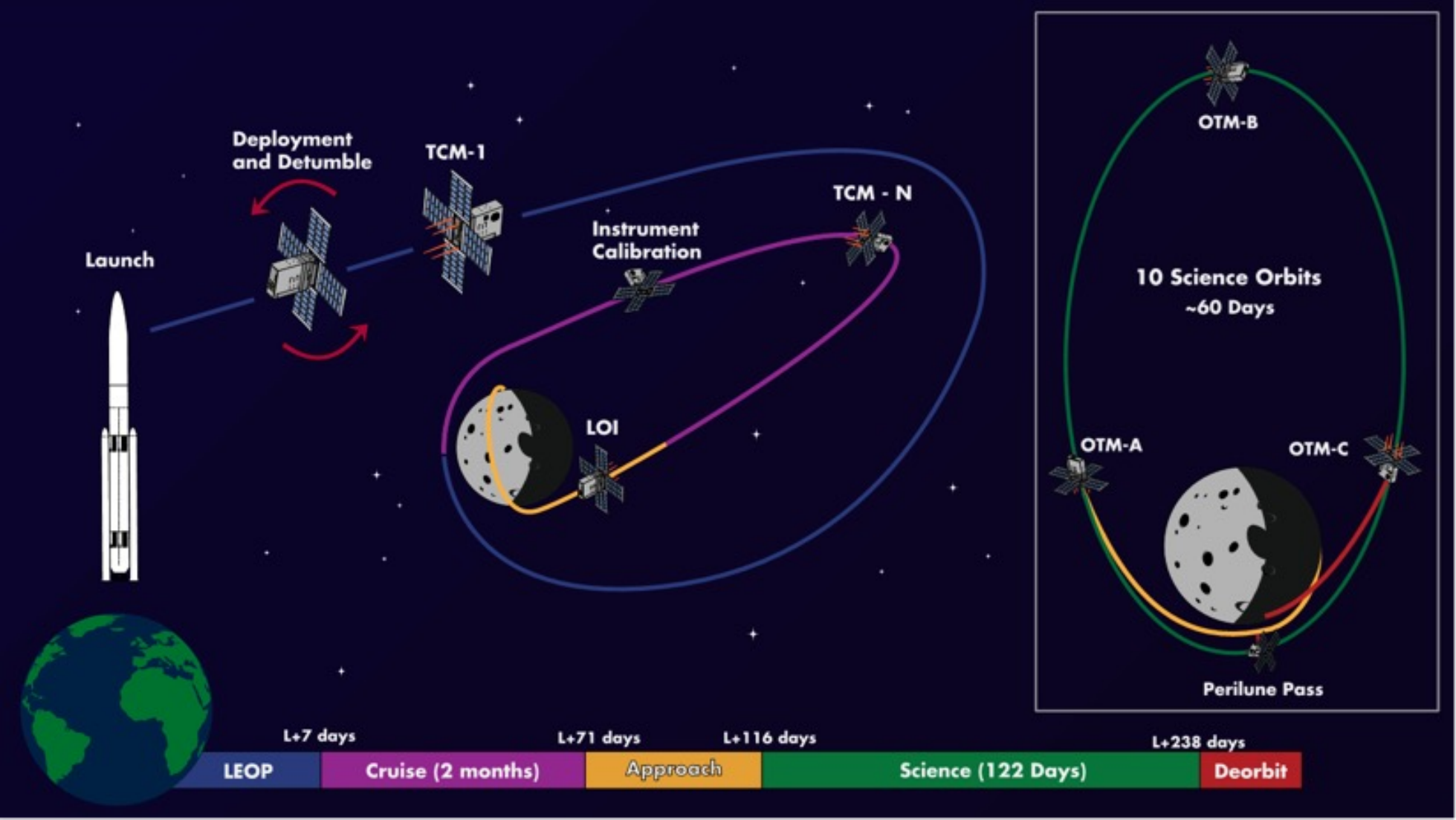}
    \caption{Original Lunar Flashlight mission concept of operations \cite{Hauge:2023}.}
    \label{fig:LFCONOPS}
\end{figure}

\begin{figure}[t!]
    \centering
    \includegraphics[width=1\linewidth]{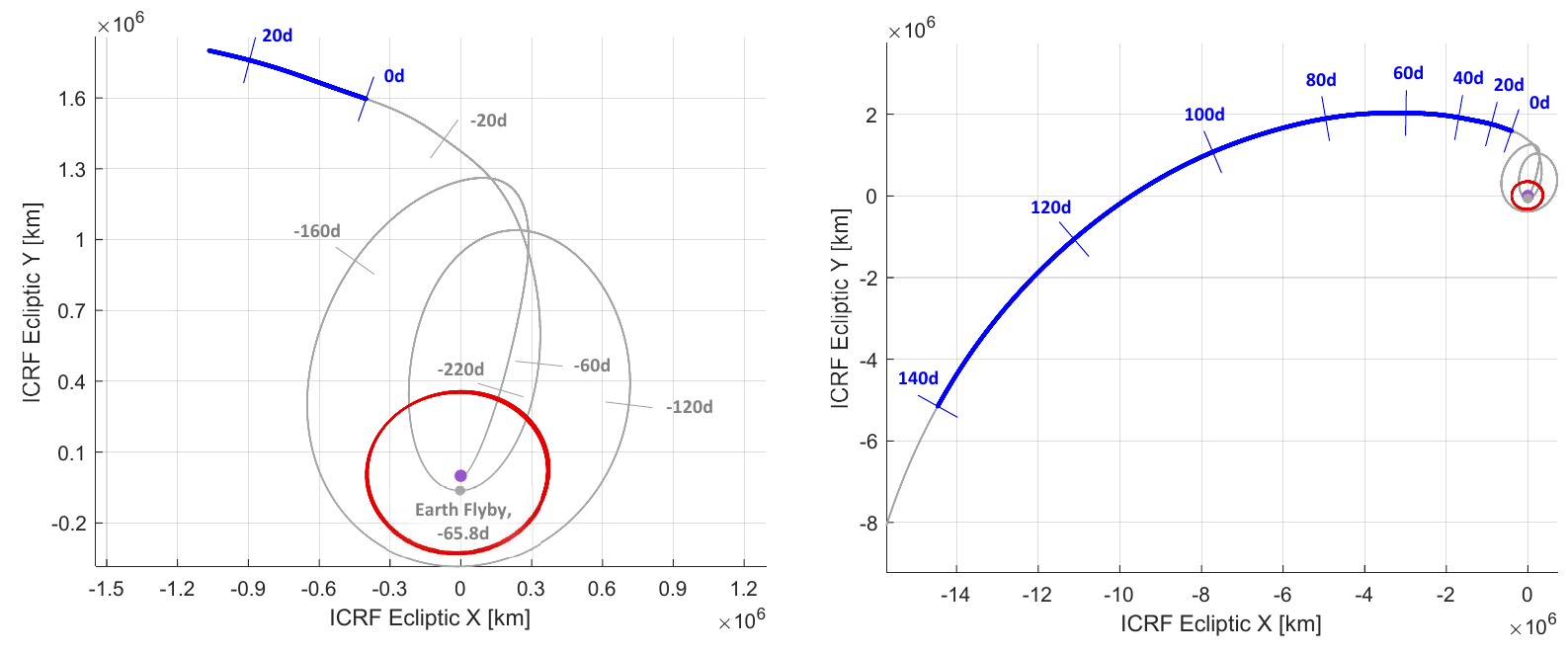}
    \caption{Visualization of as-flown Lunar Flashlight trajectory as seen in Earth-centered ecliptic ICRF. After about 5 months within the Earth-Moon system (left), LF was ejected into a heliocentric trajectory (right). Shown in dark blue is the portion of the trajectory corresponding the LONEStar imaging campaign.  Tick marks indicate the LONEStar elapsed time relative to the reference epoch 2023-JUL-22 00:00:00 UTC. The Moon’s orbit is shown in red. }
    \label{fig:LFtrajectorySmall}
\end{figure}

Lunar Flashlight was launched on 2022-DEC-11 and deployed into a trajectory which would take it to cislunar space. The original concept of operations (see Fig.~\ref{fig:LFCONOPS}) required multiple trajectory correction maneuvers (TCMs) to insert the spacecraft into a target near rectilinear halo orbit (NRHO) around the moon. However, an anomaly with the propulsion system prevented the spacecraft from achieving the target NRHO. After several attempts to recover the propulsion system, the official mission was declared over in May 2023 after successfully completing all of its technology demonstration objectives. After an Earth flyby, the spacecraft’s trajectory exited the Earth-Moon system and entered into heliocentric orbit (see Fig.~\ref{fig:LFtrajectorySmall}). The current LF orbit has a 15-year synodic period with Earth and the spacecraft will return to the Earth-Moon vicinity in 2038. Other than the propulsion system, the spacecraft was functioning nominally, and NASA officially transferred the spacecraft to Georgia Tech for additional experimentation in August 2023 \cite{LPIB:2023}.

The GT-led LF extended mission began in mid-2023 and lasted until December 2023. Since an abundance of activities had been conducted using the propulsion unit during the primary mission, the on-orbit demonstration of a number of novel OPNAV techniques was chosen as the primary extended mission objective.  Results from other mission activities are available elsewhere \cite{Smith:2023}. 

\begin{figure}[t!]
    \centering
    \includegraphics[width=0.7\linewidth]{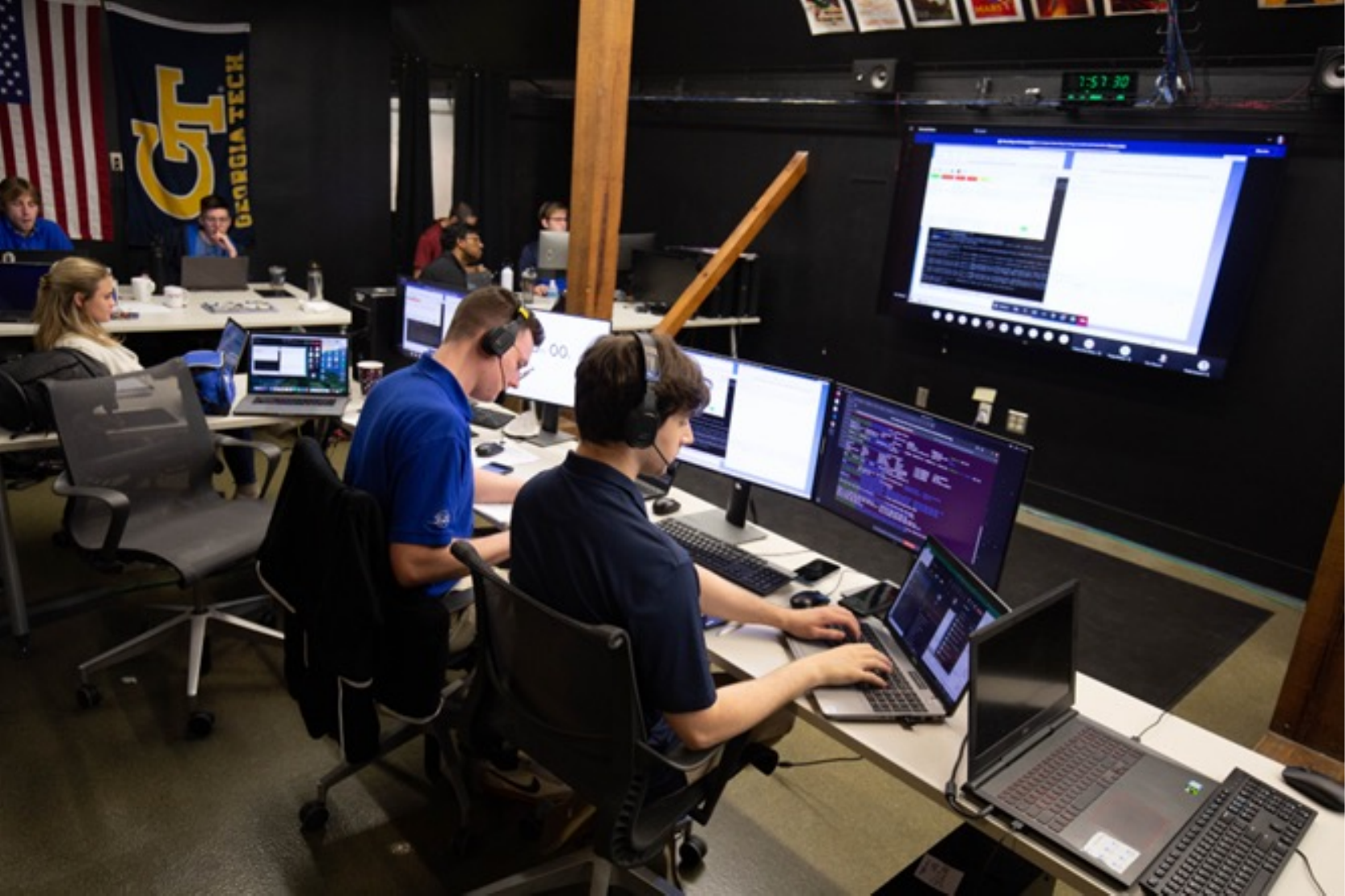}
    \caption{Georgia Tech Mission Operations Center (MOC) during Lunar Flashlight mission.}
    \label{fig:LFoperations}
\end{figure}

LF was controlled directly by GT students from a Mission Operations Center (MOC) located on the GT campus in Atlanta, Georgia (Fig.~\ref{fig:LFoperations}) \cite{Starr:2024}. To the authors’ knowledge, Lunar Flashlight is the first interplanetary spacecraft to be operated by an operations team comprised entirely of students. During the course of the LF mission, GT conducted more than 500 Deep Space Network (DSN) contacts with the spacecraft, consisting of both ``autonomous" and ``operator in the loop'' interactions. 

Lunar Flashlight operations relied on Earth-based radiometric navigation with DSN. The JPL-led Lunar Flashlight Navigation Team processed two-way Doppler, two-way ranging, and delta-differential one-way ranging (DDOR) navigation observables from DSN between launch (2022-DEC-11) and about 2023-JUN-27 14:00:00 UTC. However, the portion of the trajectory covering the times of the LONEStar experiment are primarily based on DSN observations between 2023-MAY-24 and 2023-JUN-27. The final DSN-based navigation solution was delivered in late June 2023 \cite{Cox:2023} and this product was used as the reference trajectory for LONEStar. Since LF did not attempt to use any onboard thrusters after 2023-MAY-05 (19 days before the start of the final DSN tracking data arc), the final DSN-based trajectory remained unperturbed by propulsive maneuvers and was found to be valid for the entirety of the LONEStar investigation. 

\section{OPNAV Instrument}

\subsection{Overview of Instrument}
\begin{figure}[b!]
    \centering
    \includegraphics[width=1\linewidth]{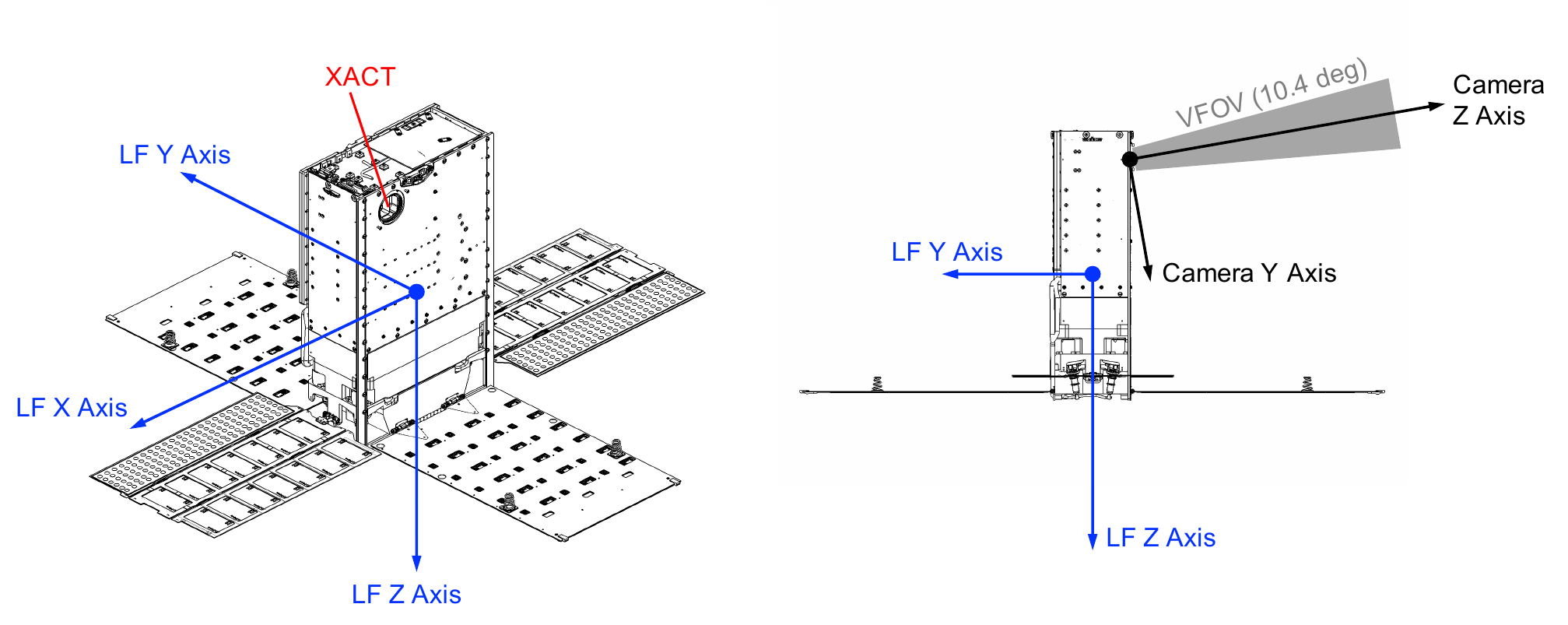}
    \caption{Illustration of Lunar Flashlight spacecraft configuration, important coordinate frames, and orientation of camera field of view (FOV).}
    \label{fig:LFFramesAndFOV}
\end{figure}

The Blue Canyon Technologies XACT-50 is an essential component of the LF guidance, navigation, and control (GNC) system \cite{Lai:2020,Sternberg:2022}.
While intended for attitude determination, the star tracker within the XACT-50 module is also capable of acting as a camera. This instrument has a $1024 \times 1280$ pixel image sensor with overall specifications as shown in Table~\ref{tab:CameraSpecs}. The camera is mounted on the –Y face of the LF bus, with the boresight rotated by 10 degrees from the face normal about the LF +X axis (i.e., canted away from the solar panels), as shown in Fig.~\ref{fig:LFFramesAndFOV}.

\begin{table}[h!t]
    \centering
    \caption{Summary of camera specifications. The FOV and IFOV numbers reported here are based on the in-flight calibration discussed in Section~\ref{Sec:GeomCal}.}
    \begin{tabular}{lcc}
    \hline
    Parameter & Value\\
    \hline
    Field of View (FOV) & $10.4 \times 13.0$ deg \\
    Instantaneous FOV (IFOV) & 36.6 arcsec \\
    Image Sensor Resolution & $1024 \times 1280$ pixel \\
    Sun Keep Out Zone (KOZ) & 45 deg \\
    Pixel bit depth & 10 bits \\
    \hline
    \end{tabular}
    \label{tab:CameraSpecs}
\end{table}

\begin{figure}[h!]
    \centering
    \includegraphics[width=0.6\linewidth]{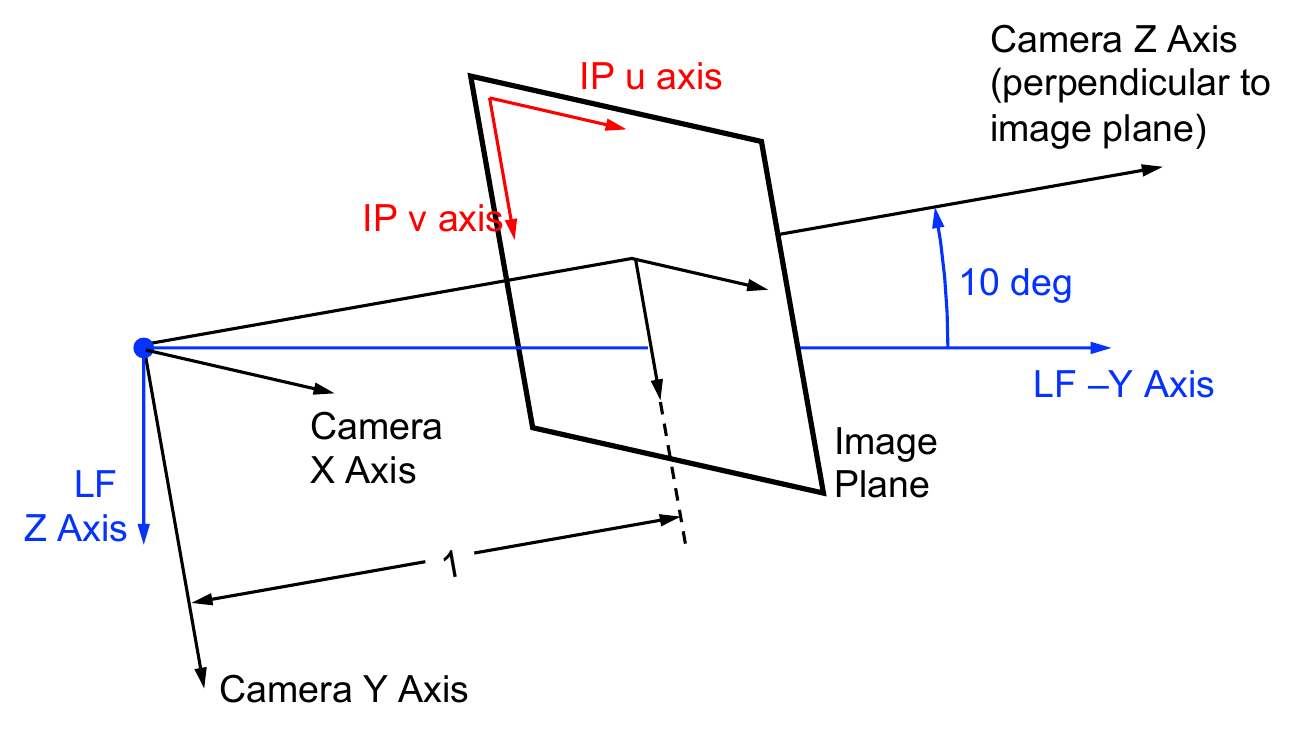}
    \caption{Illustration of camera frame, image plane, and [u,v] coordinate conventions.}
    \label{fig:ImagePlaneDefinition}
\end{figure}

Consistent with Fig.~\ref{fig:LFFramesAndFOV}, and following the conventions of Ref.~\cite{Christian:2021}, the camera frame is defined to have the +Z axis along the boresight direction (positive out of the camera). Moreover, when looking out of the camera at the image plane, the +X axis is to the right (direction of increasing image columns) and the +Y axis is down (direction of increasing image rows). This orientation is illustrated in Fig.~\ref{fig:ImagePlaneDefinition}.

The XACT contains a buffer capable of storing up to five images at a time. Once the buffer is full, the five images must be offloaded to the LF flight computer before additional images can be collected. This set of five sequential images is referred to as an \textit{Image Block}. The time between sequential images within a single Image Block varied, but was as short as about five seconds apart. An entire Image Block sequence (including five images and associated metadata) took about 110 minutes to transfer from the XACT buffer to the LF flight computer. To ensure a complete transfer of data, the start of any new Image Block was always scheduled at least 10 minutes after the conclusion of data transfer for the previous Image Block. As a result, two sequential Image Blocks were always separated by at least 120 minutes. Multiple coordinated Image Blocks in a row are referred to as an \textit{OPNAV Pass}.

In most cases, each LONEStar image was given an alphanumeric label with the first three numbers corresponding to the Image Block, followed by a letter \textit{a-e} denoting the image number within that imaging block. For example, image 536b is the second image within Image Block 536. In some instances, the target was changed in the middle of an Image Block---for example, image 593b (the second image within Image Block 593) captured Jupiter, but the subsequent three images in the Image Block captured Saturn. In these cases, the number is iterated by one to correspond to a new target, and the letter reindexes from \textit{a} (therefore, these final three images capturing Saturn were labeled 594a/b/c). 

\subsection{Lunar Flashlight Imaging Operations}

LONEStar included several OPNAV experiments. Each of these experiments required a particular sequencing of images within each Image Block and OPNAV Pass to achieve the desired objectives. The timing, pointing, and camera settings for each image were collaboratively designed between the LF Operations and OPNAV teams.

\subsubsection{LONEStar Imaging Campaign}
\label{Sec:ImagingCampaign}
\begin{figure}[t!]
    \centering
    \includegraphics[width=1\linewidth]{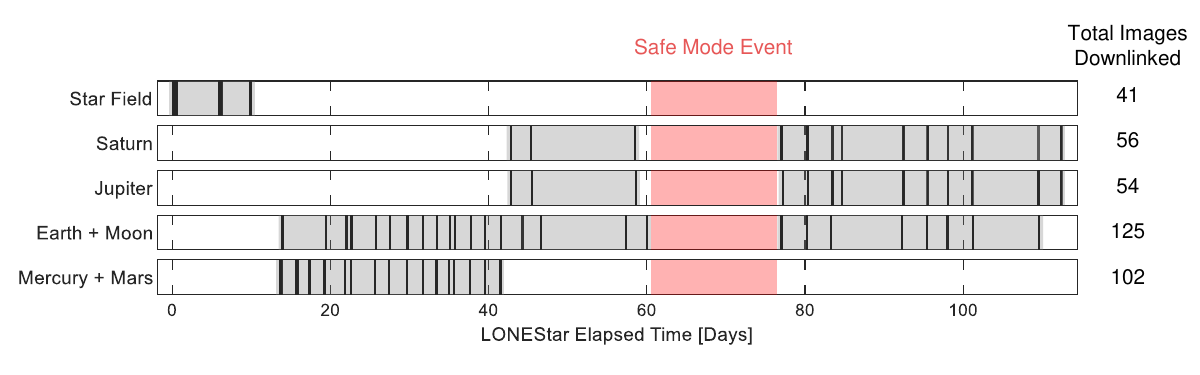}
    \caption{Timeline of LONEStar image acquisition. Shaded areas indicate windows where OPNAV targets were scheduled for imaging, with dark lines indicating image capture epochs. The reference epoch is taken to be 2023-JUL-22 00:00:00 UTC.}
    \label{fig:ImagingTimeline}
\end{figure}

The LONEStar Imaging Campaign (see Fig.~\ref{fig:ImagingTimeline}) was scheduled from late July 2023 through late November 2023. The campaign began by capturing a series of star field images for geometric calibration (see Section~\ref{Sec:GeomCal}). After acquiring star field images, attention was turned to OPNAV activities. The Earth and Moon (usually close enough to one another to be seen together in a single image) were imaged regularly for the remainder of the LONEStar campaign. Favorable geometry enabled Mercury and Mars to be captured in a single frame for most of August 2023. Following the conclusion of the Mercury and Mars imaging campaign, LONEStar turned to imaging Jupiter and Saturn. Since the apparent angle between Jupiter and Saturn was too large for concurrent imaging (varying from 66-72 deg throughout the campaign), these planets were imaged sequentially. The LONEStar experiments required these images to be taken in close succession, thus dictating that they be collected within a single Image Block (since Image Blocks are separated from one another by at least 120 minutes). Consequently, within the same Image Block, LF was commanded to capture two images of one planet (either Jupiter or Saturn) and three of the other. 

Not all of the images captured were usable for subsequent OPNAV analysis. For each target, many early Image Blocks consisted of sweeps through different exposure times and gains in order to best expose the celestial body. Only a subset of these images capture the target at an appropriate, OPNAV-quality exposure. Thus, the OPNAV results in the following sections sometimes have considerably fewer data points than one might infer from Fig.~\ref{fig:ImagingTimeline} alone.

A large interruption of the imaging campaign occurred in September (shown red in Fig.~\ref{fig:ImagingTimeline}). During this time, the spacecraft experienced a safe mode event following a latch-up believed to have been caused by solar radiation. The recovery and subsequent investigation into this event took 15 days, during which there were no images captured or downlinked.

\subsubsection{Image Block Design}

An ideal OPNAV image contains a well-exposed celestial body in the foreground against a backdrop of well-exposed stars to anchor the camera’s inertial attitude. In most cases, however, the dynamic range of the camera is insufficient to detect stars without overexposing the planets. The usual solution, which was employed by LONEStar, is to bracket short exposure OPNAV images (where no stars are visible) on both sides with longer exposure star images (where the celestial body is over-exposed). The attitude of the intermediate OPNAV image is estimated by interpolating between the attitude computed from the surrounding star images. Following this approach, the first and fifth (last) images of each LONEStar Image Block are almost always long-exposure star images. The second, third, and fourth images were typically taken with a much shorter exposure time and/or lower sensor gain setting to capture the OPNAV target without saturation. The time between sequential images in a single Image Block varied, but was no quicker than about five seconds.

For each image within an Image Block, the LF Operations and OPNAV team determined the camera boresight direction necessary to acquire the desired OPNAV image. Along with vehicle-level constraints (e.g., laser payload’s detector Sun keep-out-zone (KOZ)), the OPNAV pointing direction was used to construct a commanded LF body-frame quaternion. This commanded quaternion was visually inspected using the JPL-developed TBALL software application. If found acceptable by the LF Operations team, the attitude was added to the Image Block sequence.

\subsubsection{Sequencing, Uplink/Downlink, and Formatting}

Once an Image Block sequence was developed by the LF Operations and OPNAV teams, it was incorporated into the LF Master Events Timeline (MET). The MET was then parsed by a customized LF sequence generator that produced the commands that were uplinked to the spacecraft during each human-in-the-loop DSN contact.

\begin{figure}[b!]
    \centering
    \includegraphics[width=1\linewidth]{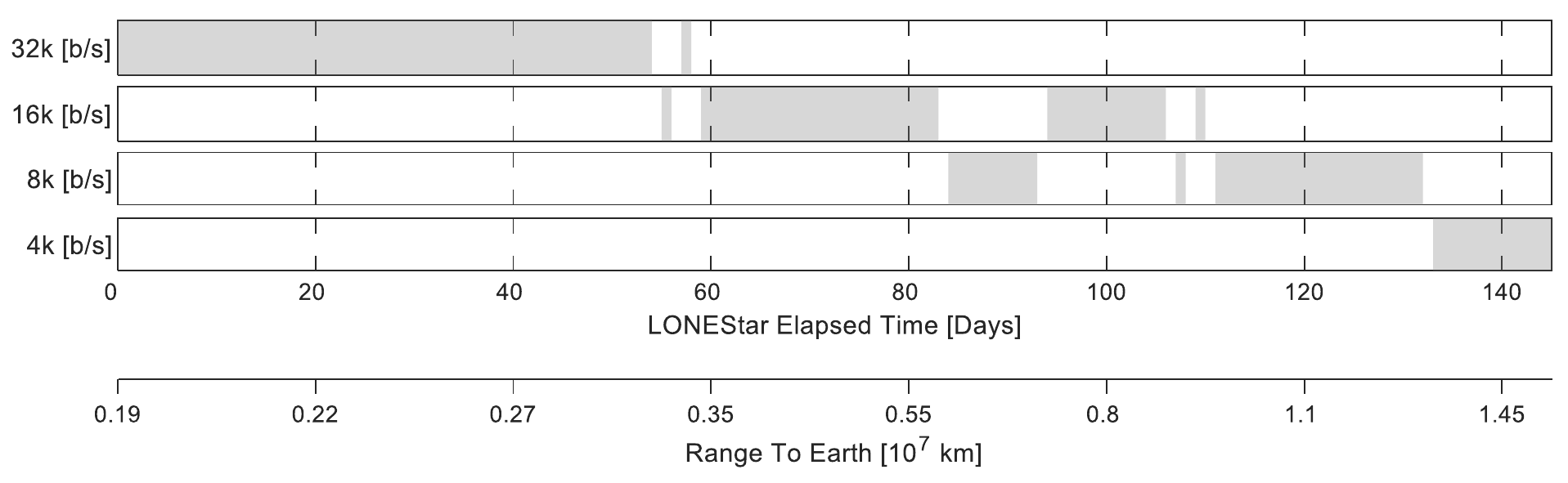}
    \caption{Timeline of data rates over the course of the LONEStar Imaging Campaign. The shaded regions refer to the time frame each data rate was used. The reference epoch is taken to be 2023-JUL-22 00:00:00 UTC.}
    \label{fig:LFDataRate}
\end{figure}

Following the successful completion of an LF OPNAV Image Block sequence, the LF Operations team facilitated the downlink of each image via DSN. Each image downlink was scheduled into the MET, which allowed the team to easily adjust for the current downlink data rate, other scheduled activities, and contact duration. The required time to downlink a single image varied from 7.3 minutes (at 32,000 bits per second) to 58.3 minutes (at 4,000 bits per second). See Fig.~\ref{fig:LFDataRate} for a timeline of the different data rates used to downlink images during LONEStar. The length of time communicating at higher data rates was extended by implementing commanding for the spacecraft to point its antenna towards Earth during a contact---the effects of this change may be seen by the jump back up to 16,000 bits per second around the 94th day of the LONEStar experiment.

Images were downlinked as a binary file, with three 10-bit pixels packed into every 4 bytes. These binary files were then unpacked and converted into TIFF images for subsequent processing. Additionally, the spacecraft telemetry, event records, and sequences were downlinked and parsed to provide information within a metadata file for each downlinked image including the commanded quaternion, image capture time, camera settings, and sequenced commands. 

\section{Camera Calibration}
\label{Sec:CameraCal}
Accurate OPNAV requires a well-calibrated camera. Of particular concern is the geometric calibration of the camera, which establishes the relationship between three-dimensional (3D) directions and two-dimensional (2D) image points. Such a correspondence enables observed points to be mapped to physical lines-of-sight (LOS) directions, which in turn may be processed by the desired OPNAV algorithms. Following standard practice, in-flight geometric calibration for LONEStar was performed using images of star fields.

\subsection{Point Spread Function (PSF) Characterization}
\label{Sec:PSF}
As is typical for star trackers \cite{Liebe:2002}, the XACT camera is intentionally defocused to improve star centroiding performance. The shape of the resulting blur (i.e., the impulse response in the spatial domain) is described by the point spread function (PSF), which is often approximated with a bivariate Gaussian distribution \cite{King:1971,Winick:1986}. Characterization of the PSF shape is an essential camera calibration task. Indeed, many OPNAV image processing techniques (e.g., subpixel horizon localization \cite{Renshaw:2020}, image restoration by deconvolution) require knowledge of the PSF as an input to the algorithm. 	

Following historical precedent, the PSF was initially modeled as a circularly symmetric, bivariate Gaussian above a constant background 
\begin{equation}
    J(u,v) = J_B + J_0 \, \text{exp} \left[- \frac{1}{2} \frac{(u - u_c)^2 + (v - v_c)^2}{\sigma^2} \right]
\end{equation}
where $J_B$ is the background brightness, $J_0$ is the PSF amplitude, $(u_c,v_c)$ is the PSF center location, and $\sigma$ is the PSF width. The units of $J,J_B,J_0$ are assumed to be digital number (DN).

The fit residuals for this model, however, were not always as low as desired, and so a bivariate Laplace distribution (which has higher kurtosis) was also considered 
\begin{equation}
    J(u,v) = J_B + J_0 \, \text{exp} \left[ - \frac{\sqrt{(u - u_c)^2 + (v - v_c)^2}}{w} \right]
\end{equation}

\begin{figure}[t!]
    \centering
    \includegraphics[width=0.8\linewidth]{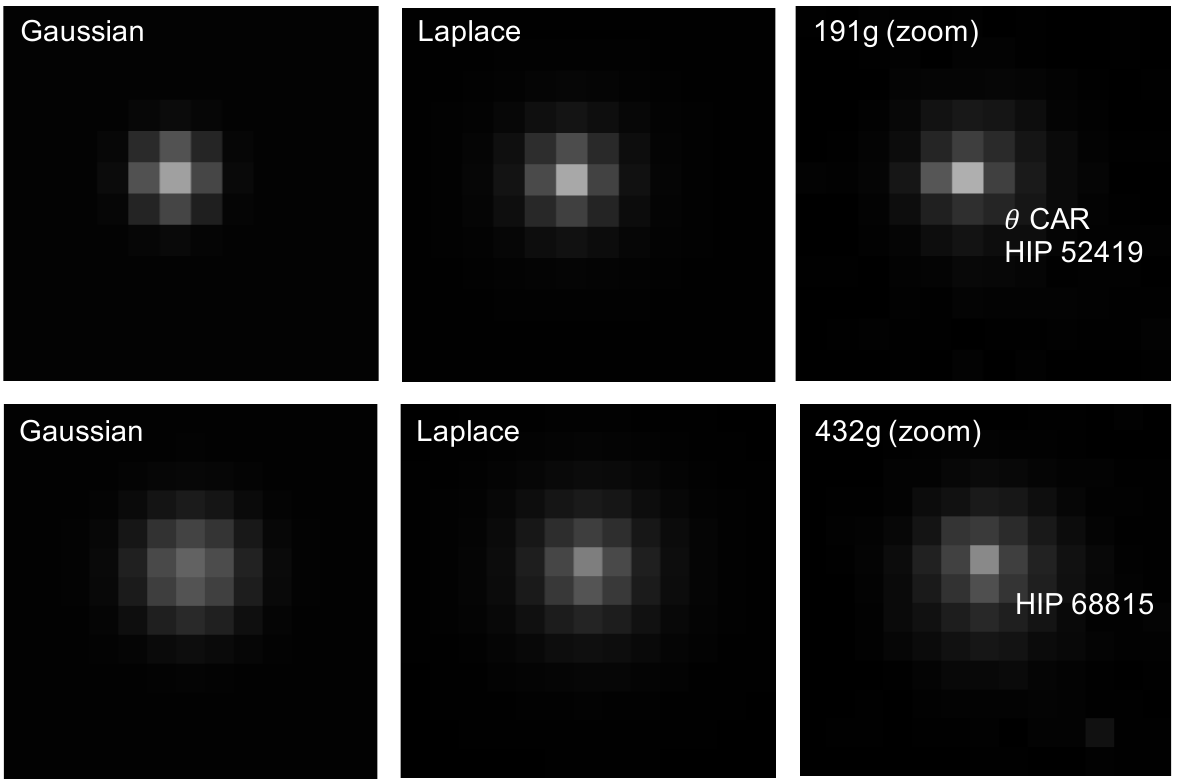}
    \caption{Comparison of best-fit Gaussian PSF (left) and best-fit Laplace PSF (middle) with actual LF image patches (right).  Top row shows an example patch of LF image 191g centered around star HIP 52419 ($\theta$ CAR, visual magnitude of 2.74). Bottom row shows an example patch of LF image 432g centered around star HIP 68815 (visual magnitude of 5.69).}
    \label{fig:PSFzoom}
\end{figure}

For both models, the five free parameters of the model are estimated by minimizing the residuals between the predicted and observed PSFs in small patches centered about each star. To match against the observed PSF (which is quantized), the continuous PSF is numerically integrated over the bounds of each pixel. The resulting nonlinear least-squares problem is solved with the Levenberg-Marquardt algorithm (LMA). Example results from two stars used in the calibration process are shown in Fig.~\ref{fig:PSFzoom}.
\begin{figure}[t!]
    \centering
    \includegraphics[width=0.7\linewidth]{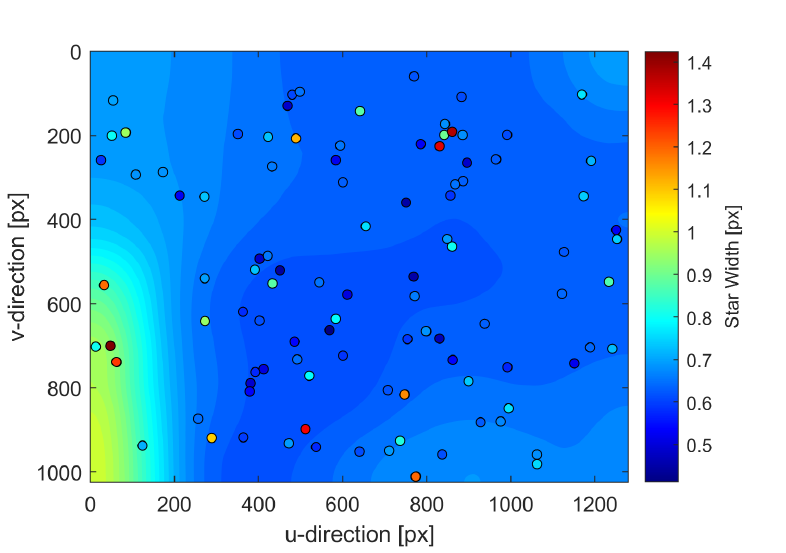}
    \caption{Laplace PSF width as estimated from 108 stars spread across the entire FOV. Contours shown here have been smoothed by application of a median filter. The data points represent the estimated widths for the 108 sampled stars.}
    \label{fig:PSFheatmap}
\end{figure}

The PSFs for 108 stars from 19 images were used to characterize the PSF throughout the sensor FOV.  The post-fit residuals were found to be smaller for the Laplace PSF than for the Gaussian PSF for 107 of the 108 stars. The PSF estimates were rather sparse and noisy (for both the Gaussian and Laplace models), and so a median filter was used to reject outliers and construct a smoothed visualization of PSF variation. Results from this analysis for the Laplace PSF are shown in Fig.~\ref{fig:PSFheatmap}. The average PSF width is found to be about $w$ = 0.72 pixels, which corresponds to a full-width, half-maximum (FWHM) of about $2\text{ln}(2)w= 2\text{ln}(2)0.72= 1.0$ pixels. 

\subsection{Camera Model}
\label{Sec:CamModel}
The camera calibration procedure aims to estimate the free parameters in a model that maps directions in the International Celestial Reference Frame (ICRF) into image pixel coordinates. Thus, before proceeding further, it is appropriate to describe this model in more detail.

Consider a celestial object (e.g., star, planet) with an ICRF direction given by $\be_i$. Since the LF spacecraft is moving at a speed of around 29-30 km/s relative to the Solar System Barycenter (SSB), it is important to account for stellar aberration. Defining the LF spacecraft's ICRF velocity as $\bv_I$, the observer's velocity as a fraction of the speed-of-light may be computed as $\bbeta = \bv_I / c$. Thus, to first order in $||\bbeta||$, the apparent ICRF direction $\be_i'$ is given by \cite{Shuster:2003}
\begin{equation}
    \be_i' = \left( \bI_{3 \times 3} - \left[ \left( \bbeta \times \be_i \right) \times \right]\right) \be_i
\end{equation}
The apparent ICRF direction $\be_i'$ must be rotated into the camera frame. Let the rotation from ICRF to the camera frame be given by $\bT_C^I$, such that the apparent LOS direction in the camera frame $\ba_i'$ is given by
\begin{equation}
    \ba_i' = \bT_{C}^{I} \be_i'
\end{equation}
The apparent direction in the camera frame $\ba_i'$ may be projected onto the image plane with the pinhole camera model, which may be compactly represented with the proportionality relationship \cite{Christian:2021}
\begin{equation}
    \bar{\bx}_i = \begin{bmatrix}
        x_i \\ y_i \\ 1
    \end{bmatrix} \propto \ba_i' = \bT_C^I \be_i'
\end{equation}
where $\bar{\bx}_i$ is the apparent image plane location in homogeneous coordinates. Recall here that the image plane is a fictitious plane located outside the camera at $z_C=1$ (see Fig.~\ref{fig:ImagePlaneDefinition}). 

The idealized image plane coordinates $\bar{\bx}_i$ must then be perturbed due to imperfections in the optical system. The LONEStar project accounts for this with the Brown-Conrady model \cite{Conrady:1919,Brown:1966,Brown:1971}, which is the same model used for the Orion OPNAV camera on Artemis I \cite{Christian:2016,Samaan:2018} and OSIRIS-REx TAGCAMS \cite{Bos:2020}. That is, the distorted (observed) image plane coordinates may be computed as
\begin{equation}
    \bx_{d_i} = \begin{bmatrix}
        x_{d_i} \\ y_{d_i}
    \end{bmatrix} = (1 + k_1 r_i^2 + k_2 r_i^4 + k_3 r_i^6) \begin{bmatrix}
        x_i \\ y_i
    \end{bmatrix} + \begin{bmatrix}
        2 p_1 x_i y_i + p_2 (r_i^2 + 2 x_i^2) \\
        p_1 (r_i^2 + 2 y_i^2) + 2 p_2 x_i y_i
    \end{bmatrix}
\end{equation}
where $r_i^2=x_i^2+y_i^2$. This can be written in homogeneous coordinates as $\bar{\bx}_{d_i}=[\bx_{d_i};1]$. Finally, the distorted image plane coordinates may be converted to pixel coordinates with the camera calibration matrix $\bK$ as

\begin{equation}
    \bar{\bu}_i = \begin{bmatrix}
        u_i \\ v_i \\ 1
    \end{bmatrix} = \begin{bmatrix}
        d_x & 0 & u_p \\
        0 & d_y & v_p \\
        0 & 0 & 1
    \end{bmatrix} \begin{bmatrix}
        x_{d_i} \\ y_{d_i} \\ 1
    \end{bmatrix} \text{ or } 
    \bar{\bu}_i = \bK \bar{\bx}_{d_i}
\end{equation}
where $d_x$ and $d_y$ are the ratio of focal length to pixel pitch and ($u_c$,$v_c$) are the pixel coordinates where the camera boresight pierces the image plane. If the pixels are square then $d_x$=$d_y$. Moreover, assuming small angles, note that $IFOV \approx 1 / d_x$.

Thus, in total, the camera model used for LONEStar has nine free parameters: five lens distortion parameters from the Brown-Conrady model ($k_1,k_2,k_3,p_1,p_2$) and four parameters for transforming from image plane to pixel coordinates ($d_x,d_y,u_p,v_p$).

\subsection{Image Processing}
\label{Sec:StarIP}
The calibration procedure requires as an input the measured pixel locations of stars in an image. This is an image processing task which aims to find the center of brightness of purposefully defocused stars. The LF OPNAV team implemented a standard star detection and centroiding algorithm.

\begin{figure}[t!]
    \centering
    \includegraphics[width=1\linewidth]{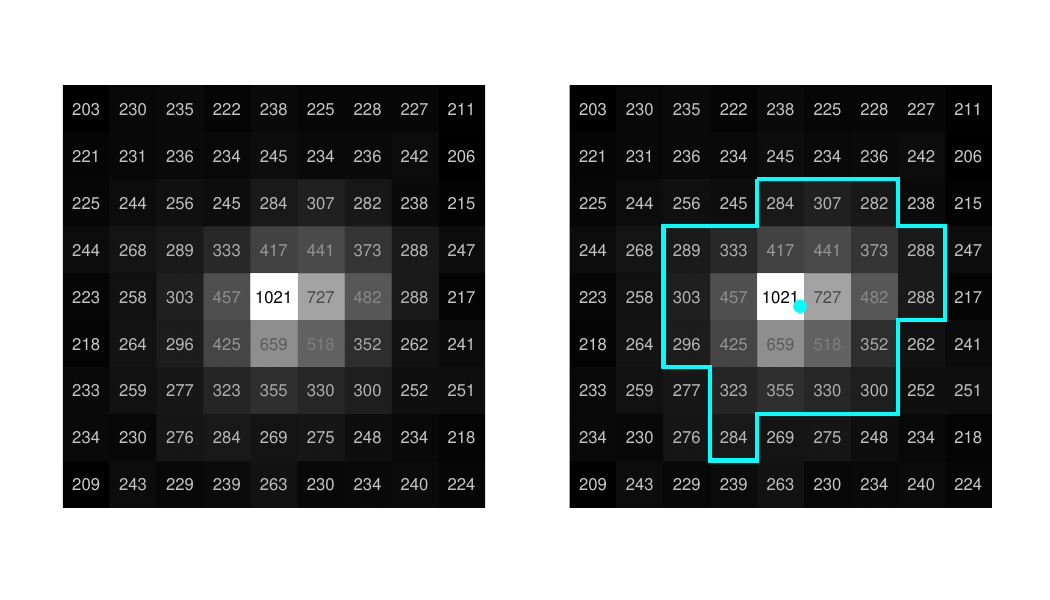}
    \caption{LF star centroiding was performed using a center of brightness (COB) algorithm. A segment of LF image 191d centered around star HIP 56561 ($\lambda$ CEN, visual magnitude of 3.11) is shown on the left. The star border used by the COB algorithm (cyan outline) and the corresponding centroid (cyan dot) are shown on the right. Each pixel is annotated with its digital number (DN).}
    \label{fig:centroidedStar}
\end{figure}

Many star field images---especially those near the Sun or Earth---experienced a noticeable amount of stray light. This was expected given the instrument’s specifications. To remove the resulting background gradient, a field flattening operation was performed. This consisted of applying a median filter to construct a background image, and then removing that background from the original image. From the flattened image, a binary image containing clusters of potential stars was formed by applying a global threshold to the flattened image. Then, a connected components analysis was used to find contiguous groups of pixels in the binary image, where only groups having $n \geq 9$  total pixels were considered as candidate stars. Final star centroids in the image were then computed for each group of pixels using a simple center of brightness (COB) computation \cite{Auer:1978,Stone:1989}.
\begin{equation}
    \label{eq:COB}
    u_{COB} = \frac{\sum_{i = 1}^n DN_i u_i}{\sum_{i = 1}^n DN_i}, v_{COB} = \frac{\sum_{i = 1}^n DN_i v_i}{\sum_{i = 1}^n DN_i}
\end{equation}
where $DN_i$ is the digital number representing the intensity of the $i$-th pixel in the group. Non-star centroids were rejected by reprojecting the Hipparcos star catalog into the image (using the best available attitude estimate) and only retaining measured centroids that corresponded to a catalog star. An example output of this procedure is shown in Fig.~\ref{fig:centroidedStar}, which shows a $9 \times 9$ section of a star field image centered on a star candidate. 

\subsection{Geometric Calibration}
\label{Sec:GeomCal}

Geometric calibration is a parameter estimation problem that seeks to find camera model coefficients which bring projections of known star LOS directions into alignment with the imaged candidate stars. The objective here is to estimate (or fix) the nine free parameters in the model described in Section~\ref{Sec:CamModel}.

Ground truth ICRF star LOS directions were constructed using the Hipparcos catalog \cite{Perryman:1997} and the standard five-parameter model given by \cite{ESA:1997}
\begin{equation}
    \be = \langle \be_0 + (t - t_{ep}) (\mu_\alpha \bp + \mu_{\delta} \bq) - \frac{\varpi r(t)}{1 AU} \rangle
\end{equation}
This model accounts for perturbations in the reference SSB direction due to both (1) the star’s proper motion ($\mu_\alpha$ and $\mu_\delta$) since the Hipparcos catalog epoch and (2) the annual parallax ($\varpi$) due to LF's position. Observed star LOS directions were acquired from a total of 35 star field images, with 282 stars uniquely identified across these images using the procedure described in Section~\ref{Sec:StarIP}. 

With corresponding Hipparcos ICRF directions $\{\be_i\}_{i=1}^n$ and measured image centroids $\{\bar{\bu}_i\}_{i=1}^n$ in hand, camera parameter estimation was performed by minimizing the reprojection errors of Hipparcos into the digital image. This was done by optimizing on the nine parameters of the camera model and corrections to the \textit{a priori} attitudes of the 35 star field images. This nonlinear least-squares problem was solved using LMA. 

LONEStar introduced a few assumptions to reduce the number of estimated camera parameters. First, the sensor pixels were assumed to be square and evenly spaced, leading to $d_x = d_y$. Furthermore, the optical center was taken to be the geometrical center of the image plane, and thus $u_p$ and $v_p$ were considered fixed exactly at the image center. In doing so, errors in boresight location were absorbed into the image attitude corrections as discussed in Ref.~\cite{Christian:2016}. This left $d_x$ as the only intrinsic parameter to be estimated. 

It is also possible to simplify the parameterization of the full Brown-Conrady model. After experimentation similar to that in Ref.~\cite{Christian:2016}, it was found that only the decentering coefficients, $p_1$ and $p_2$, and the first radial coefficient, $k_1$, were important. Thus, the LONEStar camera model assumed $k_2 = k_3 = 0$.

Including image attitude correction in the calibration process is essential and was found to significantly enhance the quality of the in-flight LONEStar calibration. 

\begin{figure}[b!]
    \centering
    \includegraphics[width=1\linewidth]{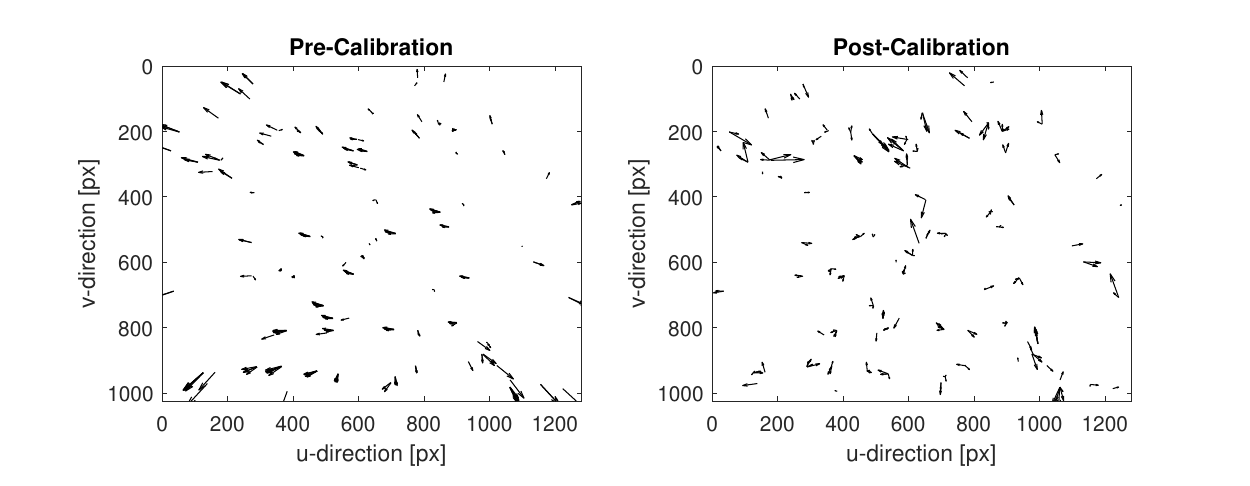}
    \caption{Quiver plot showing the reprojection residuals (scaled up for visualization) for all 282 stars used in the calibration. Structure in the pre-calibration resilduals (left) is largely removed after calibration is applied (right).}
    \label{fig:CalibQuiver}
\end{figure}

\begin{figure}[t!]
    \centering
    \includegraphics[width=1\linewidth]{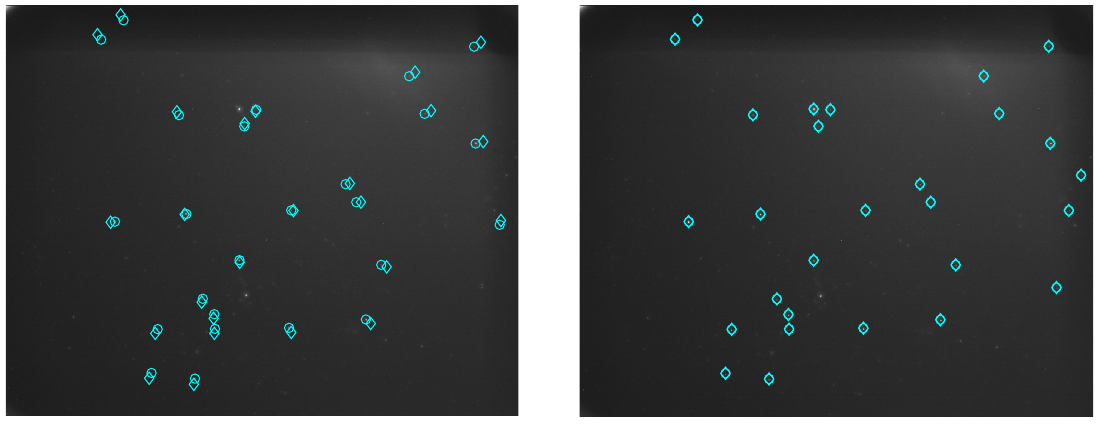}
    \caption{Reprojections of the Hipparcos catalog onto Image 191e before (left) and after (right) camera calibration. Observed stars are circles, and reprojections are diamonds. Note the pincushion distortion on the left.}
    \label{fig:CalibProj}
\end{figure}

\begin{figure}[t!]
    \centering
    \includegraphics[width=1\linewidth]{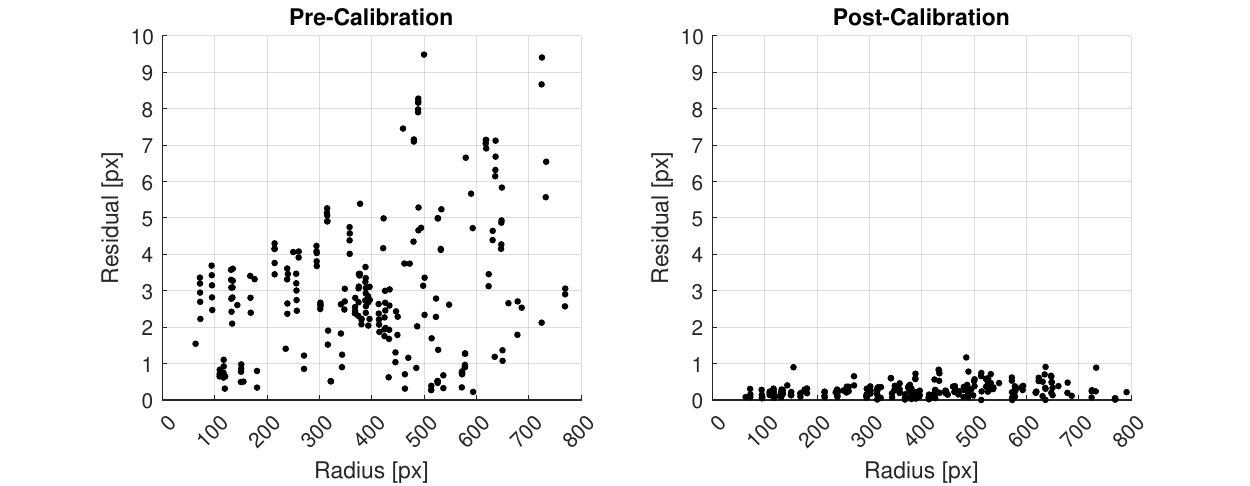}
    \caption{Magnitude of star centroid residuals as a function of radius from the center of the image. Pre-calibration residuals (left) are substantially reduced after the calibration is applied (right). Post-calibration residuals are subpixel across the entire image, indicating that the distortions are well-explained by the Brown-Conrady model.}
    \label{fig:CalibResiduals}
\end{figure}

Prior to calibration (assuming a perfect pinhole camera), a clear pincushion distortion pattern was observed (see left frames of Fig.~\ref{fig:CalibQuiver} and Fig.~\ref{fig:CalibProj}). This idealized model leads to reprojection errors on the order of about 5-10 pixels (see left frame of Fig.~\ref{fig:CalibResiduals}), depending on the distance from the image center. Application of the LONEStar calibration procedure described here reduces the reprojection error to the subpixel level across the entire image (see right frame of Fig..~\ref{fig:CalibResiduals}). Qualitatively, the high-level pincushion distortion is entirely removed in the right frame of Fig.~\ref{fig:CalibProj}.  Moreover, the residuals in Fig.~\ref{fig:CalibQuiver} appear to be randomly directed after calibration, indicating that the global structure of the lens distortion has been removed.

\subsection{OPNAV System Effective Dynamic Range}
\label{Sec:DynamicRange}
It is necessary to characterize the effective dynamic range of the end-to-end OPNAV system. The effective dynamic range (i.e., the difference between the brightest and dimmest object an optical system can detect) depends on both the camera specifications and the image processing algorithms. For example, the centroiding algorithm used by LONEStar (see Section~\ref{Sec:StarIP}) is designed for relatively high signal-to-noise ratio (SNR) objects and a larger effective dynamic range could likely be achieved with low-SNR star centroiding algorithms (e.g., image coadding \cite{Zackay:2017}). The same 35 star field images used for calibration were interrogated to empirically bound the effective dynamic range for the LONEStar experiment. For each calibration image, this analysis considered only the stars with no saturated pixels that were also successfully matched to a known star from the Hipparcos catalog. The difference in Hipparcos visual magnitude between the brightest and dimmest successfully matched star was computed for each image, giving us an empirical lower-bound on the dynamic range. The system-level dynamic range was found to be approximately $\Delta m = 5$. 

\section{Celestial Triangulation with Distant Planets}
\label{Sec:CelTriangPlanetsTop}

\subsection{Overview of Celestial Triangulation}

The selection of the specific triangulation algorithm to use for a particular spacecraft localization problem is important. Different triangulation algorithms minimize different cost functions and, consequently, are not interchangeable when LOS measurements, planet ephemerides, and camera attitude estimates are noisy. A comprehensive study of triangulation methods is provided in Ref.~\cite{Henry:2023a}, including derivations of the analytic error covariance for each. Further study of triangulation performance with different types of uncertainty may be found in Ref.~\cite{Henry:2024}. However, for the purposes of LONEStar, three specific triangulation algorithms are of primary concern: Direct Linear Transform (DLT), midpoint algorithm, and Linear Optimal Sine Triangulation (LOST). 

\subsubsection{Direct Linear Transform (DLT)}
\label{Sec:DLT}
Suppose that a spacecraft with ICRF position $\br_I$ views a distant planet at position $\bp_{I_i}$. It follows, then, that the ICRF LOS direction from the spacecraft to the planet is a point in $\PP^2$ given by

\begin{equation}
    \bell_{C_{i}} \propto \bT_{C_i}^{I} (\bp_{I_i} - \br_{I})
\end{equation}
Without (yet) considering the optimality in the presence of noisy measurements, the DLT proceeds with the purely geometric argument that the troublesome proportionality may be removed by taking the cross-product of both sides with the known measurement $\bell_{C_{i}}$. Doing so,
\begin{equation}
    [\bell_{C_i} \times ] \bell_{C_i} = [\bell_{C_i} \times ] \bT_{C_i}^{I} (\bp_{I_i} - \br_I) = \textbf{0}_{3 \times 1}
\end{equation}
hence
\begin{equation}
    [\bell_{C_i} \times ] \bT_{C_i}^I \br_I = [\bell_{C_i} \times ] \bT_{C_i}^I \bp_{I_i}
\end{equation}
For many LOS observations $\{\bell_{C_i}\}_{i=1}^n$ this relation may be stacked into a linear system to solve for the unknown $\br_I$
\begin{equation}
    \begin{bmatrix}
        [\bell_{C_1} \times ] \bT_{C_1}^{I} \\
        \vdots \\
        [\bell_{C_n} \times ] \bT_{C_n}^{I}
    \end{bmatrix} \br_I = \begin{bmatrix}
        [\bell_{C_1} \times ] \bT_{C_1}^{I} \bp_{I_1} \\
        \vdots \\
        [\bell_{C_n} \times ] \bT_{C_n}^{I} \bp_{I_n}
    \end{bmatrix}
\end{equation}
In this work, since the planet positions $\bp_{I_i}$ are assumed known from the DE440 ephemeris files \cite{Park:2021}, it is possible to directly solve for $\br_I$.

Note, however, that the scaling of the LOS direction is arbitrary and it is often helpful to describe a particular scaling of the LOS direction as $\bell_{C_i} \propto w_i \ba_i$ where, as before, $\ba_i$ is a camera frame unit vector.  Consequently, the DLT may be written as 
\begin{equation}
    \label{eq:WeightedDLT}
    \begin{bmatrix}
        w_1 [\ba_1 \times ] \bT_{C_1}^{I} \\
        \vdots \\
        w_n [\ba_n \times ] \bT_{C_n}^{I}
    \end{bmatrix} \br_I = \begin{bmatrix}
        w_1 [\ba_1 \times ] \bT_{C_1}^{I} \bp_{I_1} \\
        \vdots \\
        w_n [\ba_n \times ] \bT_{C_n}^{I} \bp_{I_n}
    \end{bmatrix}
\end{equation}
where different choices of the LOS scaling parameter $w_i$ (which act as weights on corresponding measurements) will lead to different answers in the presence of measurement noise. It will now be shown how different cost functions lead to different choices of $w_i$.

\subsubsection{Midpoint Algorithm}
\label{Sec:Midpoint}
An inertial LOS measurement describes a direction in 3D space. If that LOS direction is constrained to pass through a particular 3D point (e.g., the location of the observed planet at a specified time) then the result is a line of position (LOP) on which the spacecraft must lie. If the LOS measurements are perfect, two (or more) LOPs exactly intersect at a point. If the LOS measurements are noisy, two (or more) LOPs do not intersect at all. In this case, one reasonable cost function for optimal triangulation is to find the spacecraft location $\br_I$ with the minimum (perpendicular) distances to each of the LOPs. This is called the midpoint algorithm since, for two LOPs, the resulting estimate lies halfway along the line connecting the LOPs at their closest point. 

To proceed, recognize that the perpendicular distance of a point $\br_I$ from a LOP with direction $\be_i$ through point $\bp_{I_i}$ may be computed as
\begin{equation}
    d_i = \left(\bI - \ba_i \ba_i^T\right) \bT_{C_i}^{I} \left(\br_I - \bp_{I_i}\right) = -[\ba_i \times]^2 \bT_{C_i}^{I} \left(\br_I - \bp_{I_i}\right)
\end{equation}
The midpoint cost function proposed above would then be
\begin{equation}
    \text{min } J(\br_I) = \frac{1}{2} \sum_{i = 1}^{n} d_i^2 = \frac{1}{2} \sum_{i = 1}^{n} \left( \br_I - \bp_{I_i} \right)^T \bT_{I}^{C_i} \left[ \ba_i \times \right]^4 \bT_{C_i}^{I} \left( \br_{I} - \bp_{I_i} \right) 
\end{equation}
Applying the first differential condition and recalling that $\left[\ba_i \times \right]^4 = -\left[ \ba_i \times \right]^2$ produces an optimal estimate of $\br_I$ satisfying
\begin{equation}
    \frac{\partial J}{\partial \br_I} = \sum_{i = 1}^{n} \left( \br_I - \bp_{I_i} \right)^T \bT_{I}^{C_i} \left[ \ba_i \times \right]^2 \bT_{C_i}^{I} = \textbf{0}_{1 \times 3}
\end{equation}
Rewriting slightly yields \cite{Szeliski:2022}
\begin{equation}
    \left( \sum_{i = 1}^n \bT_{I}^{C_i} \left[ \ba_i \times \right]^2 \ \bT_{C_i}^{I} \right) \br_{I} = \sum_{i = 1}^{n} \bT_{I}^{C_i} \left[ \ba_i \times \right]^2 \bT_{C_i}^{I} \bp_{I_i}
\end{equation}
which is nothing more than the Normal equations for the DLT with $ w_i = 1 $. Thus, in practice, triangulation by the midpoint algorithm is found as the solution to the equivalent linear system of the form (just Eq.~\eqref{eq:WeightedDLT} with $ w_i = 1 $)
\begin{equation}
    \begin{bmatrix}
        [\ba_1 \times ] \bT_{C_1}^{I} \\
        \vdots \\
        [\ba_n \times ] \bT_{C_n}^{I}
    \end{bmatrix} \br_I = \begin{bmatrix}
        [\ba_1 \times ] \bT_{C_1}^{I} \bp_{I_1} \\
        \vdots \\
        [\ba_n \times ] \bT_{C_n}^{I} \bp_{I_n}
    \end{bmatrix}
\end{equation}

\subsubsection{Linear Optimal Sine Triangulation (LOST)}

In cases where measurements are simultaneous (e.g., two celestial bodies in one image) and where errors in the LOS directions are dominated by centroid localization in the image, the maximum likelihood estimate (MLE) may be found in closed form with the LOST algorithm. Thus, for the LOST algorithm, one seeks to minimize the cost function
\begin{equation}
    \text{min } J(\br_I) = \frac{1}{2} \sum_{i = 1}^{n} \left( \tilde{\bu}_i - \bu_i \right)^T \bR_{\bu_i}^{-1} \left( \tilde{\bu}_i - \bu_i \right)
\end{equation}
which is nothing more than the reprojection errors as weighted by the measurement covariance. The analytic solution to this optimization problem is given in Ref.~\cite{Henry:2023a} and only the final result is shown here.

Assuming the centroiding errors are isotropic and uncorrelated, the LOST weighting of the DLT is 
\begin{equation}
    w_i = \frac{|| \bar{\bx}_i || }{\sigma_x} \frac{|| \bT_{I}^{C_i} \bar{\bx}_i \times \bT_{I}^{C_i} \bar{\bx}_j ||}{|| d_{ij} \times \bT_{I}^{C_i} \bar{\bx}_j||}
\end{equation}
To avoid unnecessary normalization of image plane coordinates, note that $ \bar{\bx}_i = \| \bar{\bx}_i \| \ba_i$ such that one may write $w_i \left[ \ba_i \times \right] = q_i \left[ \bar{\bx}_i \times \right]$ where

\begin{equation}
    q_i =  \frac{1}{\sigma_x} \frac{|| \bT_{I}^{C_i} \bar{\bx}_i \times \bT_{I}^{C_i} \bar{\bx}_j ||}{|| d_{ij} \times \bT_{I}^{C_i} \bar{\bx}_j||}
\end{equation}
The LOST algorithm removes the redundant row in $q_i \left[\bar{\bx}_i \times \right]$ with the $2 \times 3$ matrix $\bS = \left[ \bI_{2 \times 2} , \textbf{0}_{2 \times 1} \right]$, such that
\begin{equation}
    \label{eq:LOSTLinSys}
    \begin{bmatrix}
        q_1 \bS [\bar{\bx}_1 \times ] \bT_{C_1}^{I} \\
        \vdots \\
        q_n \bS [\bar{\bx}_n \times ] \bT_{C_n}^{I}
    \end{bmatrix} \br_I = \begin{bmatrix}
        q_1 \bS [\bar{\bx}_1 \times ] \bT_{C_1}^{I} \bp_{I_1} \\
        \vdots \\
        q_n \bS [\bar{\bx}_n \times ] \bT_{C_n}^{I} \bp_{I_n}
    \end{bmatrix}
\end{equation}
where $\bp_{I_i}$ is the location of the planet when the observed light was reflected by the body (some time $\Delta t$ before the time of image capture). 

Finally, an important feature of the LOST method is that light time-of-flight (LTOF) may be accounted for directly within the triangulation solution to within a few milliarcseconds without iteration. Introducing the term 
\begin{equation}
    \bm_i = \frac{\| \bar{\bx}_i \|}{\sigma_x} \frac{\bv_i}{c}
\end{equation}
where $\bv_i$ is the velocity of the observed celestial body and $c$ is the speed of light, the LTOF-corrected triangulation solution is simply \cite{Henry:2023b}
\begin{equation}
    \label{eq:LOSTwithLTOF}
    \begin{bmatrix}
        q_1 \bS [\bar{\bx}_1 \times ] \bT_{C_1}^{I} \\
        \vdots \\
        q_n \bS [\bar{\bx}_n \times ] \bT_{C_n}^{I}
    \end{bmatrix} \br_I = \begin{bmatrix}
        \bS [\bar{\bx}_1 \times ] \bT_{C_1}^{I} (q_1 \bp_{I_1}^+ - \bm_{I_1}) \\
        \vdots \\
        \bS [\bar{\bx}_n \times ] \bT_{C_n}^{I} (q_n \bp_{I_n}^+ - \bm_{I_n})
    \end{bmatrix}
\end{equation}
where $\bp_i^+$ is the planet location at the time of image capture.

\subsection{Image Processing and Attitude Detection}
\label{Sec:IPandAttDet}

Successful triangulation relies on ICRF LOS measurements to the observed planets. The distant planets observed by LONEStar (Mercury, Mars, Saturn, Jupiter) were unresolved and appear similar to stars. This allows for use of the same centroiding algorithm as described in Section~\ref{Sec:StarIP}, but with some minor adjustments to brightness and cluster size thresholding. 

The planet centroid location provides a LOS measurement in the camera frame that must be transformed into ICRF for triangulation and navigation. Ideally, the attitude could be determined from background stars within the OPNAV image itself. However, the LONEStar effective dynamic range (see Section~\ref{Sec:DynamicRange}) and viewing geometry (see scenario-specific discussions), prevented concurrent imaging of planets and stars. When possible, short exposure OPNAV images were bracketed on both sides with long exposure star images. The attitude at the intermediate time of the OPNAV image was then estimated using spherical linear interpolation (SLERP) \cite{Shoemake:1985}. For some OPNAV images, it was only possible to obtain one corresponding long exposure star image. For example, when performing sequential imaging of Jupiter and Saturn, the limitation of five images per Image Block means that only one of the OPNAV images could be bracketed on both sides with star images (usually Jupiter). The remaining planet OPNAV image (usually Saturn) had only a single star image.

Attitude determination with these long-exposure star field images was computed as an update to the commanded attitude from LF telemetry. Star LOS measurements were obtained and matched against a Hipparcos reprojection computed with the known commanded attitude. In some scenarios (particularly with images of Mercury and Mars), excessive background light in the image drove the choice for a low signal-to-noise threshold while screening star candidates, resulting in a multitude of false (non-star) candidates. Matching between star candidates and known catalog stars was thus a multi-step process that proved effective at rejecting false-positive stars in the face of exceptionally noisy images. First, the nearest catalog neighbors to each candidate point are obtained, and a simple threshold is applied, filtering out candidates with no close-by catalog star. Then, a nearest-neighbor ratio test is conducted for each candidate star, ensuring that a match with any nearby catalog entry is unique. It is natural to expect that any attitude offset from the reprojection will manifest as a shift of pixels in a consistent direction and magnitude across the entire image. As a result, the final step is to calculate the median pixel residual direction and magnitude from the set of multiple potentially valid observation-catalog pairs, and retain only pairs that fall within a specified tolerance of these median values. 

\subsection{Instantaneous Triangulation: Mercury and Mars}
\label{Sec:InstantTriMercuryMars}
\begin{figure}[b!]
    \centering
    \includegraphics[width=1\linewidth]{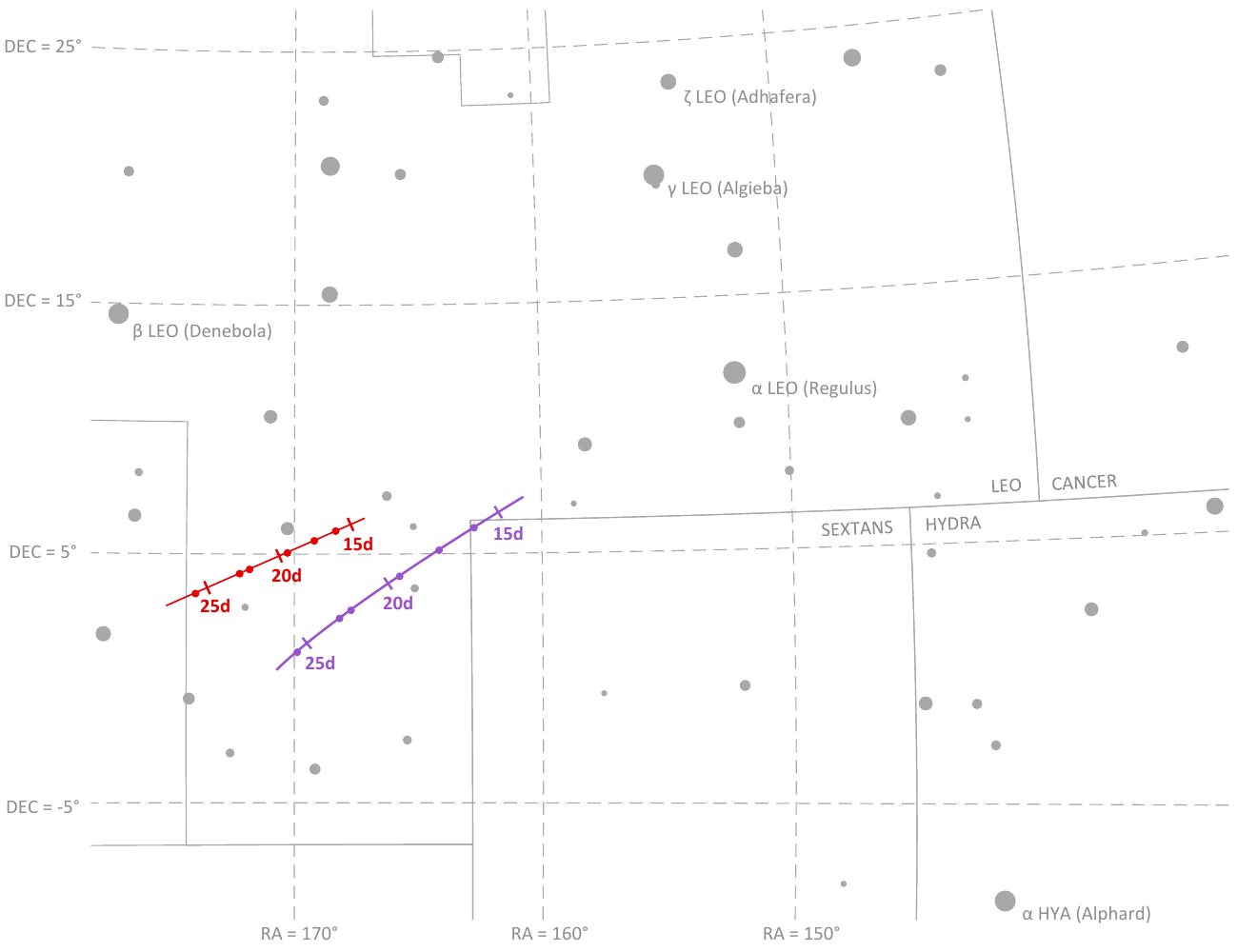}
    \caption{Apparent motion of Mercury (purple) and Mars (red) on the celestial sphere as seen by Lunar Flashlight. The measured LOS directions to these planets obtained from LONEStar OPNAV images are shown as dots. Tick marks along the planet tracks indicate elapsed time from the LONEStar reference epoch.}
    \label{fig:MercuryMarsRADEC}
\end{figure}

\begin{figure}[t!]
    \centering
    \includegraphics[width=0.8\linewidth]{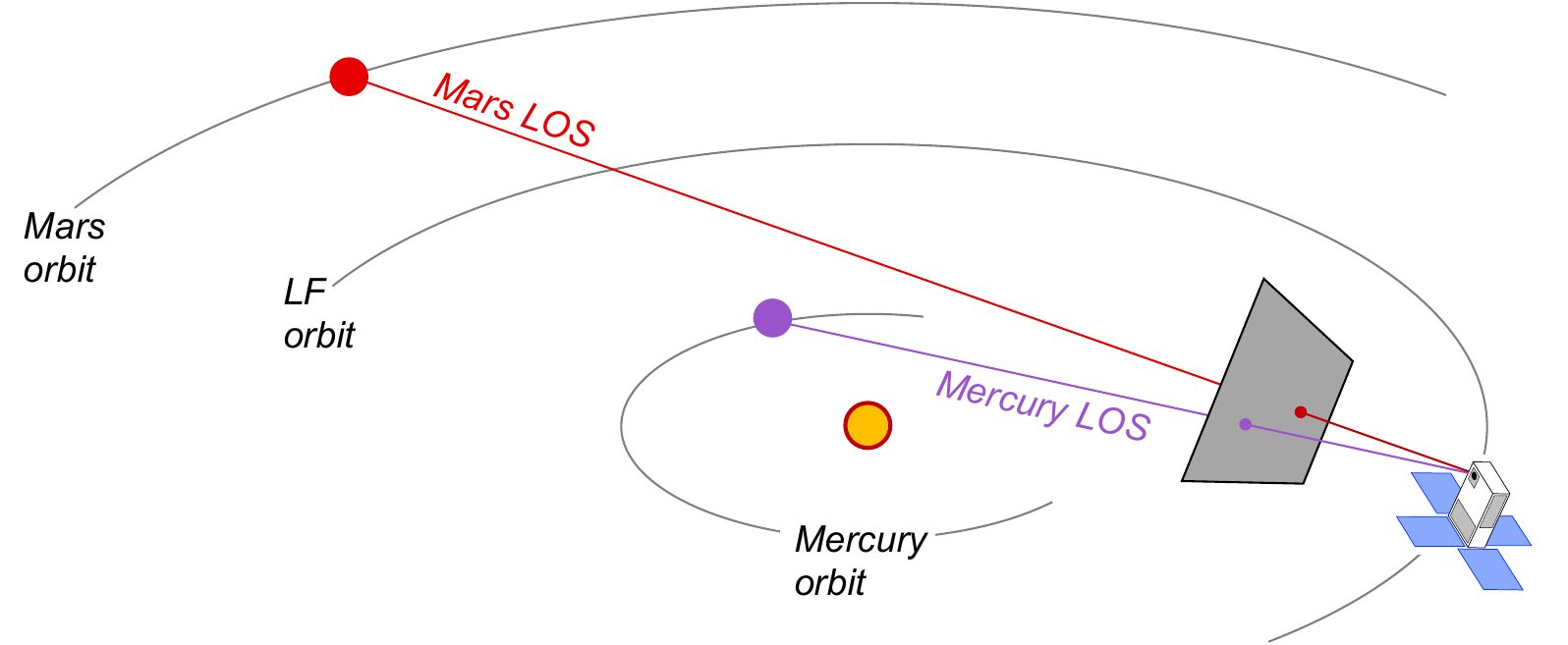}
    \caption{Lunar Flashlight was able to instantaneously triangulate its position with simultaneous observations of Mercury and Mars.}
    \label{fig:MercuryMarsCartoon}
\end{figure}

For a brief period of time in August 2023, favorable geometry allowed Lunar Flashlight to simultaneously observe Mercury and Mars in a single image. Six simultaneous observations of Mercury and Mars were successfully acquired as part of the LONEStar imaging campaign. The apparent path of both Mercury and Mars (as seen from LF) against the star field background may be seen in Fig.~\ref{fig:MercuryMarsRADEC}, with the resulting measured LOS directions to each planet shown as dots on the celestial sphere. This was the only time during the LONEStar imaging campaign when two distant planets were simultaneously visible with the camera. By capturing two planets in the same image (see Fig.~\ref{fig:MercuryMarsCartoon}), it is possible to demonstrate instantaneous triangulation using the LOST algorithm.

\begin{figure}[t!]
    \centering
    \includegraphics[width=0.8\linewidth]{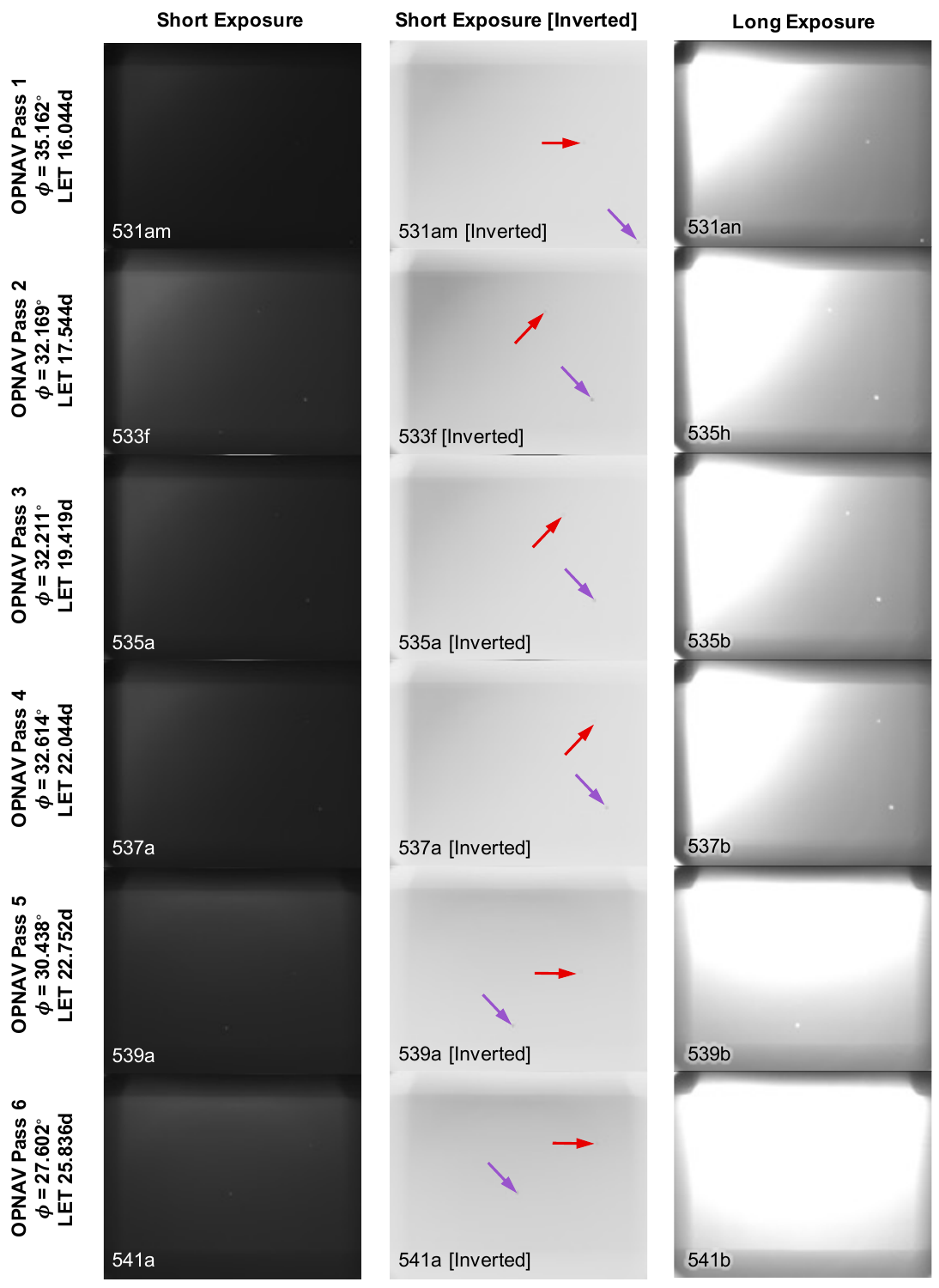}
    \caption{Tiling of all images used to observe Mars and Mercury. Arrows in the inverted image point to Mercury (purple) and Mars (red). Image capture times are reported in LONEStar Elapsed Time (LET) with units of days.}
    \label{fig:MercuryMarsTiling}
\end{figure}

Although the August 2023 geometry permitted simultaneous imaging of two planets (apparent inter-planet angle of about 4.5-5.5 deg, see Table~\ref{tab:MarsMercuryInstTriang}), the lighting condition was challenging. Indeed, the Sun was within the camera’s manufacturer-recommended 45 deg Sun KOZ for the entirety of this imaging opportunity. The lighting was best (although still not good) at earlier dates, and deteriorated as time advanced and the angle between the planet and Sun decreased (as can be seen from top to bottom in Fig.~\ref{fig:MercuryMarsTiling} and Table~\ref{tab:MarsMercuryInstTriang}). Stray light from the Sun was a pervasive and performance-limiting effect for all of the Mercury and Mars OPNAV images. Nevertheless, the OPNAV solutions presented here are quite reasonable, especially when considering that the images were captured substantially outside of the instrument’s recommended operating conditions.

\begin{table}[h!t]
    \centering
    \caption{Apparent angular separation between Mercury and Mars for the instantaneous triangulation experiment.}
    \begin{tabularx}{\textwidth}{lccYYY}
    \toprule
    \multirow{3}{*}{Image} & \multirow{3}{2.6 cm}{\centering LONEStar \\ Elapsed Time [days]}  & \multirow{3}{1.6 cm}{\centering Sun Angle \\ ($\phi$) [deg]}  & \multicolumn{3}{c}{Apparent Angle Between Mars and Mercury}\\
    \cmidrule(lr){4-6}
     &  &  & DSN + DE440 [deg] & Measured [deg] & Residual [arcsec] \\
    \midrule
    531am & 16.044 & 35.162 & 5.53432 & 5.53906 & 17.064 \\
    533f  & 17.544 & 32.169 & 5.03656 & 5.03164 & 17.712 \\
    535a  & 19.419 & 32.211 & 4.61647 & 4.61332 & 11.340  \\
    537a  & 22.044 & 32.614 & 4.40068 & 4.39791 & 9.972  \\
    539a  & 22.752 & 30.438 & 4.41016 & 4.40610  & 14.616 \\
    541a  & 25.836 & 27.602 & 4.72606 & 4.72148 & 16.488 \\
    \bottomrule
    \end{tabularx}
    \label{tab:MarsMercuryInstTriang}
\end{table}

The instrument did not have sufficient dynamic range (see Section~\ref{Sec:DynamicRange}) to simultaneously capture well-exposed planets and background stars, and thus pairs of short and long-exposure images were required. The long exposure and short exposure images were separated by about 5 seconds from one another.  It was not generally possible to bracket short-exposure planet images on both sides with long-exposure star images for the Mercury and Mars observations. An attitude stability study performed with images of Jupiter and Saturn (see Section~\ref{Sec:SequentialTriJupSat}) suggests that the LF pointing errors over a 5-second interval are likely on the order of 15-20 arcsec (i.e., about 0.5 pixel). 

\begin{figure}[b!]
    \centering
    \includegraphics[width=1\linewidth]{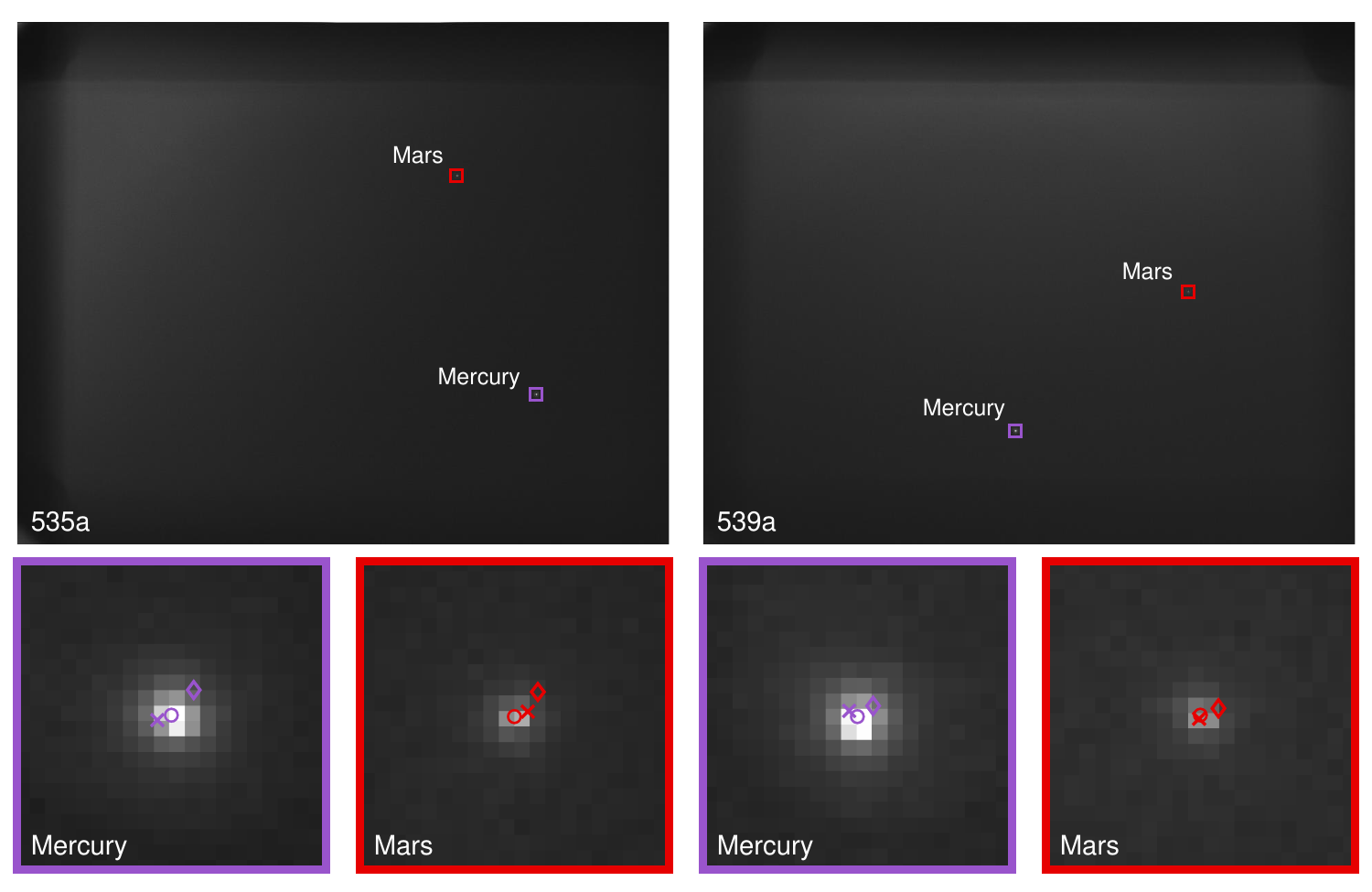}
    \caption{Two observation images with zoomed in views of Mars and Mercury. The $\circ$ indicates a measured centroid, $\diamond$ indicates a centroid projection from the JPL-produced navigation solution, and $\times$ indicates the reprojection from the OPNAV-produced solution.}
    \label{fig:MercuryMarsCentroid}
\end{figure}

\begin{figure}[b!]
    \centering
    \includegraphics[width=1\linewidth]{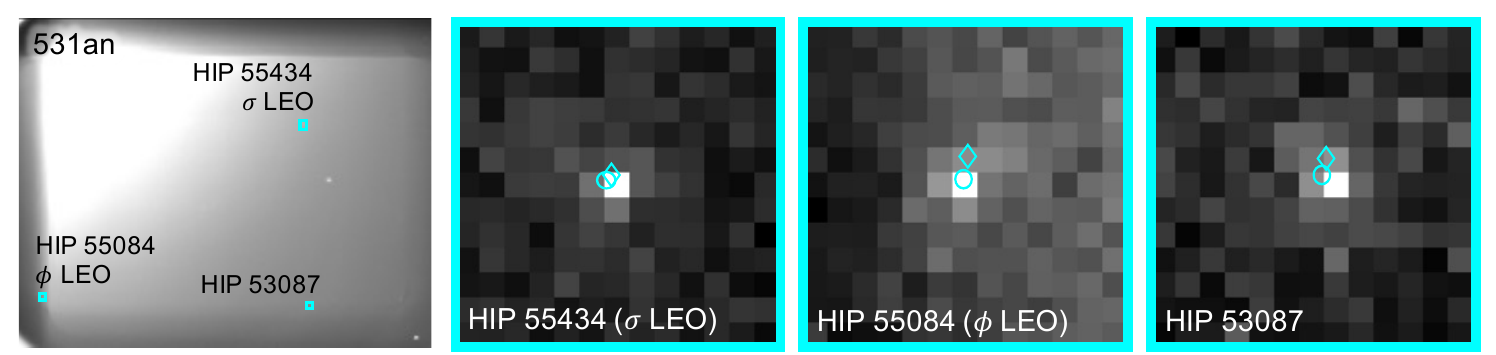}
    \caption{Three stars from a long-exposure image of Mars and Mercury. The $\circ$ indicates a measured centroid and the $\diamond$ indicates a centroid projection from the Hipparcos star catalog. Note that the brightness of each patch around these stars has been renormalized such that the darkest pixel is black and the brightest is white.}
    \label{fig:MercuryMarsStar}
\end{figure}

Short-exposure images were of reasonable quality even within the Sun KOZ, with properly exposed planets that could be easily centroided (see Fig.~\ref{fig:MercuryMarsCentroid}). These short exposure images had no detectable stars. Conversely, long exposure times revealed numerous stars in each image, but the planets became saturated and a significant amount of stray light saturated entire regions of the image. Despite these challenges, star centroiding, matching, and attitude determination was possible (see Fig.~\ref{fig:MercuryMarsStar}; note that the Bayer designations in this figure---and all subsequent figures---were determined from Ref.~\cite{Kostjuk:2002}). Unfortunately, as the sun angle $\phi$ decreased, so too did the quality of the attitude correction that could be derived from these long-exposure images, in turn worsening the quality of the triangulation solution.

For simultaneous LOS measurements, LOST provides the statistically optimal triangulation solution without iteration. Performance of the LOST algorithm (see Eq.~\eqref{eq:LOSTLinSys} and Ref.~\cite{Henry:2023a}) for localizing LF using each of the six images containing both Mercury and Mars is summarized in Table~\ref{tab:MarsMercuryCamRes}. Residuals are reported relative to the JPL-produced reference trajectory obtained from radiometric observables.

\begin{table}[t!]
    \centering
    \caption{Instantaneous triangulation residuals using images containing both Mercury and Mars.}
    \begin{tabularx}{\textwidth}{l P{1.1 cm} P{1.1 cm}P{1.1 cm} P{1.1 cm} P{1 cm} P{1.8 cm}P{1.8 cm}}
    \toprule
    \multirow{4}{*}{Image} & \multicolumn{3}{c}{Camera Frame Residual [km]}  & \multicolumn{4}{c}{Residual Norm}\\
    \cmidrule(lr){2-4} \cmidrule(lr){5-8}
     & \multirow{3}{*}{X} & \multirow{3}{*}{Y} & \multirow{3}{*}{Z} & \multirow{3}{*}{$10^5$ km} & \multirow{3}{1 cm}{\centering Earth Radii} & Normalized by range to Mercury & \multirow{3}{1.8 cm}{\centering Normalized by range to Mars} \\
    \midrule
    531am & 1,749.1 & -23,350 & 63,582 & 0.6776  & 10.623  & 0.0004888 & 0.0001892 \\
533f  & 32,900  & -44,241 & -238,522 & 2.4481  & 38.384  & 0.0018089 & 0.0006817 \\
535a  & 23,135  & -63,283 & -468,496 & 4.7332  & 74.211  & 0.0036064 & 0.0013140 \\
537a  & 16,407  & -56,088 & -531,709 & 5.3491  & 83.868  & 0.0042614 & 0.0014789 \\
539a  & 37,628  & -45,662 & -756,179 & 7.5849  & 118.923 & 0.0061173 & 0.0020948 \\
541a  & 71,145  & -57,681 & -188,000 & 18.8223 & 295.113 & 0.0160341 & 0.0051750 \\
    \bottomrule
    \end{tabularx}
    \label{tab:MarsMercuryCamRes}
\end{table}

Assuming isotropic centroid errors with image plane covariance $\sigma_x=(\sigma_u)(IFOV)$ the error covariance of the MLE triangulated solution (whether by LOST or a classical iterative scheme) is given by \cite{Henry:2023a}
\begin{equation}
    \label{eq:LOSTcovariance}
    \bP_r = \left[ \sum_{i = 1}^n q_i^2 \bT_{P}^{C_i} \left[ \bK_i^{-1} \bu_i \times \right]^{T} \bS^T \bS \left[ \bK_i^{-1} \bu_i \times \right] \bT_{C_i}^{I} \right]^{-1}
\end{equation}
where everything in this equation is known and $\bP_r$ may be readily computed. In the special case of a narrow FOV camera and only two observations, the generic covariance may be simplified and the total error $\sigma_{tot} = \sqrt{Tr(\bP)}$ may be approximated as \cite{Broschart:2019}
\begin{equation}
    \label{eq:JPLcovariance}
    \sigma_{tot} = \sigma_{x} \frac{\sqrt{\rho_1^4 + \rho_1^2 \rho_2^2 \sin^2{\theta_{12}} + 2 \rho_1^2 \rho_2^2 + \rho_2^4}}{\sqrt{\rho_1^2 + \rho_2^2} \sin{\theta_{12}}}
\end{equation}
Of note is that the simplified total error in Eq.~\eqref{eq:JPLcovariance} from Ref.~\cite{Broschart:2019} was derived in a completely different manner than the generic covariance in Eq.~\eqref{eq:LOSTcovariance} from Ref.~\cite{Henry:2023a}. Nevertheless, some simple analysis will show the results to be consistent with one another, which is numerically demonstrated in Table~\ref{tab:MarsMercuryErrStat}.

\begin{table}[b!]
    \centering
    \caption{Instantaneous triangulation error statistics using images containing both Mercury and Mars.}
    \begin{tabularx}{\textwidth}{lP{2 cm} P{2 cm} YYY}
    \toprule
    \multirow{3}{*}{Image} & \multirow{3}{2 cm}{\centering Residual Norm \\ \hspace{0pt} [$10^5$ km]}  & \multicolumn{3}{c}{Total Error - 1 Sigma [km]} & \multirow{3}{2 cm}{\centering Mahalanobis \\ Distance [-]}\\
    \cmidrule(lr){3-5}
     &  &  $\sqrt{Tr(\bP)}$ & Approximated & Difference  \\
    \midrule
    531am & 0.6776  & 880,997   & 885,221   & -4,224 & 0.998 \\
    533f  & 2.4481  & 971,891   & 971,642   & 249    & 2.131 \\
    535a  & 4.7332  & 1,059,016 & 1,058,811 & 205    & 2.677 \\
    537a  & 5.3491  & 1,107,899 & 1,109,167 & -1,268 & 2.673 \\
    539a  & 7.5849  & 1,107,464 & 1,106,409 & 1,055  & 1.687 \\
    541a  & 18.8223 & 1,033,201 & 1,031,179 & 2,022  & 3.993 \\
    \bottomrule
    \end{tabularx}
    \label{tab:MarsMercuryErrStat}
\end{table}

Additionally, it is important to consider light time-of-flight (LTOF) effects. There are a great many ways to account for LTOF, and results for two such methods are summarized in Table~\ref{tab:MarsMercuryLTOFCorr}. The first method uses an iterative approach to appropriately backdate the ephemeris catalog query to the time when photons were reflected by the celestial body instead of the time of the image collection. This was mechanized for LONEStar using the converged Newtonian (CN) option within the SPICE toolkit \cite{Acton:1996,Acton:2018}. Another option is the analytic and non-iterative LTOF correction that may be written directly into LOST (see Eq.~\ref{eq:LOSTwithLTOF} and Ref.~\cite{Henry:2023b}). Since both of the LTOF corrections give essentially the same solution, all subsequent discussions use the non-iterative method from Ref.~\cite{Henry:2023b}.

\begin{table}[h!t]
    \centering
    \caption{Comparison of instantaneous triangulation results using various light time-of-flight corrections.}
    \begin{tabularx}{0.75\textwidth}{lYYY}
    \toprule
    \multirow{3}{*}{Image} & \multicolumn{3}{c}{Residual Norm [km]} \\
    \cmidrule(lr){2-4} 
    & SPICE CN LTOF Correction & LOST LTOF Correction & \multirow{2}{*}{Difference}  \\
    \midrule
    531am & 67,657    & 67,758    & -101  \\
    533f  & 245,154   & 244,811   & 343   \\
    535a  & 473,943   & 473,316   & 627   \\
    537a  & 535,668   & 534,911   & 757   \\
    539a  & 759,473   & 758,490   & 983   \\
    541a  & 1,884,248 & 1,882,233 & 2,015 \\
    \bottomrule
    \end{tabularx}
    \label{tab:MarsMercuryLTOFCorr}
\end{table}

\subsection{Sequential Triangulation: Jupiter and Saturn}
\label{Sec:SequentialTriJupSat}
In most cases, simultaneous access to two or more planet LOS measurements is not achievable. This makes instantaneous triangulation of the type discussed in Section~\ref{Sec:InstantTriMercuryMars} impossible. However, if two of more planets can be viewed in rapid succession, an acceptable triangulation solution may still be formed. LONEStar demonstrated this concept on sequential images of Jupiter and Saturn.

\begin{figure}[t!]
    \centering
    \includegraphics[width=0.6\linewidth]{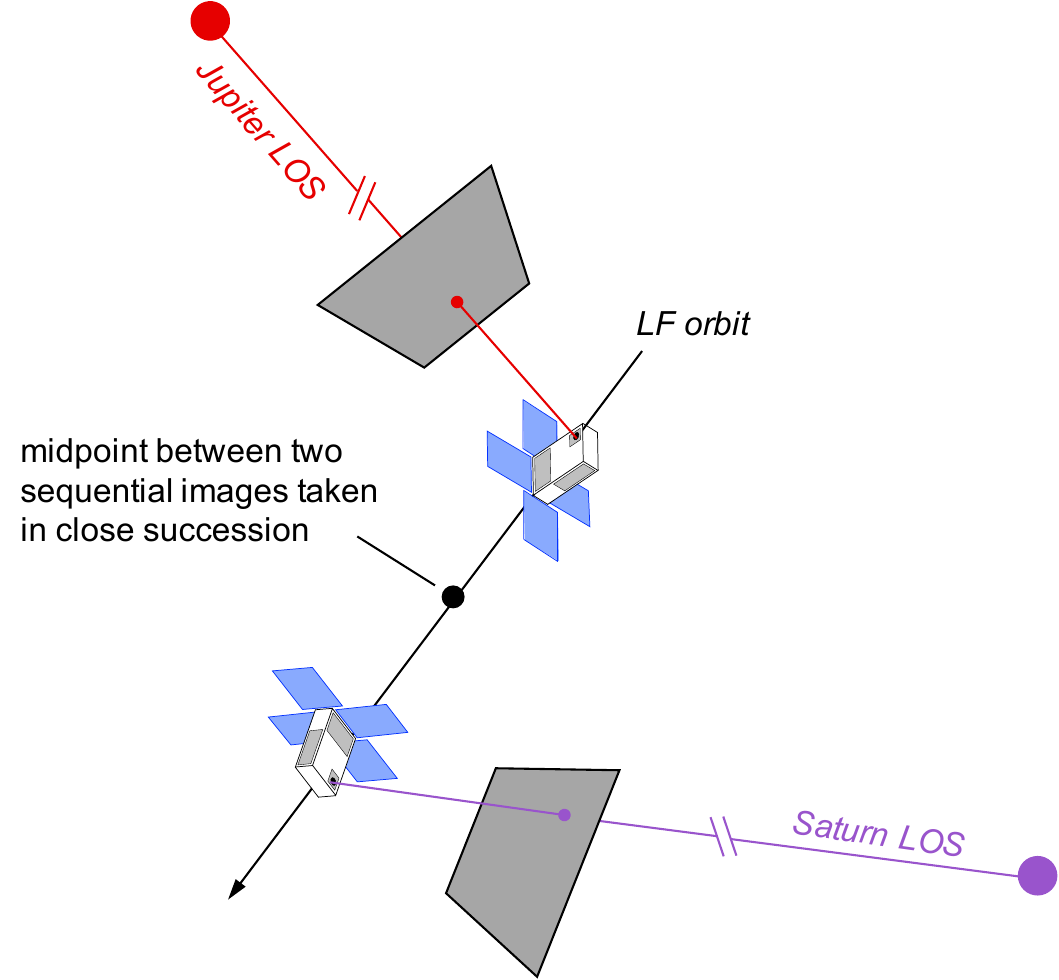}
    \caption{Using the midpoint algorithm, Lunar Flashlight was able to triangulate its position with two sequential observations of Jupiter and Saturn.}
    \label{fig:JupiterSaturnCartoon}
\end{figure}

LONEStar performs triangulation with sequential images using the midpoint algorithm. The selection of the midpoint algorithm instead of LOST is intentional. While LOST (and its variants) provide statistically optimal triangulation, the maximum likelihood cost function on which LOST is based fundamentally assumes that the observer resides at a single point. With sequential images from a moving spacecraft, however, each observation originates from a different point. When there are only two observations, a reasonable estimate of the spacecraft’s location at a time halfway between the observations is the point halfway along the line having the shortest distance between the two LOPs (see Fig.~\ref{fig:JupiterSaturnCartoon}). This is exactly the position that the midpoint algorithm provides (see Section~\ref{Sec:Midpoint}). Of special note is that the midpoint algorithm only makes sense when there are exactly two images, and it would not be appropriate for sequential triangulation with three (or more) sequential images. See Section~\ref{Sec:DynamicTri} for a discussion of how to process many sequential images.

\begin{figure}[b!]
    \centering
    \includegraphics[width=1\linewidth]{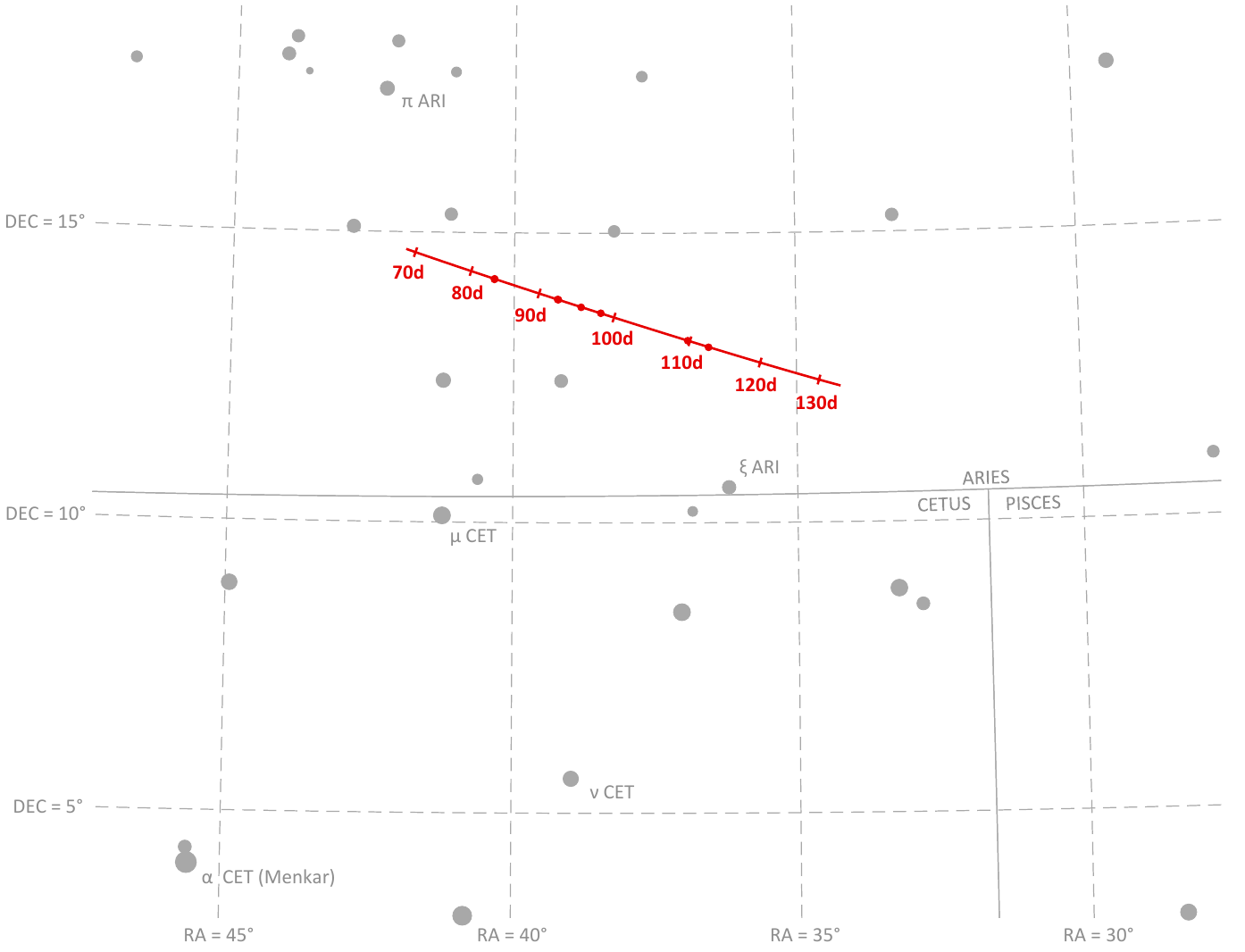}
    \caption{Apparent motion of Jupiter (red) on the celestial sphere as seen by Lunar Flashlight. The measured LOS directions to Jupiter obtained from LONEStar OPNAV images are shown as dots. Tick marks along Jupiter's track indicate elapsed time from the LONEStar reference epoch.}
    \label{fig:JupiterRADEC}
\end{figure}

\begin{figure}[t!]
    \centering
    \includegraphics[width=1\linewidth]{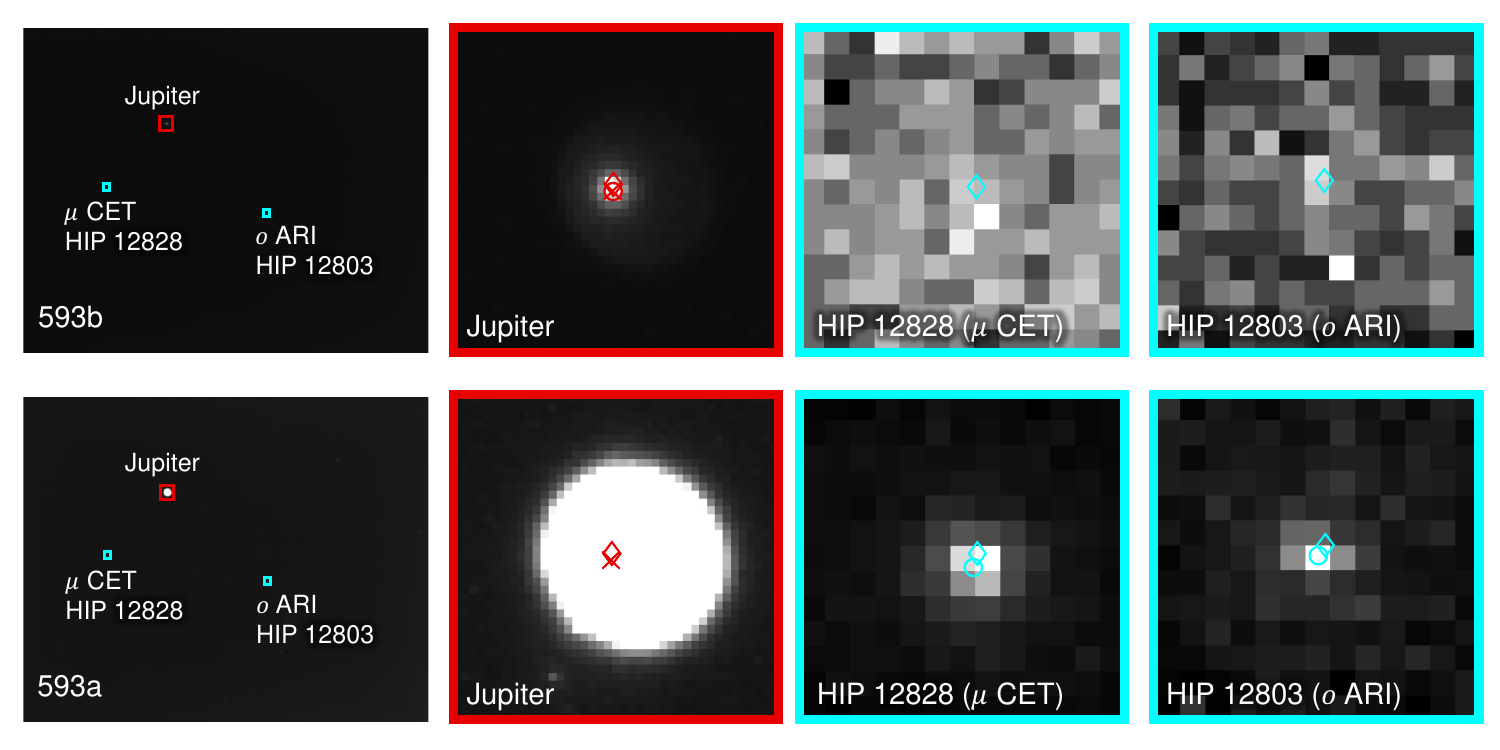}
    \caption{A pair of sequential short exposure (top) and long exposure (bottom) images of Jupiter. The OPNAV LOS measurement is obtained from the top image where Jupiter is properly exposed, while the camera attitude is obtained from the bottom image where stars are visible. The $\circ$ indicates a measured centroid, $\diamond$ indicates a centroid projection from the JPL-produced navigation solution, and $\times$ indicates the reprojection from the OPNAV-produced solution. Each zoomed patch has been renormalized such that the darkest pixel is black and the brightest is white.
    }
    \label{fig:Jupiter593ab}
\end{figure}

\begin{figure}[b!]
    \centering
    \includegraphics[width=0.8\linewidth]{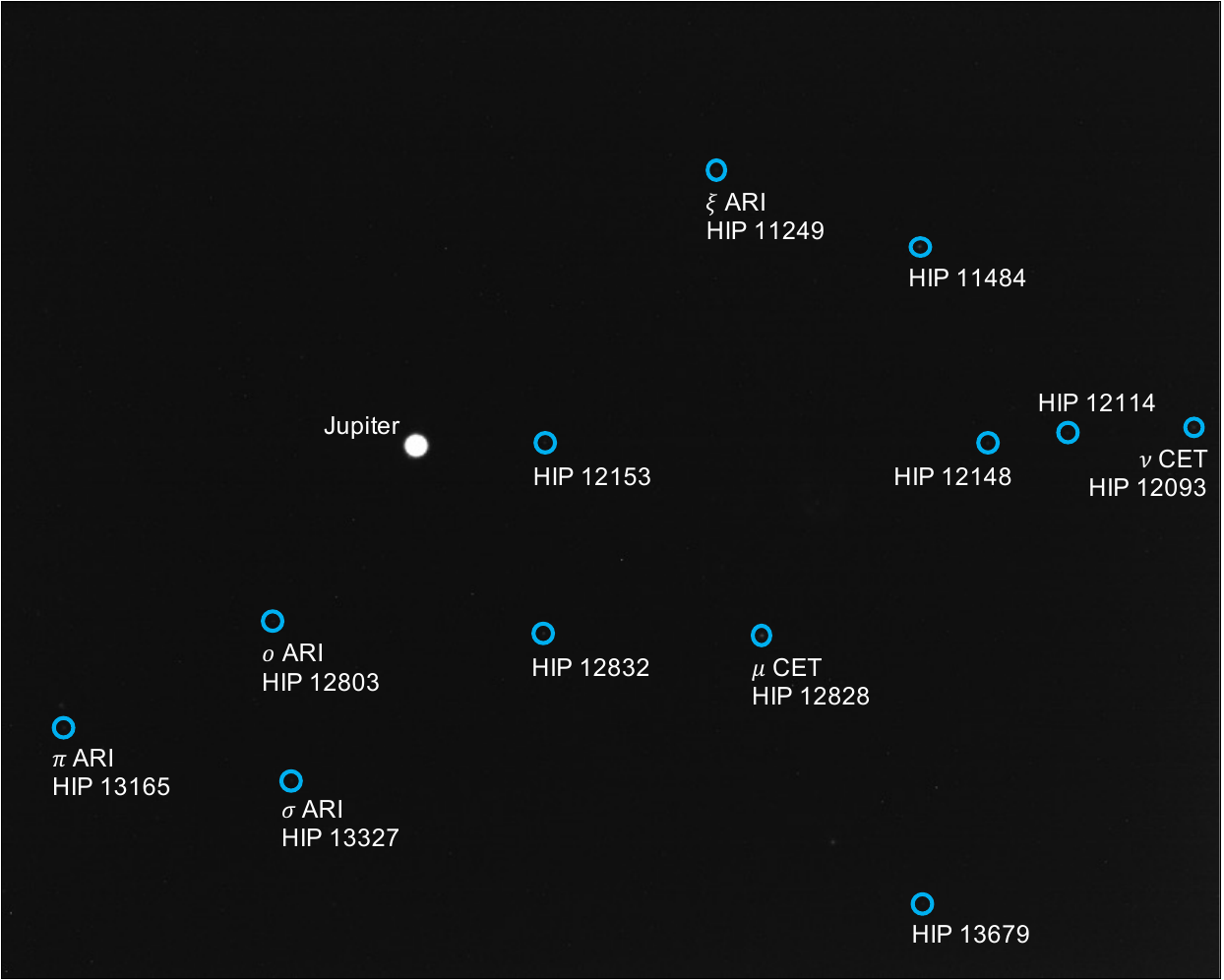}
    \caption{Image 590c, a long-exposure image showing a saturated Jupiter, alongside the 12 stars of magnitude 6 or brighter visible in the image. 
    }
    \label{fig:Starfield590c}
\end{figure}

The apparent magnitude of Jupiter as seen from Lunar Flashlight was about $m \approx -2.85$ during days 80-120 of the LONEStar imaging campaign. During this same period of time, Jupiter appeared to be moving across a rather dim portion of the constellation Aries and adjacent to a bright portion of the constellation Cetus (see Fig.~\ref{fig:JupiterRADEC}). The brightest star within any of the Jupiter OPNAV images (HIP 14135, $\alpha$ CET) had a visual magnitude of $m$ = 2.54. Thus, the magnitude separation between the brightest star and Jupiter is $\Delta m \approx 2.54 - (-2.85) \approx 5.39$. Recognizing that the effective dynamic range of the sensor is only $\Delta m \approx 5$, this difference of $\Delta m = 5.39$ is just beyond the limits of the LONEStar pipeline and the system cannot detect a star without also saturating Jupiter. Consequently, Jupiter observations required separate images for Jupiter and for stars, such as shown in Fig.~\ref{fig:Jupiter593ab}. The star field images routinely matched about 7-12 stars to the Hipparcos database (e.g., Fig.~\ref{fig:Starfield590c}).

\begin{figure}[p]
    \centering
    \includegraphics[width=1\linewidth]{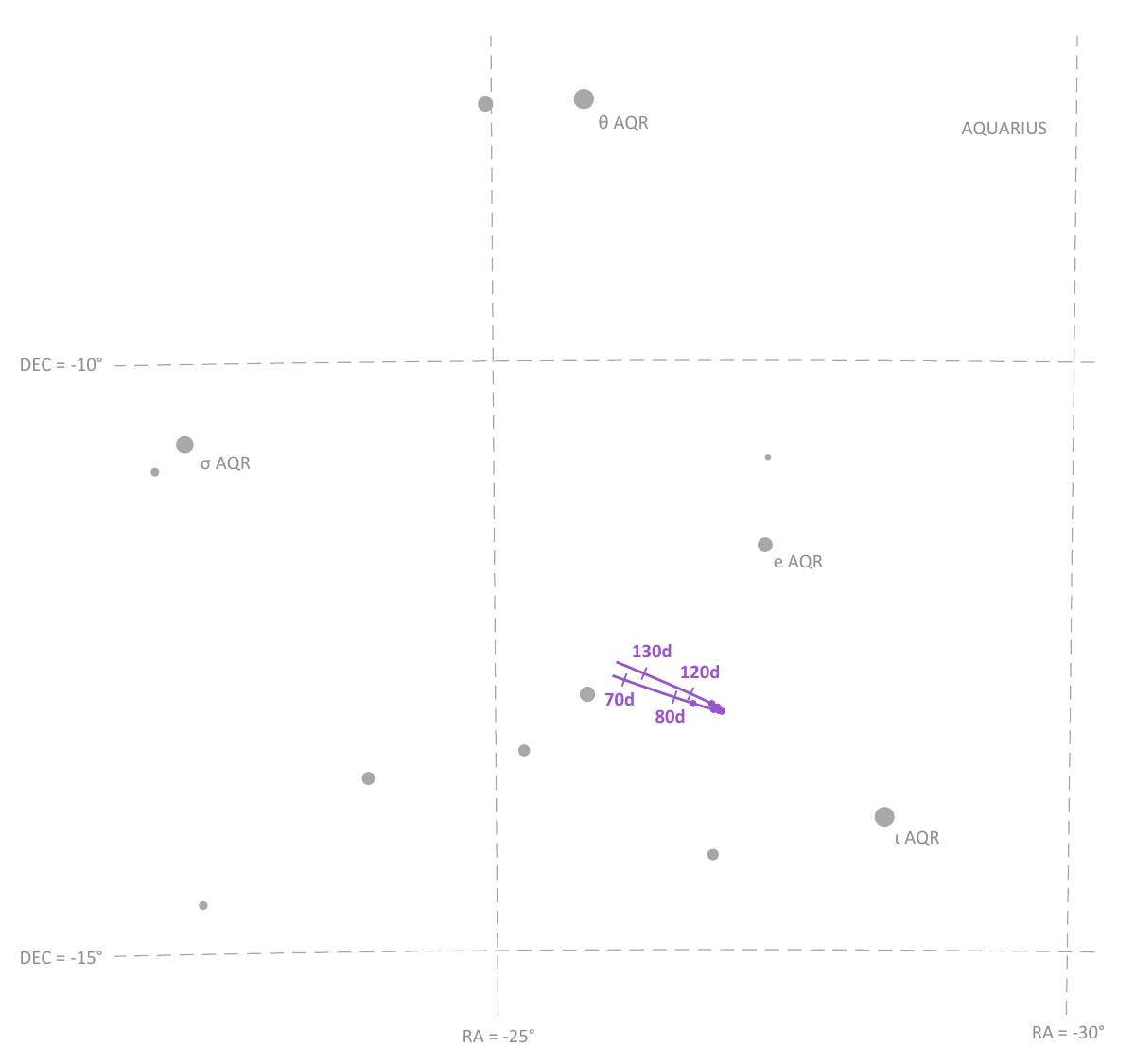}
    \caption{Apparent motion of Saturn (purple) on the celestial sphere as seen by Lunar Flashlight. The measured LOS directions to Saturn obtained from LONEStar OPNAV images are shown as dots. Tick marks along Saturn's track indicate elapsed time from the LONEStar reference epoch.}
    \label{fig:SaturnRADEC}
\end{figure}

\begin{figure}[p]
    \centering
    \includegraphics[width=1\linewidth]{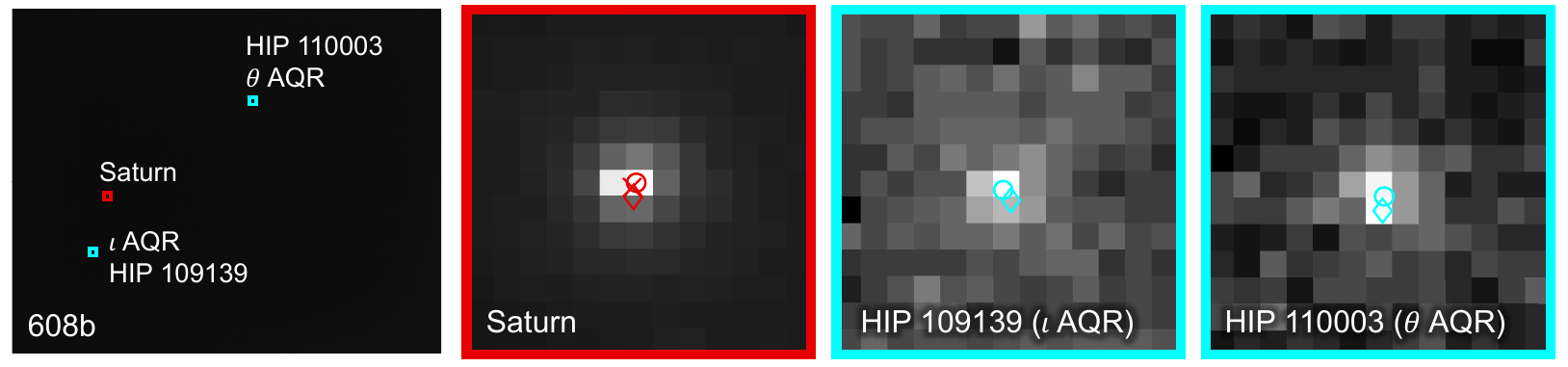}
    \caption{Example of a short-exposure image of Saturn, demonstrating that stars and Saturn are simultaneously observable. The $\circ$ indicates a measured centroid, $\diamond$ indicates a centroid projection from the JPL-produced navigation solution, and $\times$ indicates the reprojection from the OPNAV-produced solution. Each zoomed patch has been renormalized such that the darkest pixel is black and the brightest is white.}
    \label{fig:SaturnStar}
\end{figure}

Saturn appeared substantially dimmer than Jupiter. The apparent magnitude of Saturn as seen from Lunar Flashlight was about $m$ = 0.64-0.82 during days 80-120 of the LONEStar imaging campaign. Saturn appeared to transit across a very small portion of the constellation Aquarius (see Fig.~\ref{fig:SaturnRADEC}), near a number of modestly bright stars (notably $\theta$ AQR at $m$ = 4.17 and $\iota$ AQR at $m$ = 4.29). The difference in apparent visual magnitude between Saturn and these bright stars is $\Delta m \gtrsim 3.35$, which is easily detectable within the effective dynamic range of the LONEStar system ($\Delta m \approx 5$). Consequently, it is usually possible to see two or three stars in the background of most of the Saturn images (e.g., Fig.~\ref{fig:SaturnStar}). The small number of background stars, however, led to relatively poor attitude determination performance. Indeed, experimentation demonstrated that superior image alignment could be achieved by computing the attitude from long exposure stars images (which typically contained 7-12 matched stars) using the same procedure as for Jupiter. 

As can be seen from the observed paths of Jupiter and Saturn across the celestial sphere (Fig.~\ref{fig:JupiterRADEC} and Fig.~\ref{fig:SaturnRADEC}), the apparent angle between them was about 70 deg (see Table~\ref{tab:JupSatAngSep}). This angle was measured in practice by first imaging Jupiter (with star field images), and then slewing the spacecraft to subsequently image Saturn (with star field images) within the same Image Block. The time between sequential Jupiter and Saturn images was about 135 seconds for the six observation pairs summarized in Table~\ref{tab:JupSatAngSep}.

\begin{figure}[b!]
    \centering
    \includegraphics[width=1\linewidth]{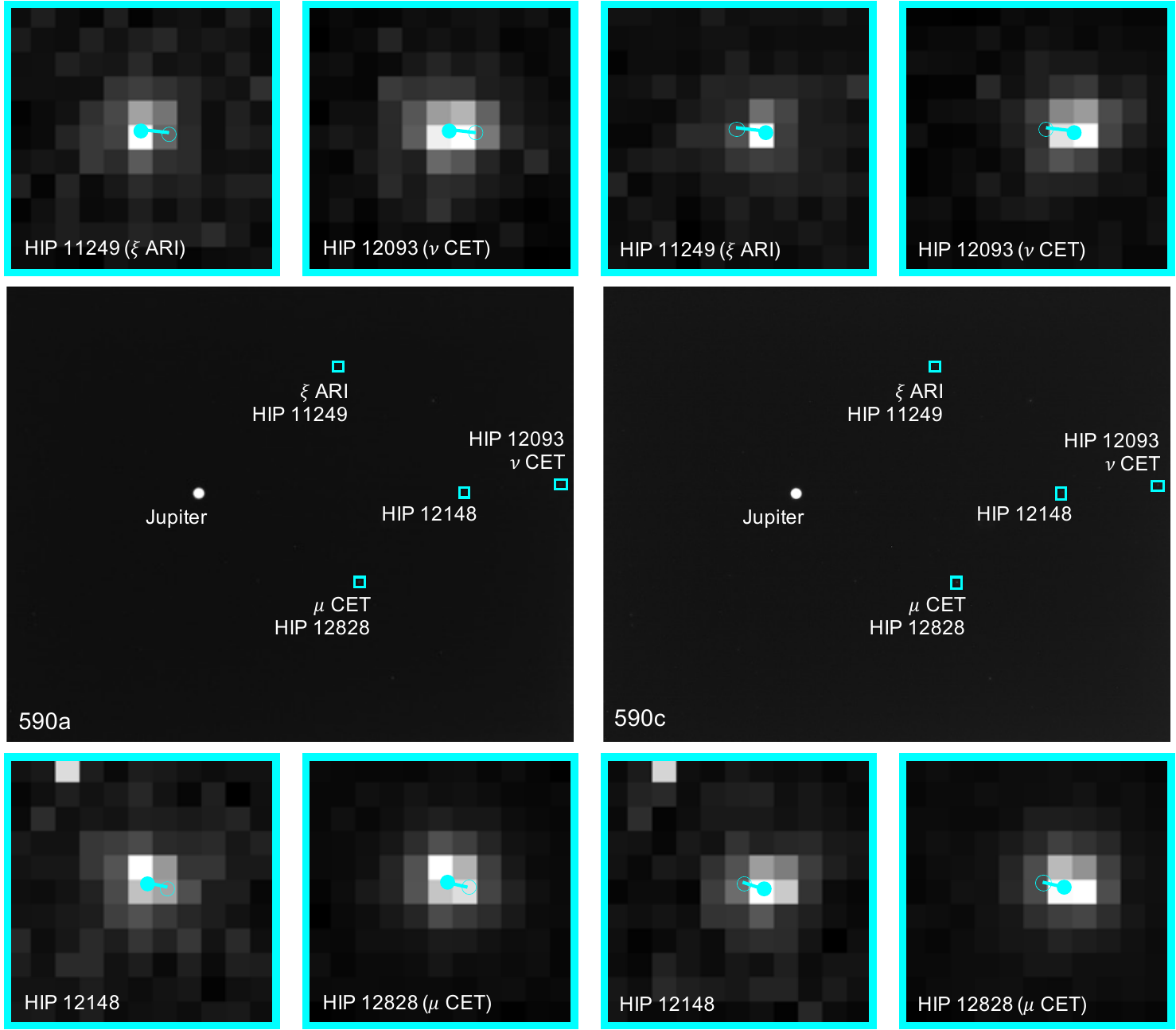}
    \caption{Two long-exposure images of Jupiter used to bracket the attitude of short-exposure OPNAV image 590b. The filled circle indicates the centroid observed in the corresponding image, with the unfilled circle indicating the centroid location in the other bracketing image. Approximately one pixel of attitude drift is observed between the image on the left (590a) and the image on the right (590c).}
    \label{fig:JupiterStarBracket}
\end{figure}

Within a particular Image Block, since the planet observations are separated by a slewing maneuver, only one of Jupiter or Saturn could be bracketed on both sides with star field images; the other planet only had a star field image on one side. Bracketing was usually performed with Jupiter since the Saturn images contained a few background stars. In one case (608b, 607b), one of the star field images could not be downlinked and thus attitude bracketing was not possible with either planet. 

The time between a short-exposure planet image and its corresponding long-exposure star image was about 5 seconds. Consequently, the time between two long-exposure star images bracketing a short-exposure planet image was about 10 seconds. The change in attitude across a bracket may be used to study the pointing stability during the LONEStar experiment, which is especially important for understanding the OPNAV pointing performance when bracketing with two star images is not possible (i.e., for one-sided attitude determination). As can be seen in Table~\ref{tab:JupSatAngSep}, the attitude change between the star bracket images was about 30-45 arcsec over the 10-second interval, which is equivalent to about one pixel (the IFOV is about 36.6 arcsec, see Table~\ref{tab:CameraSpecs}). This suggests an attitude rate on the order of about 3.0-4.5 arcsec/sec during OPNAV operations. The apparent movement of stars (by about one pixel) between two star bracket images for Jupiter may be seen in Fig.~\ref{fig:JupiterStarBracket}. By interpolating the attitude at the midpoint, the attitude error associated with the bracketed planet image is substantially reduced.

Results of triangulation using the midpoint algorithm with the measured LOS directions to Jupiter (red dots in Fig.~\ref{fig:JupiterRADEC}) and Saturn (purple dots in Fig.~\ref{fig:SaturnRADEC}) may be found in Table~\ref{tab:JupSatCamRes}. The covariance of the midpoint algorithm is analytically computed using the methods of Ref.~\cite{Henry:2023a}, and the total error $\sqrt{Tr(\bP)}$ and Mahalanobis distance are reported in Table~\ref{tab:JupSatErrStats}. The analytic approximation of total error from Eq.~\eqref{eq:JPLcovariance} and Ref.~\cite{Broschart:2019} could not be used here since different centroiding errors were assumed for Jupiter (0.5 px) and Saturn (0.25 px). In addition to planet centroiding errors, these Mahalanobis distance numbers assume a pointing error of about 0.25 px.

\begin{table}[h!]
    \centering
    \caption{Apparent angular separation between Jupiter and Saturn for the sequential triangulation experiment.}
    \begin{tabularx}{\textwidth}{lP{1.5 cm}YP{1.5 cm}P{1.5 cm}P{1 cm}P{1 cm}}
    \toprule
    \multirow{4}{*}{Image} & \multirow{4}{1.5 cm}{\centering LONEStar Elapsed Time [days]}  & \multicolumn{3}{c}{\multirow{2}{*}{Apparent Angle Between Jupiter and Saturn}} & \multicolumn{2}{c}{\multirow{2}{2.5 cm}{\centering Bracketing Star Angle [arcsec]}}\\ 
    \\
    \cmidrule(lr){3-5}
    \cmidrule(lr){6-7}
     &  & DSN + DE440 [deg] & \multirow{2}{1.5 cm}{\centering Measured [deg]} &  \multirow{2}{1.5 cm}{\centering Residual [arcsec]} & \multirow{2}{*}{Jupiter} & \multirow{2}{*}{Saturn}  \\
    \midrule
    587b, 588b & 92.545  & 70.77439 & 70.77837 & 14.328 & 38.838 & -      \\
    589b, 590b & 92.628  & 70.76461 & 70.76679 & 7.848  & 32.860 & -      \\
    593b, 594b & 95.713  & 70.39001 & 70.39410  & 14.724 & -      & 43.158 \\
    597b, 598b & 98.253  & 70.06330  & 70.06684 & 12.744 & 37.034 & -      \\
    604a, 605b & 109.711 & 68.44419 & 68.45058 & 23.004 & 36.057 & -      \\
    608b, 607b & 112.546 & 68.01979 & 68.01767 & 7.632  & -      & -     \\
    \bottomrule
    \end{tabularx}
    \label{tab:JupSatAngSep}
\end{table}

\begin{table}[h!]
    \centering
    \caption{Sequential triangulation residuals using the midpoint algorithm and observations of Jupiter and Saturn.}
    \begin{tabularx}{\textwidth}{l P{1 cm} P{1 cm} P{1 cm} P{1 cm} P{1 cm} P{1.7 cm}P{1.7 cm}}
    \toprule
    \multirow{4}{*}{Image} & \multicolumn{3}{c}{Camera Frame Residual [km]}  & \multicolumn{4}{c}{Residual Norm}\\
    \cmidrule(lr){2-4} \cmidrule(lr){5-8}
     & \multirow{3}{*}{X} & \multirow{3}{*}{Y} & \multirow{3}{*}{Z} & \multirow{3}{*}{$10^5$ km} & \multirow{3}{1 cm}{\centering Earth Radii} & Normalized by range to Jupiter & \multirow{3}{1.7 cm}{\centering Normalized by range to Saturn} \\
    \midrule
    587b, 588b & 26,311  & -33,098 & -22,046 & 0.4768 & 7.476  & 0.0000345 & 0.0000792 \\
589b, 590b & 14,060  & -14,693 & -10,318 & 0.2280 & 3.575  & 0.0000165 & 0.0000379 \\
593b, 594b & 29,931  & -54,013 & -15,875 & 0.6376 & 9.997  & 0.0000459 & 0.0001060 \\
597b, 598b & 54,019  & 32,895  & 11,759  & 0.6433 & 10.086 & 0.0000461 & 0.0001070 \\
604a, 605b & 94,101  & 11,901  & 59,125  & 1.1177 & 17.524 & 0.0000785 & 0.0001851 \\
608b, 607b & -57,315 & -70,230 & -17,264 & 0.9228 & 14.468 & 0.0000645 & 0.0001524 \\
    \bottomrule
    \end{tabularx}
    \label{tab:JupSatCamRes}
\end{table}

\begin{table}[h!]
    \centering
    \caption{Total error and Mahalanobis distance for each midpoint triangulation result.}
    \begin{tabularx}{0.75\textwidth}{lYYY}
    \toprule
    \multirow{2}{*}{Image} & Residual Norm [$10^5$ km] & \multirow{2}{*}{$\sqrt{Tr(\bP)}$ [km]} & Mahalanobis Distance [-] \\
    \midrule
    587b, 588b & 0.4768 & 123,326 & 0.830 \\
    589b, 590b & 0.2280 & 123,366 & 0.394 \\
    593b, 594b & 0.6376 & 123,920 & 1.115 \\
    597b, 598b & 0.6433 & 124,450 & 0.667 \\
    604a, 605b & 1.1177 & 127,267 & 1.396 \\
    608b, 607b & 0.9228 & 128,237 & 1.080 \\
    \bottomrule
    \end{tabularx}
    \label{tab:JupSatErrStats}
\end{table}

\clearpage
\subsection{Dynamic Triangulation: Jupiter and Saturn}
\label{Sec:DynamicTri}
With only a single camera having a narrow FOV, instantaneous triangulation by the simultaneous observation of two (or more) planets is an infrequent occurrence. This constraint may be relaxed by imaging two planets in rapid succession and triangulating with the midpoint algorithm, as was done in Section~\ref{Sec:SequentialTriJupSat} with Jupiter and Saturn. In many cases, however, it is operationally inconvenient to sequentially observe two (or more) planets in rapid succession. The concept of dynamic triangulation allows us to have a rather large amount of time between sequential observations, thus substantially simplifying OPNAV operations. Dynamic triangulation accomplishes this through a two-step process: (1) generating an initial guess with a simplified dynamical model and (2) refining this guess using the full dynamical model. 

The first step in the process is generating the initial guess, which follows a similar procedure as Ref.~\cite{Henry:2023a}. The principal concern here is that the initial guess is good enough that the refinement process in step 2 converges to the correct state. For short timespans (relative to the dynamics of the orbit), it is often convenient and practical to simply assume rectilinear motion to construct the initial guess, as was done in these LONEStar experiments. The generic framework for the first step of dynamic triangulation follows the framework of Ref.~\cite{Henry:2023a}. 

To generate the initial guess, first assume a linearized dynamical model which permits a solution of the form
\begin{equation}
    \begin{bmatrix}
        \br(t) \\
        \bv(t) 
    \end{bmatrix} = 
    \begin{bmatrix}
        \bPhi_{rr} & \bPhi_{rv} \\
        \bPhi_{vr} & \bPhi_{vv}
    \end{bmatrix}
    \begin{bmatrix}
        \br_0 \\
        \bv_0 
    \end{bmatrix} 
\end{equation}
where $\bPhi_{(\cdot)}$ are $3 \times 3$ submatrices of the full $6 \times 6$ state transition matrix (STM). To obtain only the position at some time $t_i$ for use in triangulation, the top three rows of this expression may be compactly written as 
\begin{equation}
    \br(t) = 
    \bPhi_{r}
    \begin{bmatrix}
        \br_0 \\
        \bv_0 
    \end{bmatrix}       
\end{equation}
where $\bPhi_r = [\bPhi_{rr} \  \bPhi_{rv}]$. This may be substituted in to the DLT (see Eq.~\eqref{eq:LOSTLinSys}) to obtain a linear system as a function at the arbitrarily chosen reference time
\begin{equation}
    \label{eq:DynamicTriLinSys}
    \begin{bmatrix}
        [\bell_1 \times] \bPhi_r (t_1,t_0) \\
        \vdots \\
        [\bell_n \times] \bPhi_r (t_n,t_0) 
    \end{bmatrix}
    \begin{bmatrix}
        \br_0 \\
        \bv_0
    \end{bmatrix} = 
    \begin{bmatrix}
        [\bell_1 \times] \bp_1 \\
        \vdots \\
        [\bell_n \times] \bp_n
    \end{bmatrix}
\end{equation}
Thus, the initial dynamic triangulation problem estimates the spacecraft's full translational state (both position and velocity) at the reference time, rather than only the position. 

This initial guess is then refined by minimizing the reprojection error with a nonlinear least-squares solver (e.g., LMA) and full dynamical model. This refinement follows the same batch estimation (i.e., orbit determination) procedure as discussed in Section~\ref{Sec:BatchFilter}. 

To demonstrate this approach, a subset of four measurements from Section~\ref{Sec:SequentialTriJupSat} were chosen---two of Jupiter (images 588b and 598b) and two of Saturn (images 584b and 594b)---with at least 2.5 days between each sequential measurement. This sequence of measurements spans a total of about 13 days, which is short enough (for heliocentric orbits) to assume rectilinear motion when generating the initial guess with Eq.~\eqref{eq:DynamicTriLinSys}. The initial state estimate constructed in this way yields the state vector residuals shown in the top rows of Tables~\ref{tab:DynTriPosResids} and \ref{tab:DynTriVelResids}, with a position residual of 774,503 km (121.43 Earth radii) and a velocity residual of 3,452 m/s. 

\begin{figure}[t!]
    \centering
    \includegraphics[width=0.9\linewidth]{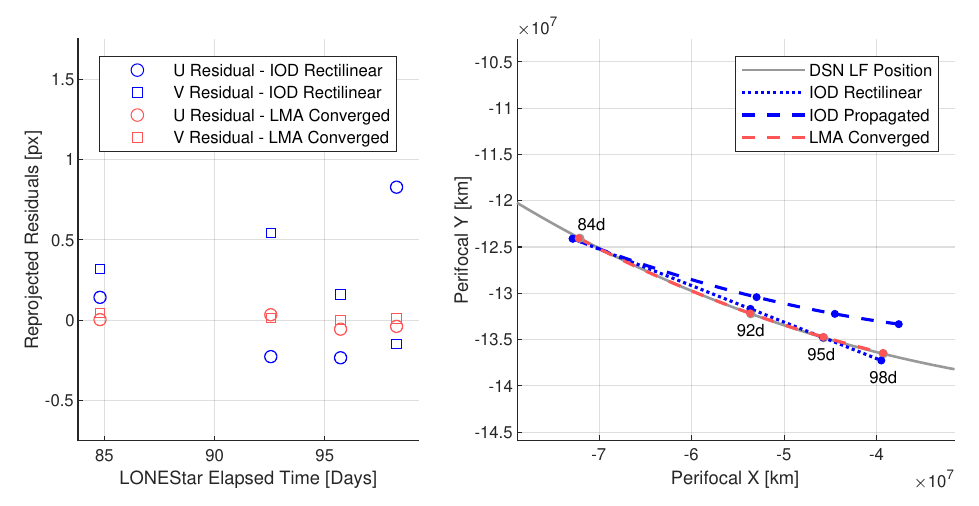}
    \caption{By minimizing the reprojection residuals (left), the dynamic triangulation IOD solution converges to the DSN trajectory (right). The axes correspond with LF’s instantaneous perifocal frame at the time of the first measurement.}
    \label{fig:DynamicTri}
\end{figure}

Minimizing the reprojection error (step 2) reduces the measurement residuals (left frame of Fig.~\ref{fig:DynamicTri}) and state residuals (Table~\ref{tab:DynTriPosResids} and Table~\ref{tab:DynTriVelResids}) by an order of magnitude, yielding a final state estimate with a position residual of 137,674 km (21.59 Earth radii) and a velocity residual of 132 m/s. Fig.~\ref{fig:DynamicTri} illustrates how sequential steps within the dynamic triangulation process converge to a result that is in excellent agreement with the reference DSN trajectory. With this data, it was shown that dynamic triangulation can use infrequent LOS measurements to perform celestial triangulation for the purposes of IOD.

\begin{table}[h!t]
    \centering
    \caption{Summary of the dynamic triangulation estimate and the converged solution position residuals at 2023-OCT-14 19:04:06 UTC.}
    \begin{tabularx}{\textwidth}{p{1.7 cm} P{1 cm} P{1.1 cm} P{1.1 cm} P{0.8 cm} P{0.8 cm} P{1.6 cm}P{1.6 cm}}
    \toprule
     & \multicolumn{3}{c}{ICRF Residual [km]}  & \multicolumn{4}{c}{Residual Norm}\\
    \cmidrule(lr){2-4} \cmidrule(lr){5-8}
     & \multirow{3}{*}{X} & \multirow{3}{*}{Y} & \multirow{3}{*}{Z} & \multirow{3}{*}{$10^5$ km} & \multirow{3}{0.8 cm}{\centering Earth Radii} & Normalized by range to Jupiter & \multirow{3}{1.6 cm}{\centering Normalized by range to Saturn} \\
    \midrule
    Step 1: IOD Rectangular & \multirow{2}{*}{642,194} & \multirow{2}{*}{-371,941} & \multirow{2}{*}{-221,589} & \multirow{2}{*}{7.745} & \multirow{2}{*}{121.43} & \multirow{2}{*}{0.0012781} & \multirow{2}{*}{0.0005670} \\
    Step 2: LMA Converged & \multirow{2}{*}{107,066} & \multirow{2}{*}{-86,482}  & \multirow{2}{*}{-3,406} & \multirow{2}{*}{1.377} & \multirow{2}{*}{21.59}  & \multirow{2}{*}{0.0002272} & \multirow{2}{*}{0.0001008} \\
    \bottomrule
    \end{tabularx}
    \label{tab:DynTriPosResids}
\end{table}

\begin{table}[h!t]
    \centering
    \caption{Summary of the dynamic triangulation estimate and the converged solution velocity residuals at 2023-OCT-14 19:04:06 UTC.}
    \begin{tabularx}{0.75\textwidth}{p{1.7 cm} YYY P{2 cm} }
    \toprule
     & \multicolumn{3}{c}{ICRF Residual [km/s]} & \multirow{2.5}{2 cm}{\centering Residual Norm [km/s]}\\
    \cmidrule(lr){2-4} 
     & X & Y & Z & \\
    \midrule
    Step 1: IOD Rectangular & \multirow{2}{*}{-3.325} & \multirow{2}{*}{-0.885} & \multirow{2}{*}{-0.280} & \multirow{2}{*}{3.452} \\
    Step 2: LMA Converged  & \multirow{2}{*}{-0.125} & \multirow{2}{*}{0.012}  & \multirow{2}{*}{-0.042} & \multirow{2}{*}{0.132} \\
    \bottomrule
    \end{tabularx}
    \label{tab:DynTriVelResids}
\end{table}

It is important to remember that the residuals in Table~\ref{tab:DynTriPosResids} and Table~\ref{tab:DynTriVelResids} are only an IOD solution produced with four measurements over a relatively short period of time. The IOD result is not used to navigate directly, but only to generate the initial guess for a batch orbit determination that processes many measurements (e.g., more than four observations, more targets than just Jupiter and Saturn) over a longer period of time (e.g., longer than 13 days). The LONEStar batch orbit determination solution is discussed in Section~\ref{Sec:BatchFilter}.

\section{Optical Navigation with the Earth and Moon}
\label{Sec:OPNAVEarthMoonTop}

\subsection{Earth and Moon Viewing Geometry}

\begin{figure}[b!]
    \centering
    \includegraphics[width=1\linewidth]{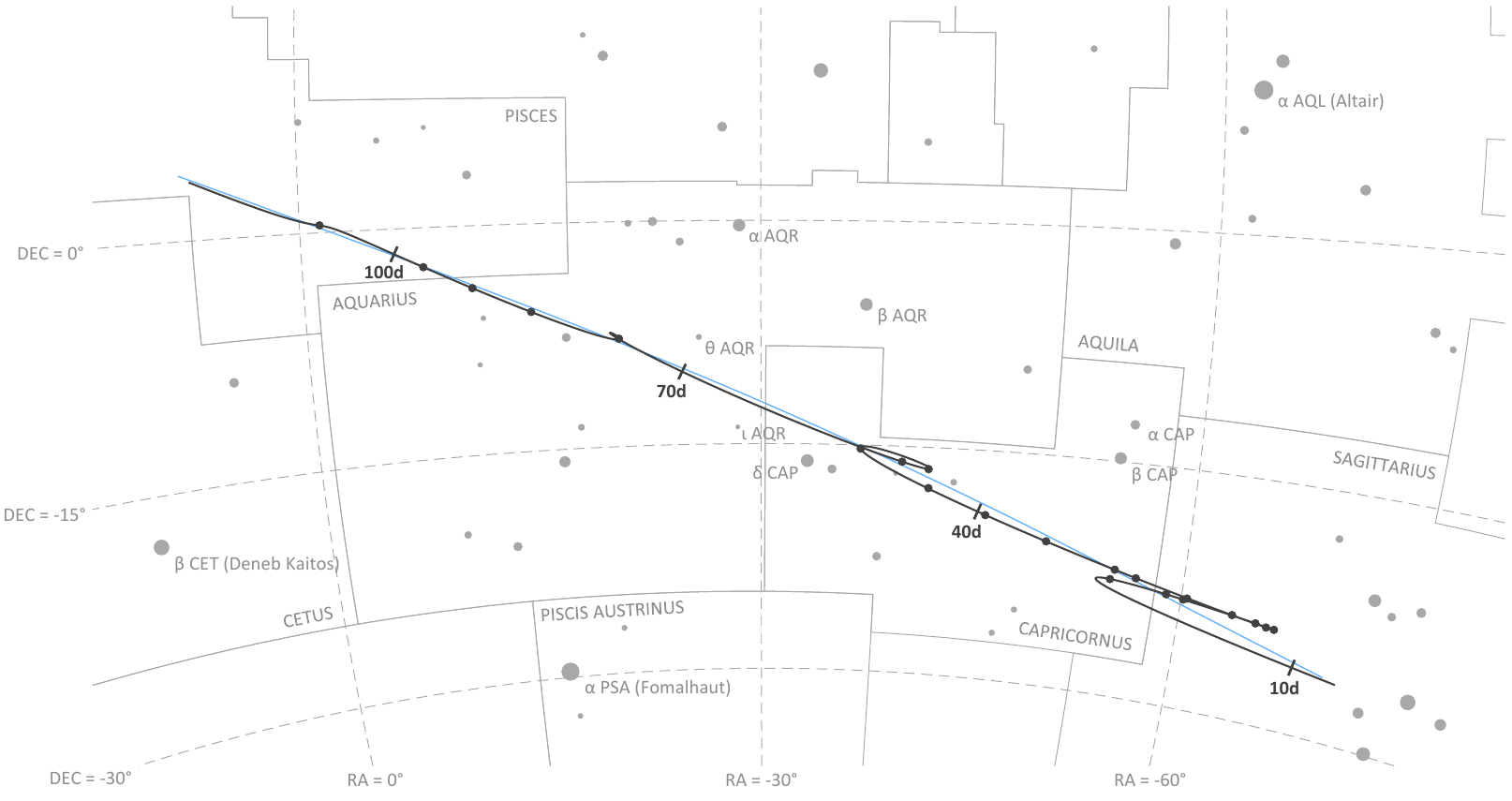}
    \caption{Apparent motion of the Moon (dark gray) on the celestial sphere as seen by Lunar Flashlight. The path of the Earth is shown in light blue for reference. The measured LOS directions to the Moon obtained from LONEStar OPNAV images are shown as dark gray dots. Tick marks along the Moon’s track indicate elapsed time from the LONEStar reference epoch.}
    \label{fig:MoonRADEC}
\end{figure}

\begin{figure}[b!]
    \centering
    \includegraphics[width=1\linewidth]{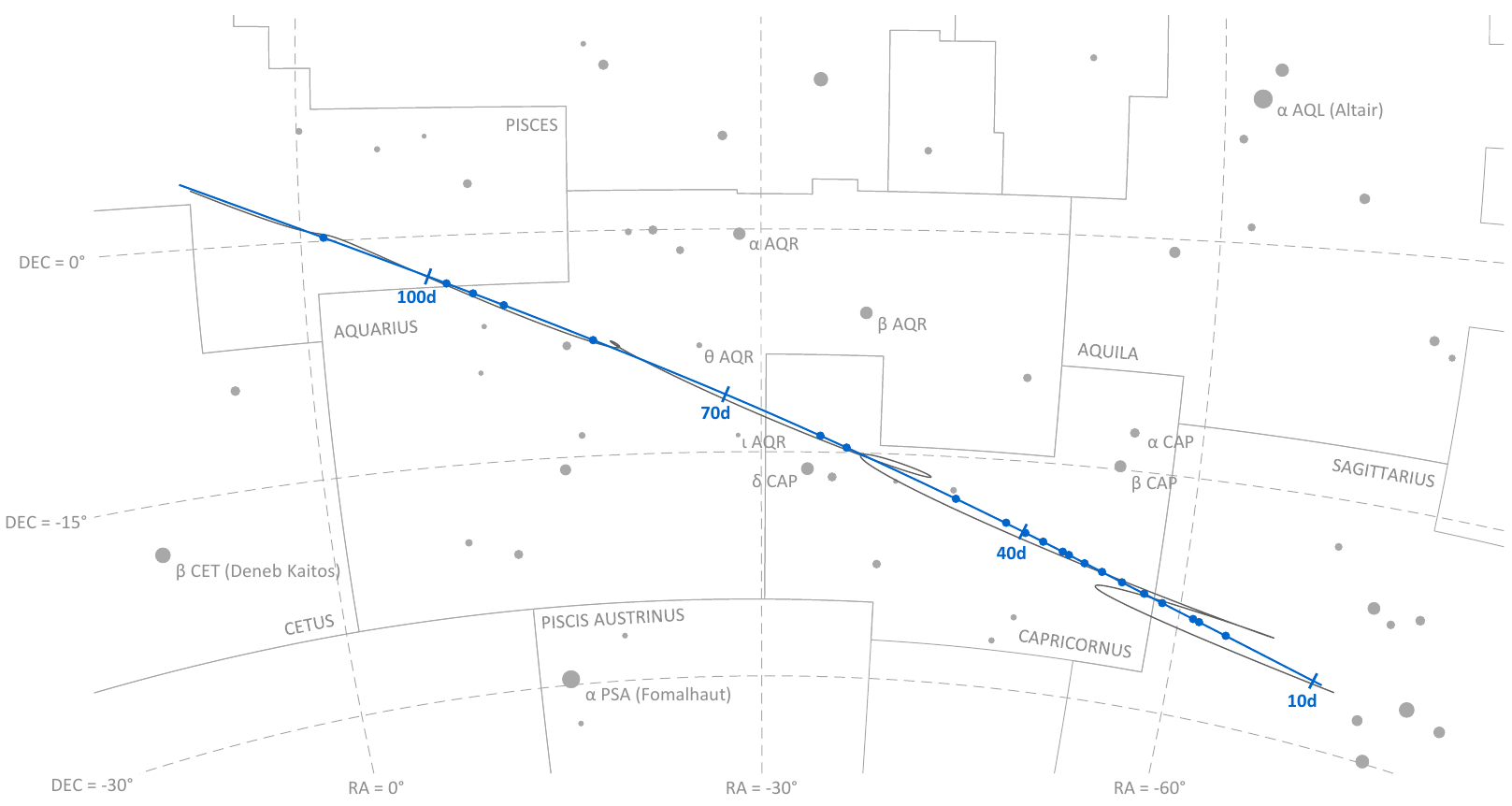}
    \caption{Apparent motion of the Earth (blue) on the celestial sphere as seen by Lunar Flashlight. The path of the Moon is shown in light gray for reference. The measured LOS directions to the Earth obtained from LONEStar OPNAV images are shown as blue dots. Tick marks along the Earth’s track indicate elapsed time from the LONEStar reference epoch.}
    \label{fig:EarthRADEC}
\end{figure}

Images of the Earth and Moon were collected throughout the entirety of the LONEStar imaging campaign (see Fig.~\ref{fig:ImagingTimeline} in Section~\ref{Sec:ImagingCampaign}). During this time, both the Earth and Moon were always visible in a single OPNAV image. For the duration of the approximately 100-day span of imaging, the Earth-Moon system appeared to transit the constellations Sagittarius, Capricornus, Aquarius, and Pisces, from the vantage point of LF (see Fig.~\ref{fig:MoonRADEC} and Fig.~\ref{fig:EarthRADEC}). Early in the LONEStar imaging campaign, the Moon’s orbit had an apparent angular diameter of 10.3 deg (bottom right of Fig.~\ref{fig:MoonRADEC}). However, as the distance between LF and the Earth-Moon system increased with time, the apparent angular diameter of the Moon's orbit dropped to only 2.5 deg (top left of Fig.~\ref{fig:MoonRADEC}).

\begin{figure}[t!]
    \centering
    \includegraphics[width=1\linewidth]{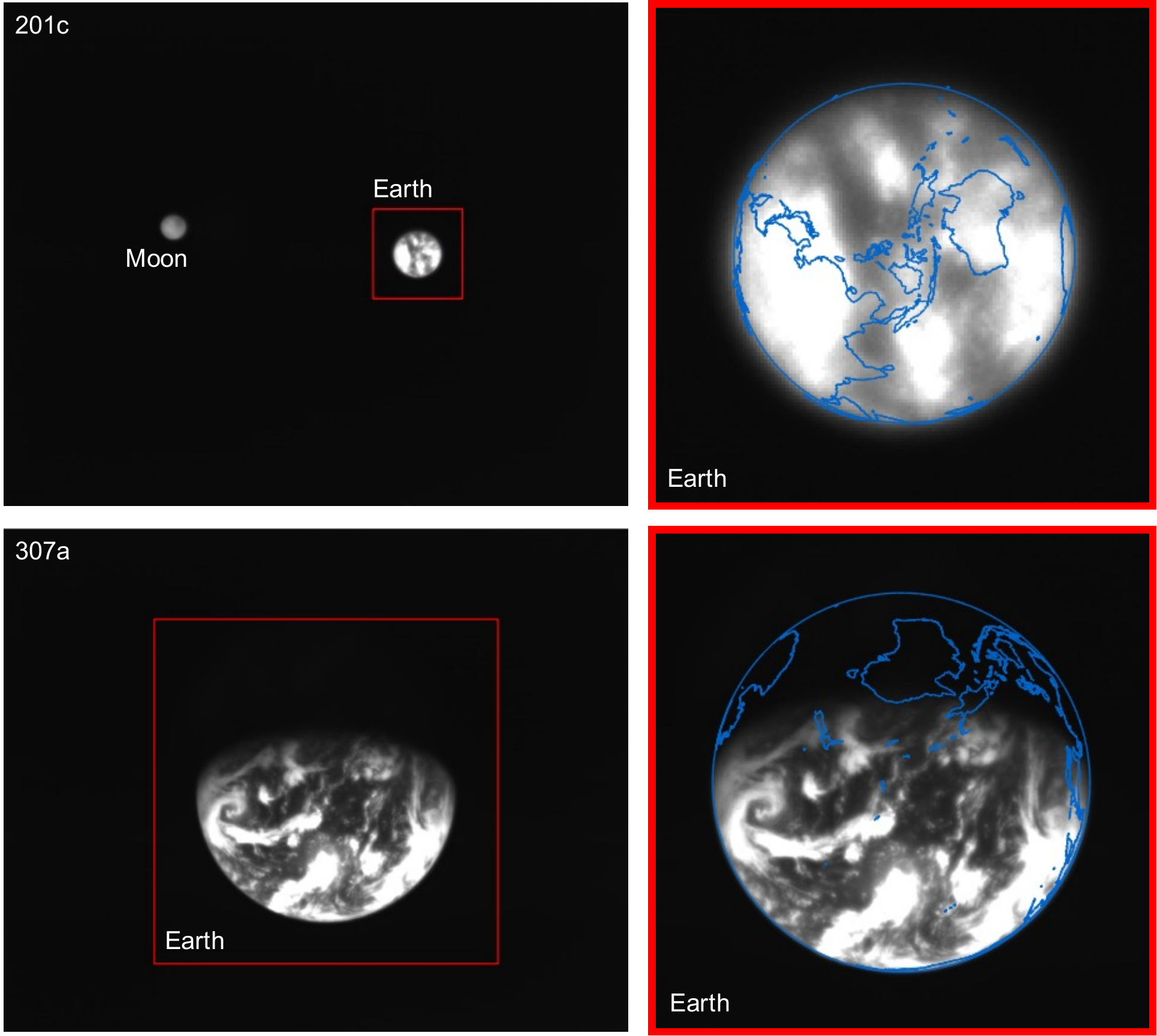}
    \caption{Close-range images of Earth captured prior to LONEStar Imaging Campaign with overlaid continental projections (blue). Image 201c (-121d) and 307a (-66d).}
    \label{fig:LargeEarth}
\end{figure}

\begin{figure}[t!]
    \centering
    \includegraphics[width=1\linewidth]{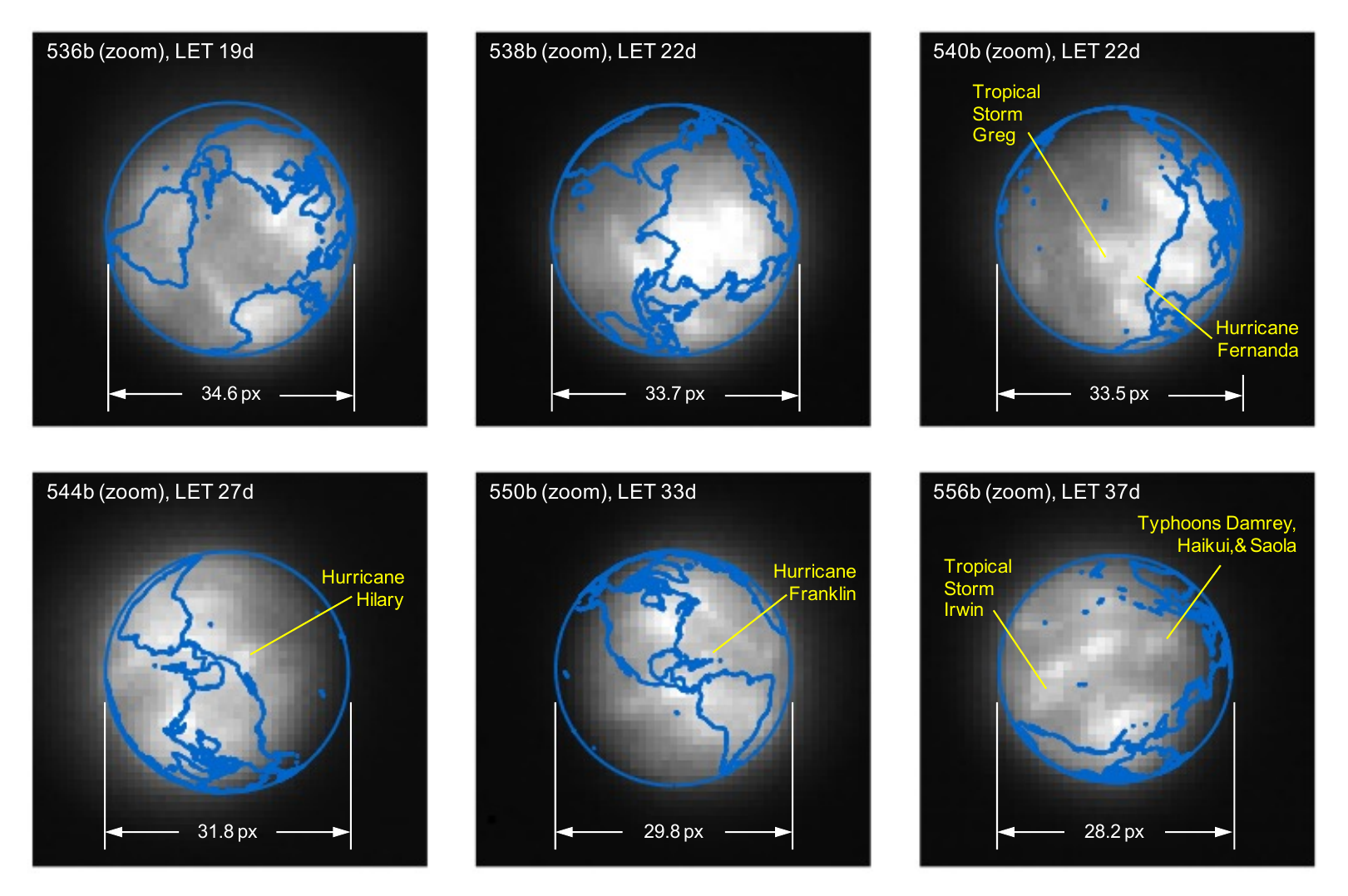}
    \caption{A series of Earth images captured during the LONEStar Imaging Campaign with overlaid continental projections (blue) and highlighted weather patterns such as hurricanes (yellow). When compared to NOAA data \cite{NOMADS:2006, GFSArchive:2015}, other prominent cloud formations are also easily recognizable.}
    \label{fig:EarthContinents}
\end{figure}

Before the beginning of the LONEStar imaging campaign, two images were acquired prior to LF’s ejection from the Earth-Moon system (see Fig.~\ref{fig:LargeEarth}). Disk-resolved imagery of the Earth remained available for much of the LONEStar imaging campaign. Although the intentional XACT defocusing does not permit crisp imagery of Earth, a number of prominent features remain visible. In particular, the bright regions of Earth have been directly correlated with prominent cloud patterns from NOAA data \cite{NOMADS:2006, GFSArchive:2015} at the specific image times. The LONEStar Earth images coincided with the northern hemisphere’s 2023 hurricane season and LF witnessed a number of named weather patterns from heliocentric space. A sampling of images with continent overlays (using the OPNAV-derived state estimates presented in subsequent sections) and callouts to known weather patterns may be found in Fig.~\ref{fig:EarthContinents}. 

\subsection{Moon Localization}
\label{Sec:MoonLocalization}

As the LF trajectory departed the Earth-Moon system (see Fig.~\ref{fig:LFtrajectorySmall}), the Moon transitioned from a resolved body to an unresolved body (subtending 10.4 pixels on 2023-AUG-10 to 2.2 pixels on 2023-NOV-08) during the LONEStar imaging campaign. Even when close, the Moon was too small for effective horizon-based OPNAV. Thus, both template matching and centroiding (with photocenter offset correction) were explored for generating lunar LOS measurements.

To centroid via template matching, 2-D renderings of the lunar surface were created to match the appearance of the Moon in the raw image. The Moon was rendered as a constant albedo sphere using the lunar-Lambert reflectance model~\cite{McEwen:1996}
\begin{equation}
    r(i, e, g) = \left[1 - \beta(g) \right] \cos i + \beta(g) \frac{2 \cos i}{\cos i + \cos e}
\end{equation}
where $i$ is the incidence angle, $e$ is the emission angle, and $g$ is the phase angle. Following the heuristic observations from Ref.~\cite{Gaskell:2008}, the phase-dependent blending of the Lambertian and Lommel-Seeliger models was assumed to be $\beta(g) = \text{exp}\left( -g / 60 \degree \right)$. The rendered sphere was then rotated into LF's camera frame such that orientation of the illuminated sphere was representative of the true image. 

\begin{figure}[b!]
    \centering
    \includegraphics[width=0.8\linewidth]{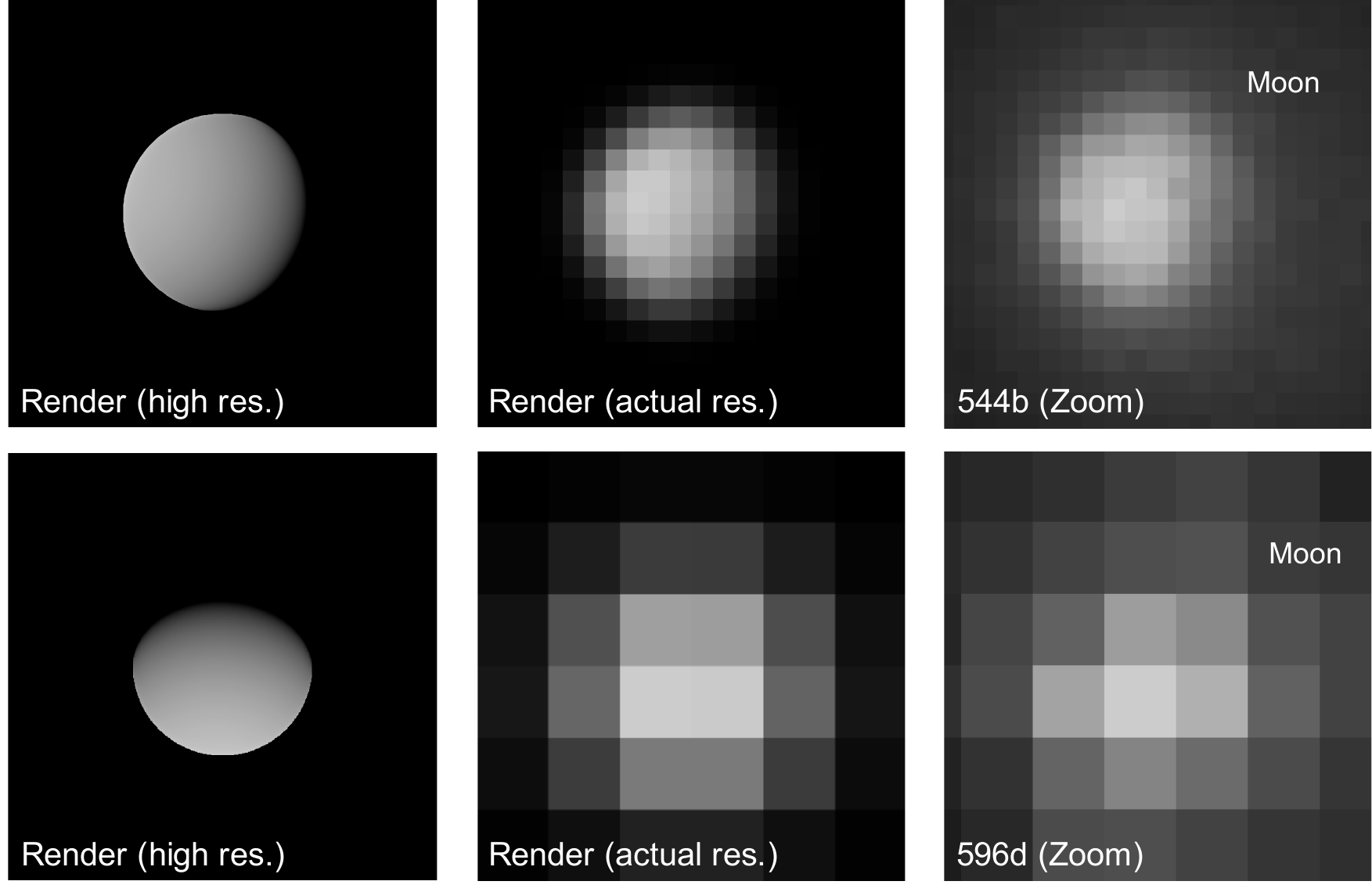}
    \caption{Template generation process for image 544b (top) and 596d (bottom), where the initial high-resolution render (left) is discretized and defocused (middle) to simulate the appearance of the moon in the raw image (right).}
    \label{fig:MoonTemplate}
\end{figure}

To construct a template at the appropriate scale, the lunar-Lambert disk was numerically integrated over the extent of each pixel. Determining the appropriate bounds for this discretization depends on an a priori estimate of the Moon’s diameter in the image, which was obtained from the DSN-based OD solution (though a functionally equivalent answer could come directly from image processing). The discretized (i.e., pixelated) disk was then defocused using the PSF from Section~\ref{Sec:PSF} to arrive at the template that was used for normalized cross-correlation. This process is illustrated in Fig.~\ref{fig:MoonTemplate}. 

\begin{figure}[b!]
    \centering
    \includegraphics[width=0.8\linewidth]{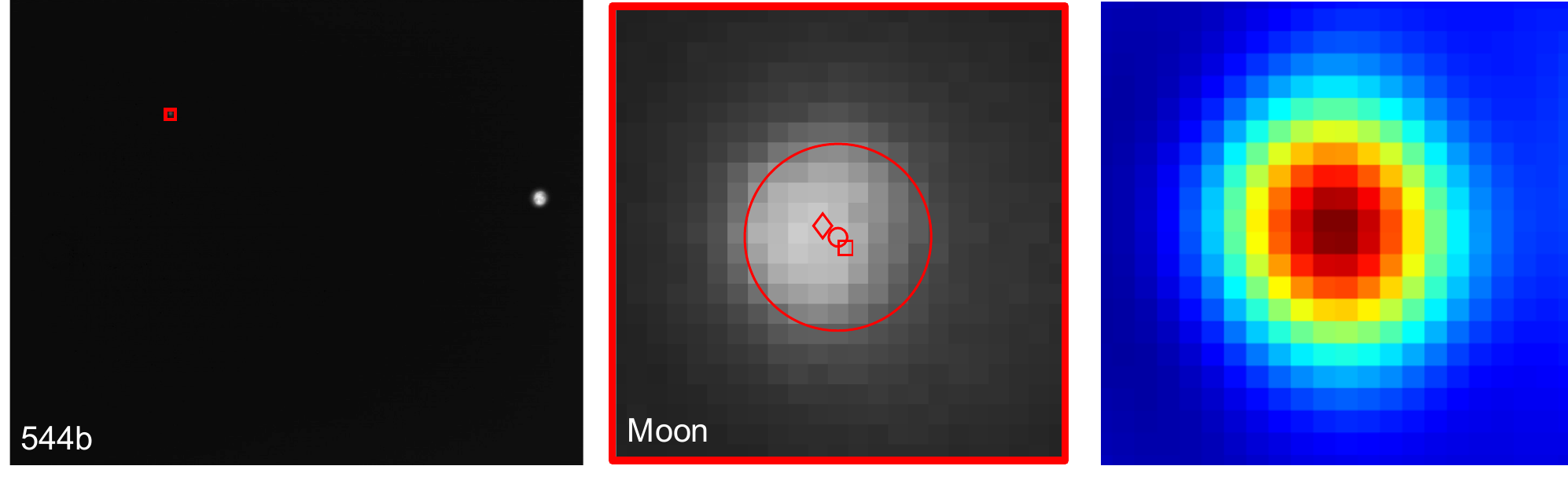}
    \caption{A zoom-in on the Moon (middle) within LF image 544b (left). The peak of the cross-correlation heatmap (right) corresponds to the $\circ$ marker in the middle image. As a comparison, the $\square$ marker shows the phase-corrected COB and the $\diamond$ marker is the reprojection of the Moon's center from the JPL-produced navigation solution. A reprojection of the Moon’s horizon is also shown. }
    \label{fig:MoonCorrPeak}
\end{figure}

Template matching is performed using normalized cross-correlation (NCC) \cite{Lewis:1995}. During the cross-correlation process, NCC normalizes the brightness of both the image and template, which circumvents the need to consider the absolute brightness of the Moon in the real image. The output of the NCC algorithm is an image-sized array of correlation scores indicating the agreement between a local image patch and the template.  A typical correlation peak is shown in Fig.~\ref{fig:MoonCorrPeak}. Subpixel template registration (i.e., Moon centroiding) is achieved by fitting a paraboloid to the correlation peak and finding the maximum.

\begin{figure}[h!]
    \centering
    \includegraphics[width=0.8\linewidth]{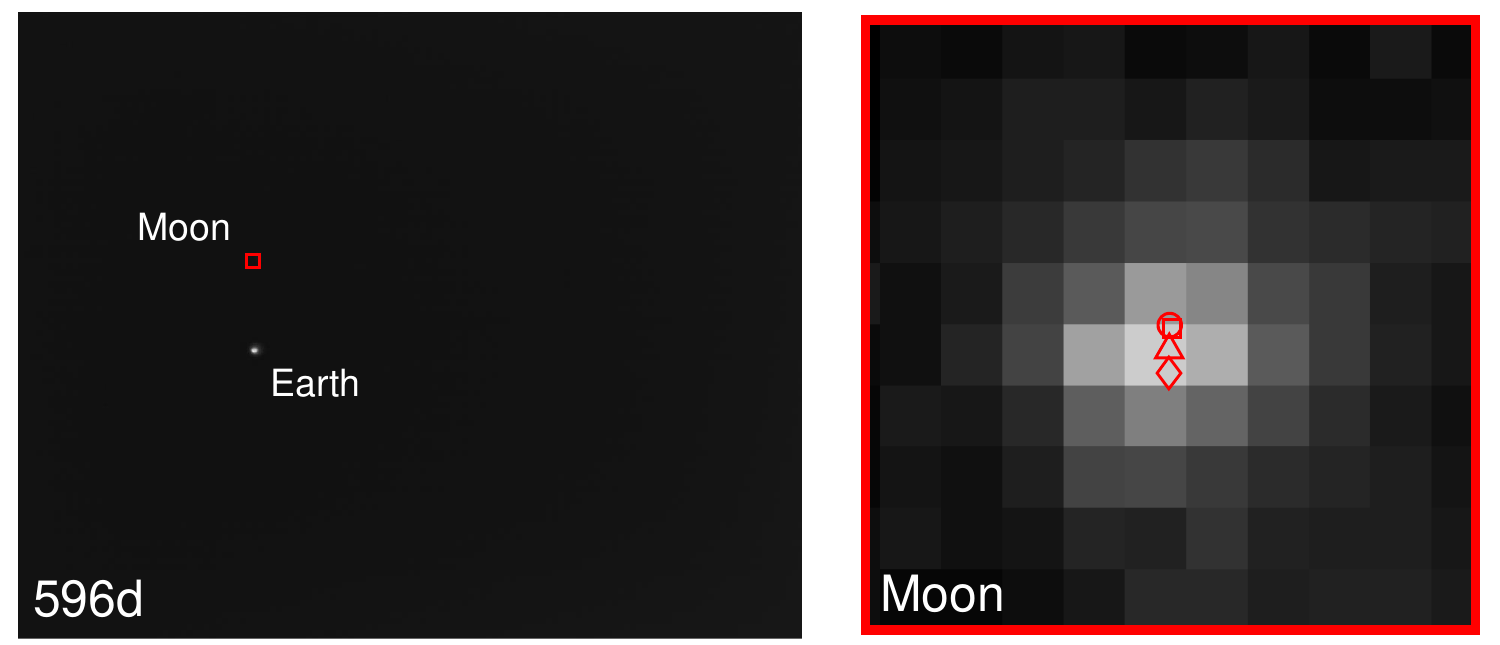}
    \caption{Comparison of different Moon centroiding algorithms (right) on LF image 596d (left). In the zoom-in on the right, the NCC centroid is the $\circ$ marker, the reprojection of the JPL-produced navigation solution is the $\diamond$ marker, uncorrected COB centroid is the $\triangle$ marker, and the photocenter corrected COB centroid is the $\square$ marker.}
    \label{fig:MoonCentroid}
\end{figure}

\begin{figure}[t!]
    \centering
    \includegraphics[width=0.8\linewidth]{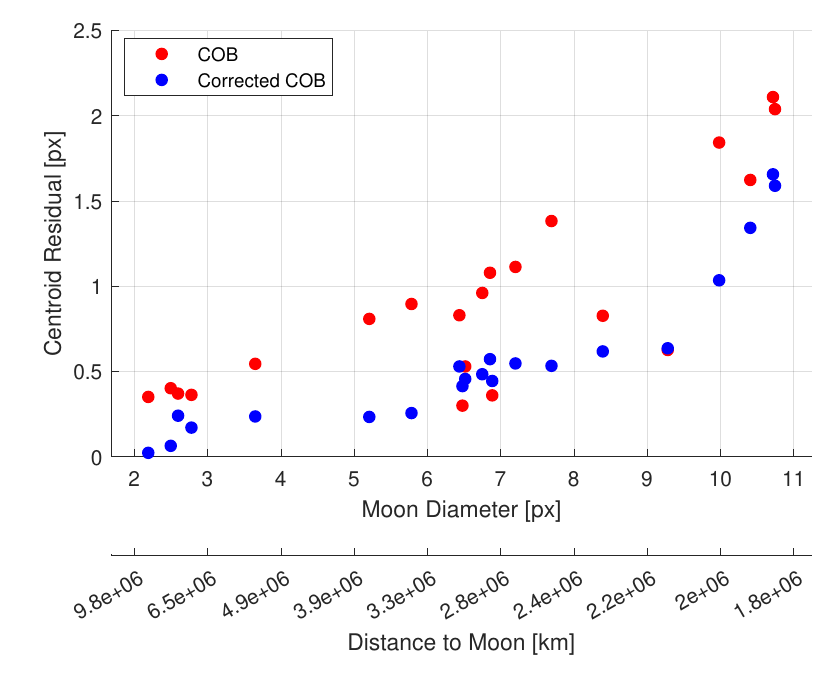}
    \caption{The pixel distance of the COB (red) and photocenter offset corrected COB (blue) from the NCC centroid as a function of the Moon’s diameter in pixels. The COB centroids with the photocenter offset correction have lower residuals than those without the correction. }
    \label{fig:MoonCentroidComp}
\end{figure}

An alternative to the NCC-based template matching method is center of brightness (COB) with a phase-dependent photocenter offset correction. Upon identifying a large contiguous cluster of bright pixels as the Moon (e.g., by proximity to the Moon’s expected location), the COB is computed in the same way as for unresolved objects (see Eq.~\eqref{eq:COB}). At non-zero phase angles, the COB is biased away from the geometric center in the direction of the Sun. As such, a photocenter offset correction may be applied in the direction opposing the Sun to compensate for this bias. Given the direction from the Sun in the image plane $\bu_{illum}$ (i.e., the $2 \times 1$ direction of incoming sunlight) and apparent radius of the Moon $R_M$ in pixels, the photocenter offset correction for a partially illuminated Lambertian sphere is given by \cite{Lindegren:1977,Kaasalainen:2004}
\begin{equation}
    \Delta \bu =  \left( R_M \frac{3 \pi}{16} \frac{1 + \cos g}{(\pi - g) \cot g + 1} \right) \bu_{illum}
\end{equation}
An example may be seen in Fig.~\ref{fig:MoonCentroid}. As can be seen in Fig.~\ref{fig:MoonCentroidComp}, the photocenter offset correction brings the raw COB into better agreement with the NCC centroids. 

Orbit determination was attempted with both the NCC centroids and offset corrected COB centroids, and NCC was found to consistently provide lower post-fit residuals. Thus, all subsequent triangulation and orbit determination results assume lunar LOS measurements produced by NCC with a lunar-Lambert template.

\subsection{Earth Localization}
\label{Sec:EarthLocalization}
OPNAV measurements of Earth were complicated by the body’s apparent size, optical effects, cloud cover, and atmosphere. Near the beginning of the LONEStar campaign (2023-AUG-10), the Earth subtended almost 35 pixels and dropped to a minimum of 11 pixels by the end of the mission (2023-NOV-08). Thus, Earth was processed as a resolved body for the duration of the experiment. Due to optical effects, there was often a noticeable coma that complicated the use of rudimentary image processing techniques. Additionally, template matching techniques (e.g., as was used for the Moon in Section~\ref{Sec:MoonLocalization}) were deemed inappropriate given the constantly evolving cloud cover which played a dominant role in the Earth’s appearance. For these reasons, Earth localization was accomplished via horizon-based OPNAV. LONEStar accomplished this with the Christian-Robinson algorithm (CRA)~\cite{Christian:2016b}, specifically using the contemporary formulation described in Ref.~\cite{Christian:2021}.

A significant challenge to using a horizon-based method on the Earth is the presence of its atmosphere, which is well-known to artificially shift the apparent location of the lit limb \cite{Christian:2016c}. Such challenges related to Earth-based OPNAV were also studied during Artemis I \cite{Inman:2024}. However, given the intentional defocusing of the LF camera and small apparent diameter of Earth, the contribution of atmospheric effects to limb localization error is comparatively small. 

\begin{figure}[b!]
    \centering
    \includegraphics[width=1\linewidth]{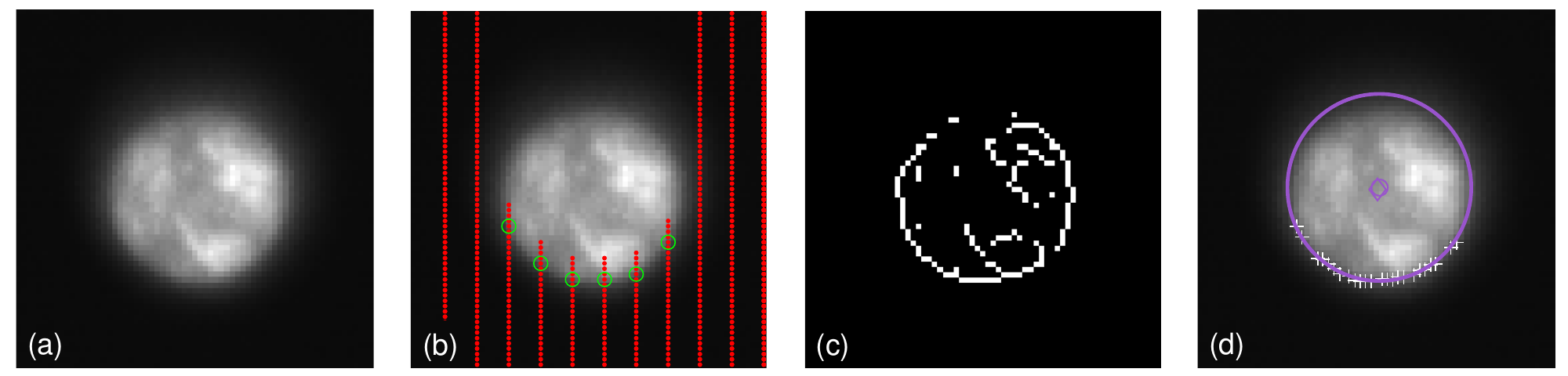}
    \caption{Earth localization pipeline. From left to right: a) zoomed view of the Earth in frame, b) scanlines in illumination direction to create a coarse mask of edge points, c) Sobel edges (unfiltered), d) the resultant subpixel edge estimates (white +), the CRA centroid ($\circ$), the reprojection of the JPL-produced navigation solution ($\diamond$), and the overlaid reprojected ellipsoid.}
    \label{fig:HorizonPipeline}
\end{figure}

\begin{figure}[t!]
    \centering
    \includegraphics[width=0.6\linewidth]{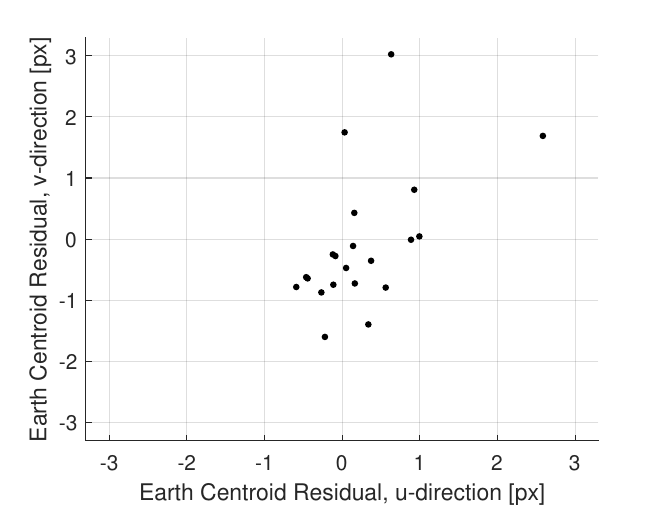}
    \caption{Residuals of Earth centroid estimates, comparing the CRA results against the DSN reprojected centroids.}
    \label{fig:EarthCRAResiduals}
\end{figure}

A straightforward pipeline similar to Ref.~\cite{Christian:2017} was implemented, as depicted in Fig.~\ref{fig:HorizonPipeline}, leading to empirically sound performance. The image was first scanned in the direction of illumination to yield a coarse mask of possible edge points, then edges from the image were extracted via the Sobel edge detection algorithm. Once these edges were filtered against the coarse mask, a denser scan in the direction of illumination was yet again conducted to remove any edges not lying on the lit limb itself. The resultant pixel-level lit limb points were then refined to subpixel accuracy via a Zernike moment-based method \cite{Christian:2017,Renshaw:2020}, where the appropriate kernel width was informed by the PSF characterization in Section~\ref{Sec:PSF}. The subpixel horizon points were then used within the CRA to produce the OPNAV measurement. Although the CRA provides the full position vector from the camera to the Earth, only the direction (i.e., a LOS measurement) was used for LONEStar. This method consistently yielded an estimate within about 1.5 pixels of what was predicted by reprojection of the JPL-produced reference trajectory (see Fig.~\ref{fig:EarthCRAResiduals}).

\subsection{Celestial Triangulation with the Earth and Moon}

\begin{figure}[b!]
    \centering
    \includegraphics[width=1\linewidth]{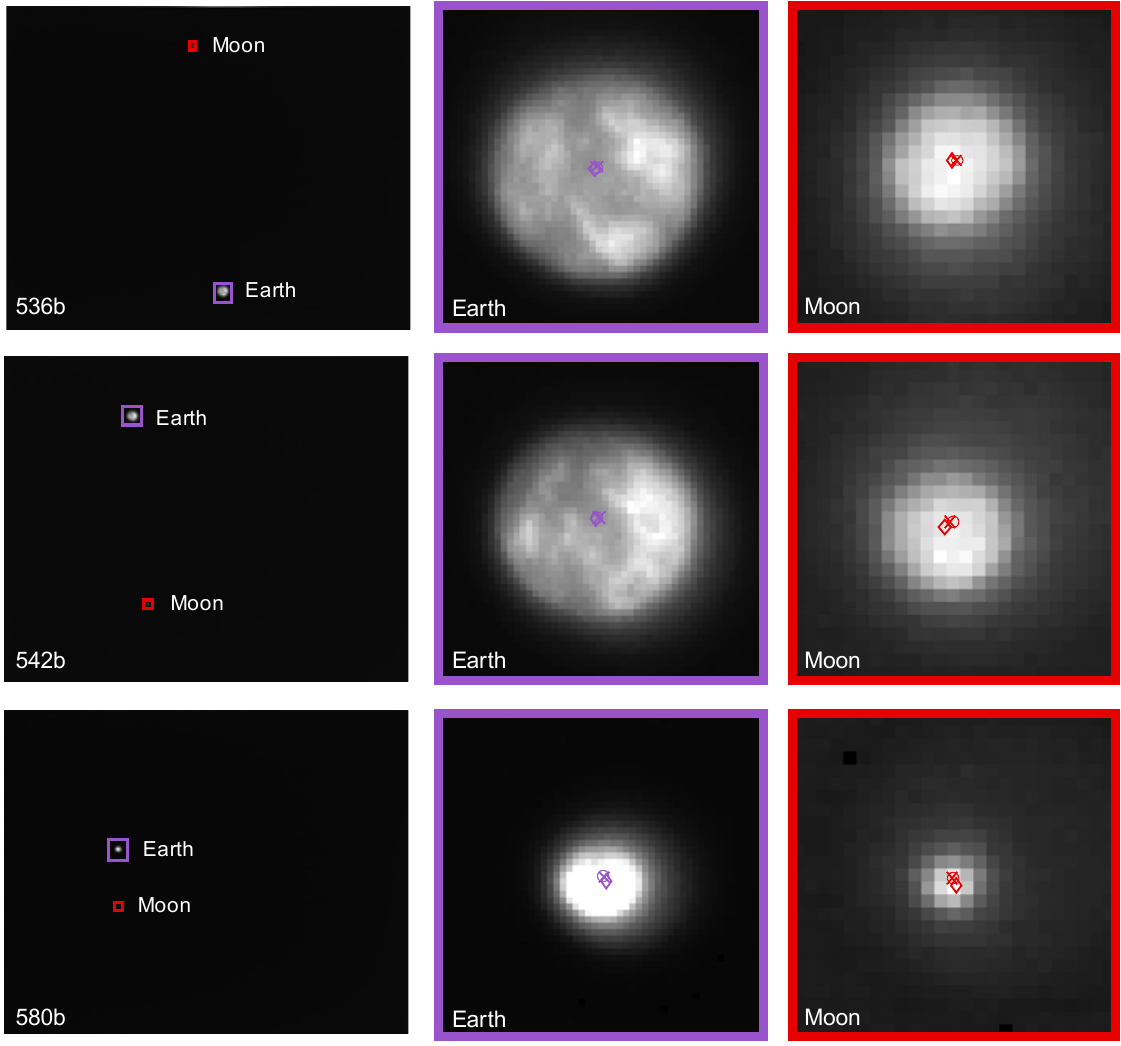}
    \caption{Selected short-exposure Earth-Moon images. Circles indicate the observed centroid, diamonds indicate a centroid projection from the JPL-produced navigation solution, and x’s indicate the reprojection from the OPNAV-produced solution.}
    \label{fig:EarthMoonShort}
\end{figure}

During the LONEStar imaging campaign, the Earth and Moon were captured simultaneously in a single image 21 times, yielding the largest observation dataset of the mission. Since both celestial bodies were seen in the same image, LOST is the appropriate triangulation method for LF localization in this case. Figure~\ref{fig:EarthMoonShort} shows short-exposure images from three such observations, illustrating the evolution of the appearance of these bodies throughout the campaign. 

In all cases, the Image Block consisted of two bracketing long-exposure images, used for attitude determination, surrounding three short-exposure images, each with small variations in exposure time and image sensor gain. Both long-exposure images were downlinked in all cases, with the exception of OPNAV Block 596, where only long-exposure image 596e was successfully downlinked. As such, the attitude for each OPNAV measurement was interpolated using SLERP from bracketing star images for 20 of the 21 observations. 

\begin{figure}[b!]
    \centering
    \includegraphics[width=1\linewidth]{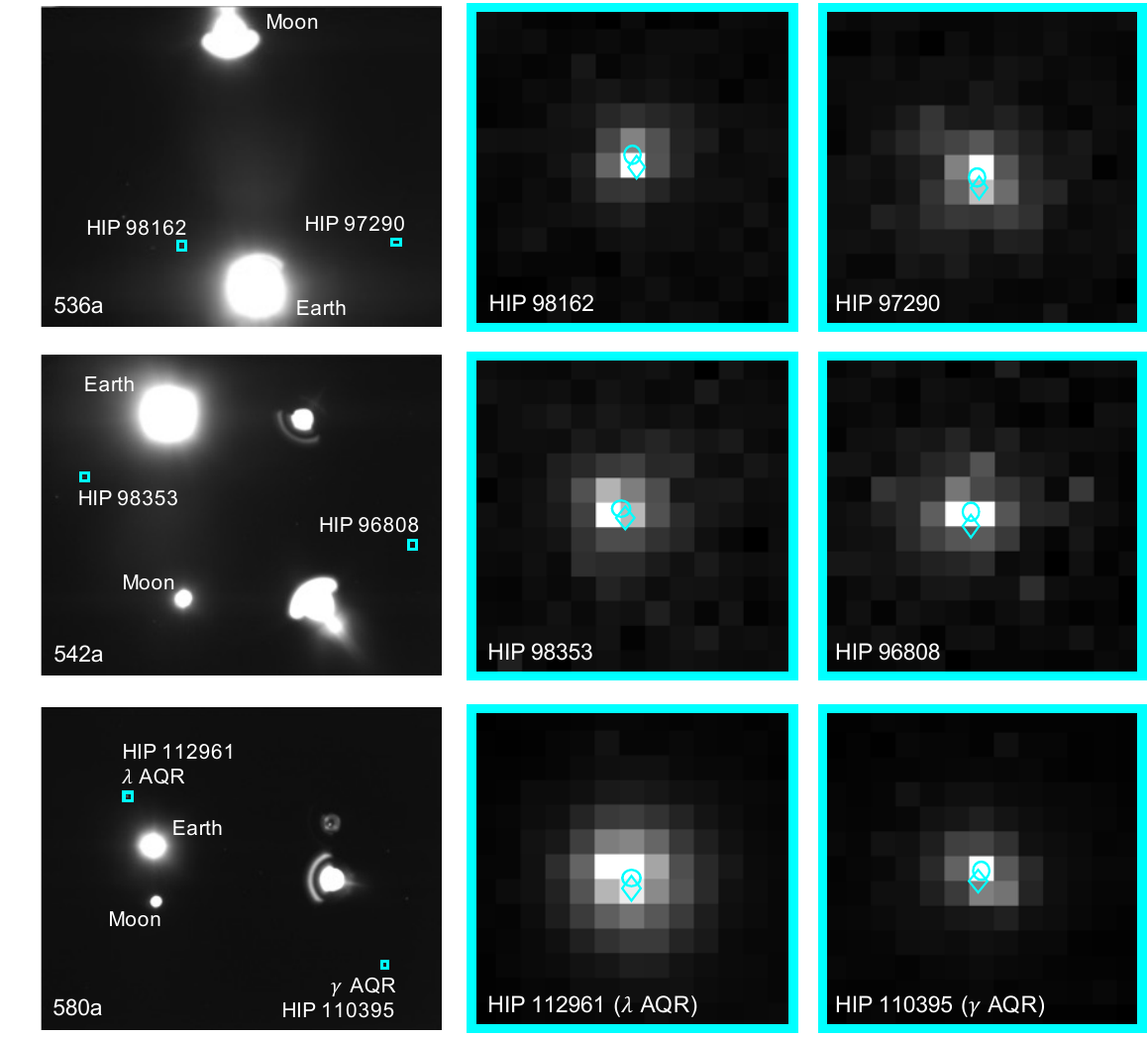}
    \caption{Selected long-exposure Earth-Moon images. Circles indicate the observed centroid, diamonds indicate a centroid projection from the JPL-produced navigation solution. Note that the brightness of each patch around these stars has been renormalized such that the darkest pixel is black and the brightest is white.}
    \label{fig:EarthMoonLong}
\end{figure}

\begin{figure}[b!]
    \centering
    \includegraphics[width=0.8\linewidth]{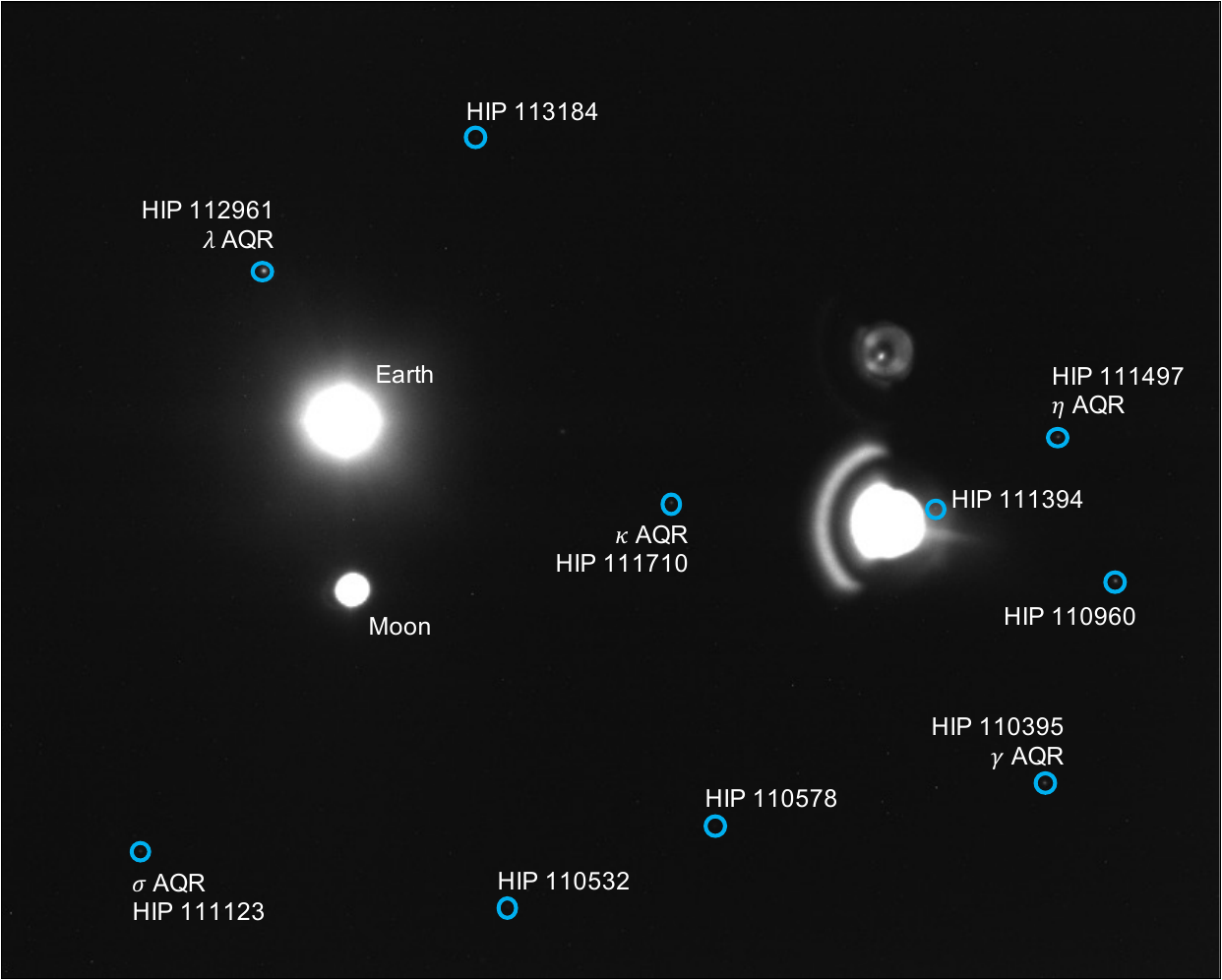}
    \caption{Image 580a, a long-exposure image showing a saturated Earth and Moon, alongside the 10 stars of magnitude 6 or brighter visible in the image. Note that while only the brightest 10 stars are labeled, 189 stars down to magnitude 9 appear in this frame.}
    \label{fig:EarthMoonStars}
\end{figure}

Attitude determination from the long-exposure images appeared uniquely challenging in this context (as compared to the distant planets), given the proximity and relative brightness of the Earth and Moon in these images. Figure~\ref{fig:EarthMoonLong} illustrates that the Earth and Moon (and a great number of pixels in their vicinity) are saturated in these instances, while optical effects within the camera result in significant artifacts elsewhere in the image as well. In most cases, however, many stars are still visible elsewhere in the image (see Fig.~\ref{fig:EarthMoonLong} and Fig.~\ref{fig:EarthMoonStars}). The robust image processing pipeline discussed in Section~\ref{Sec:IPandAttDet} was sufficient to localize these visible stars while handling all adverse optical effects, leading to successful attitude determination with no special accommodations. 

Numerical results for the example images presented in Fig.~\ref{fig:EarthMoonShort} and Fig.~\ref{fig:EarthMoonLong} are given in Tables~\ref{tab:EarthMoonCamRes} and \ref{tab:EarthMoonAngSep}. Naturally, given the relative proximity of the Earth and Moon, residuals are significantly lower than triangulation scenarios with distant planets. When normalized by range, however, these results are consistent with (i.e., same order of magnitude as) the sequential triangulation results of Jupiter and Saturn. Even as LF grew more distant from the Earth-Moon system, the apparent angle between these two bodies was still well-estimated and resulted in excellent triangulation residuals when compared to the DSN-based reference trajectory. Note that the time between bracketing star images was longer in these scenarios, with about 20 seconds between bounding images. When this is considered, the pointing drift rates observed in these images are consistent with those determined in Section~\ref{Sec:SequentialTriJupSat}. In the case of 580b, the bracketing star angle was found to be much lower, but this appears to have been a coincidence. 

\begin{table}[h!t]
    \centering
    \caption{Selected Earth-Moon LOST triangulation results.}
    \begin{tabularx}{\textwidth}{l P{1.1 cm} P{1.1 cm}P{1.1 cm} P{1.1 cm} P{1 cm} P{1.8 cm}P{1.8 cm}}
    \toprule
    \multirow{4}{*}{Image} & \multicolumn{3}{c}{Camera Frame Residual [km]}  & \multicolumn{4}{c}{Residual Norm}\\
    \cmidrule(lr){2-4} \cmidrule(lr){5-8}
     & \multirow{3}{*}{X} & \multirow{3}{*}{Y} & \multirow{3}{*}{Z} & \multirow{3}{*}{$10^5$ km} & \multirow{3}{1 cm}{\centering Earth Radii} & Normalized by range to Mercury & \multirow{3}{1.8 cm}{\centering Normalized by range to Mars} \\
    \midrule
    536b & -126.66 & 18.56  & -313.84  & 0.0034 & 0.053 & 0.0001570 & 0.0001802 \\
    542b & -121.51 & 96.56  & -740.60   & 0.0076 & 0.119 & 0.0003290 & 0.0003860 \\
    580b & 260.97  & 637.11 & 1,420.61 & 0.0158 & 0.248 & 0.0002757 & 0.0002942 \\
    \bottomrule
    \end{tabularx}
    \label{tab:EarthMoonCamRes}
\end{table}

\begin{table}[h!t]
    \centering
    \caption{Comparison of apparent vs. DSN computed angular separation of Earth and Moon.}
    \begin{tabularx}{\textwidth}{lP{1.5 cm}P{2 cm}P{2 cm}P{2 cm}Y}
    \toprule
    \multirow{4}{*}{Image} & \multirow{4}{1.5 cm}{\centering LONEStar Elapsed Time [days]}  & \multicolumn{3}{c}{\multirow{2}{*}{Apparent Angle Between Jupiter and Saturn}} & \multirow{4}{2.5 cm}{\centering Bracketing Star Angle [arcsec]}\\ 
    \\
    \cmidrule(lr){3-5}
     &  & DSN Projected [deg] & \multirow{2}{1.5 cm}{\centering Measured [deg]} &  \multirow{2}{1.5 cm}{\centering Residual [arcsec]} & \\
    \midrule
    536b & 19.586 & 7.94325 & 7.94188 & 4.932  & 71.828 \\
    542b & 25.919 & 6.03806 & 6.03518 & 10.368 & 68.768 \\
    580b & 83.461 & 1.81651 & 1.81646 & 0.180  & 14.006 \\
    \bottomrule
    \end{tabularx}
    \label{tab:EarthMoonAngSep}
\end{table}

\begin{figure}[h!]
    \centering
    \includegraphics[width=1\linewidth]{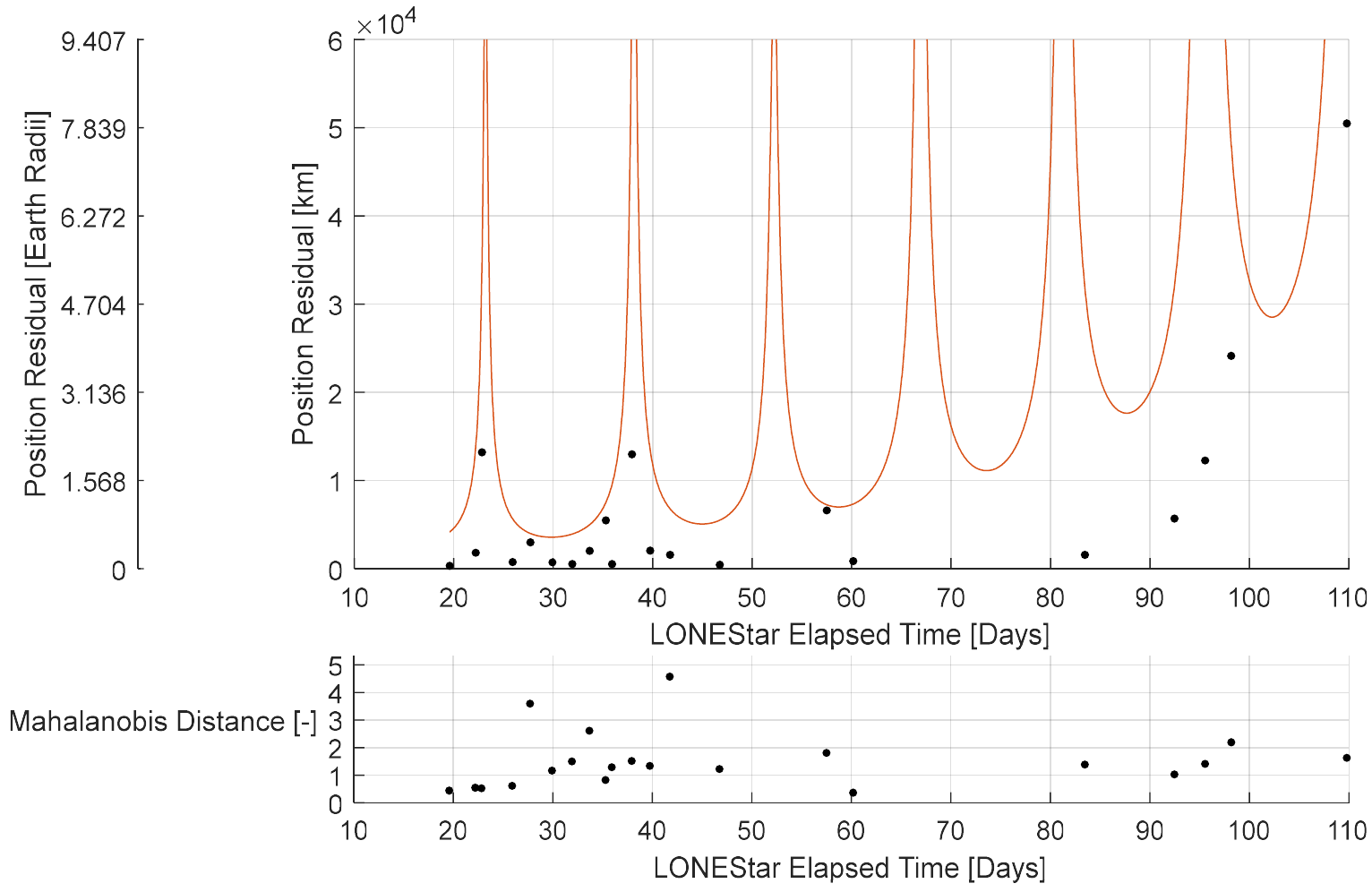}
    \caption{Earth-Moon triangulation residuals (black dots) and total error contour (orange line) as a function of time (top). These data may be used to compute the Mahalanobis distance (bottom). }
    \label{fig:EarthMoonResiduals}
\end{figure}

The particular geometry of the Earth-Moon system, as it relates to LF, leads to useful insights regarding the covariance of these measurements. Specifically, the covariance is a function of both the apparent diameter of the bodies in question (and thus their range), as well as the apparent angle between them. While the range and apparent angle of other targets was roughly constant throughout the LONEStar campaign, the orbital motion of the Moon about Earth led to a periodic apparent angle between the two. Coupled with the motion of LF moving away from the Earth-Moon system, the covariance changed considerably with time. Fig.~\ref{fig:EarthMoonResiduals} shows the norm of the triangulation residuals for each Earth-Moon OPNAV Pass, along with a continuous estimate of the instantaneous total error $\sqrt{Tr(\bP)}$ from Eq.~\eqref{eq:LOSTcovariance}. The total error is plotted as a smooth function (not just at the measurement times), to show how the changing geometry affects the instantaneous triangulation performance. The covariance computations assume 0.5 pixels of pointing error and 0.25-1.0 pixel of centroiding error for the Moon and Earth (depending on apparent diameter). The spikes in this contour correspond to the Moon passing close to the Earth, where the apparent angle between them becomes small and the triangulation geometry becomes poor. As expected, the OPNAV passes with higher residuals are strongly correlated with regions of larger OPNAV covariance. The associated Mahalanobis distance for each case is also presented below these results in Fig.~\ref{fig:EarthMoonResiduals}, with most measurements having a Mahalanobis distance below two. 

\section{OPNAV-Only Orbit Determination}
\label{Sec:BatchFilter}
The results of Section~\ref{Sec:CelTriangPlanetsTop} and Section~\ref{Sec:OPNAVEarthMoonTop} focused on instantaneous localization of LF using triangulation, CRA, and related techniques. While such instantaneous solutions are valuable, it is equally important to process the planet LOS measurements directly in a navigation filter to produce an orbit determination (OD) solution. Using a batch filter~\cite{Tapley:2004}, an OPNAV-only OD solution was produced for two scenarios. The first scenario used only the distant planet observations (Mercury, Mars, Jupiter, Saturn), which provides representative OPNAV-only navigation performance within the inner Solar System while far away from all the observed bodies (e.g., for an exploration mission during interplanetary cruise). The second scenario uses observations of the Earth and Moon in addition to all the distant planet observations (Mercury, Mars, Jupiter, Saturn), which represents the best OPNAV-only OD solution for LF that is possible with the LONEStar data set.

\subsection{Batch Filter Design}

The LONEStar batch OD solution is produced by solving a weighted nonlinear least-squares problem. This is accomplished while assuming no \textit{a priori} information from the DSN-based OD solution. A maximum likelihood estimate is constructed by attempting to find the LF state at a single reference time that best explains the entire sequence of OPNAV measurements over the 97-day imaging campaign---including 17 LOS measurements for Scenario 1 and 59 LOS measurements for Scenario 2. This is accomplished by minimizing the negative log likelihood function of the observations
\begin{equation}
    \label{eq:CostFcnBatchOD}
    \text{min } J(\bx_0) = \sum_{i = 1}^n \left( \by_i - h(\bx_i) \right)^T \bR_i^{-1} \left( \by_i - h(\bx_i) \right)
\end{equation}
which is essentially a sum of the measurement residual Mahalanobis distances. In Eq.~\eqref{eq:CostFcnBatchOD} the state $\bx_i$ is computed by propagating the reference state $\bx_0$ from the reference time $t_0$ to the measurement time $t_i$. The measurement model $h(\bx)$ is the full camera model from Section~\ref{Sec:CamModel}, which includes projective geometry, lens distortions, conversion to pixel coordinates, stellar aberration, and light time-of-flight. Attitude was not estimated by the filter, but was instead estimated from the star field background using the procedures discussed in Section~\ref{Sec:IPandAttDet}.

The dynamical model used by the batch filter to propagate the state considered gravitational attraction from major celestial bodies within the Solar System (the Sun, planets, and Earth’s moon) and solar radiation pressure (SRP). Given the influence of SRP, the SRP ballistic coefficient $\beta = C_R A / m$  was estimated as part of the orbit determination process. SRP acceleration was modeled according to \cite{Montenbruck:2000}
\begin{equation}
    \ba_{SR} = P \beta \frac{\br - \br_S}{\|\br - \br_S \|^3} AU^2
\end{equation}
where $P = 4.56 \times 10^{-6} \ Nm^{-2}$ and $\br_S$ is the position of the sun with respect to the SSB. The LF dynamics were modeled in the ICRF with an origin at the SSB. This convention has the advantage of simplifying the gravitational acceleration of the $N$-body problem to a summation of the contributing accelerations from each body. Thus, the total acceleration is given by 
\begin{equation}
    \ba = \ba_{SR} + \sum_{i = 1}^n - \frac{GM_i}{\| \br - \br_{i} \|^3} \left( \br - \br_{i} \right)
\end{equation}
where the positions of the solar system bodies were obtained from DE440 ephemerides~\cite{Park:2021}.

The measurement error covariance was empirically determined by inspection of the triangulation residuals. Measurement errors were assumed to be zero-mean, Gaussian, and isotropic with standard deviations of 0.25 pixel (Saturn), 0.5 pixel (Jupiter), 0.75 pixel (Mercury), 0.75 pixel (Mars), 0.25-1.0 pixel (Earth and Moon, depending on apparent radius).

\subsection{Scenario 1: OPNAV-Only OD with Distant Planets}

\begin{figure}[b!]
    \centering
    \includegraphics[width=1\linewidth]{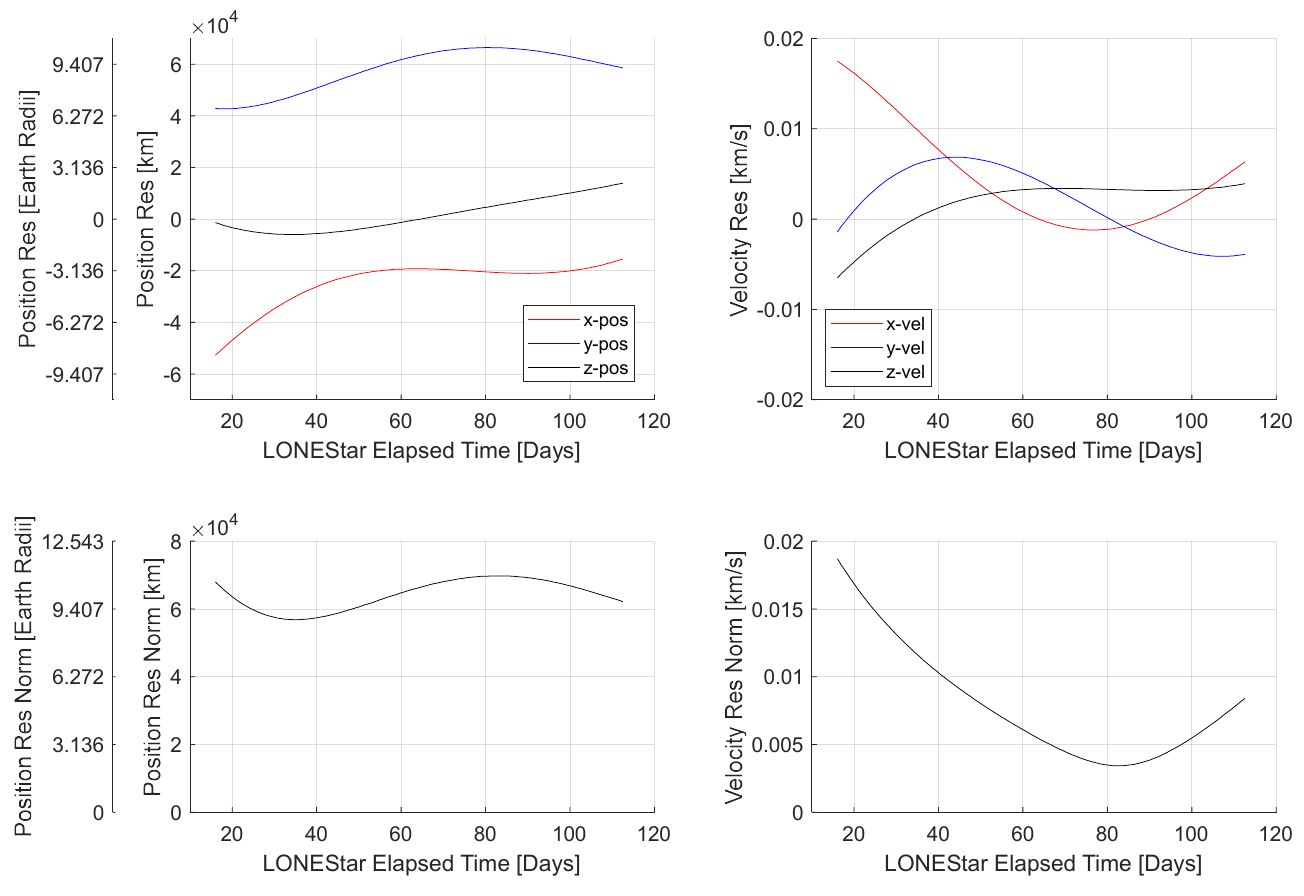}
    \caption{Orbit determination was completed with only OPNAV measurements to distant planets. The state residuals compared to the DSN solution are shown over the course of the imaging campaign.}
    \label{fig:DistPlanetsOD}
\end{figure}

The OPNAV-only distant planet OD solution included 1 Mercury observation, 1 Mars observation, 8 Jupiter observations, and 7 Saturn observations. The converged LMA solution from dynamic triangulation (discussed in Section~\ref{Sec:DynamicTri}) was used as the initial guess for the batch estimation. The resulting OPNAV-only OD result was compared with the JPL-provided OD solution based on DSN observables, with residuals as shown in Fig.~\ref{fig:DistPlanetsOD}. Here, a difference of about 6-11 Earth radii between the two OD solutions is observed over the 97-day period. This performance is achieved despite the comparatively large ranges to the distant planets at this time, which were 0.9-1.4 AU (Mercury), 2.4-2.5 AU (Mars), 4.0-4.9 AU (Jupiter), and 8.8-9.2 AU (Saturn). The OPNAV OD residuals are an order of magnitude larger than the radiometric OD residuals, therefore the reference trajectory from Earth-based tracking seems to be a reasonable surrogate for the true orbit in this scenario.

\begin{figure}[b!]
    \centering
    \includegraphics[width=0.8\linewidth]{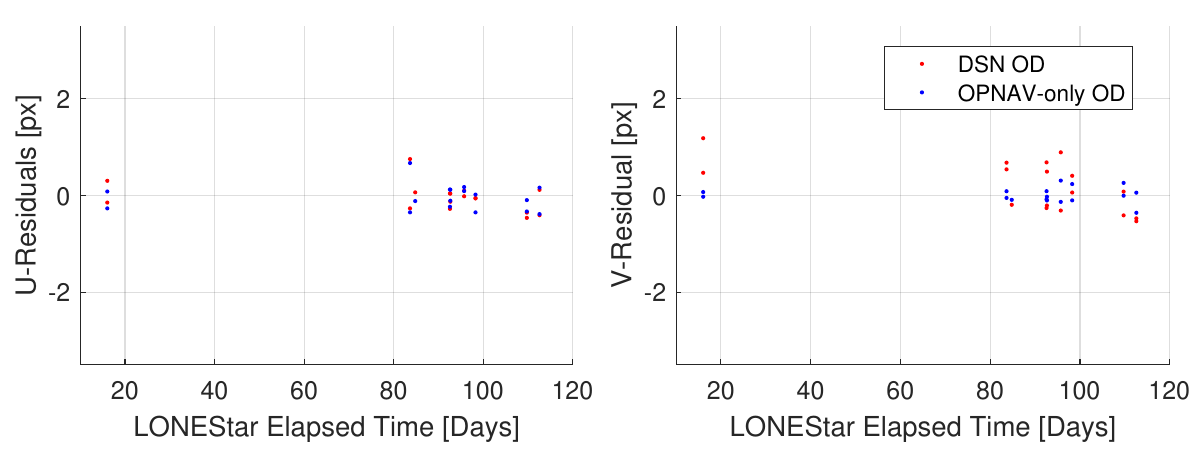}
    \caption{Distant planet OPNAV measurement residuals from the batch estimation compared to DSN residuals.}
    \label{fig:DistPlanetsResidual}
\end{figure}

Finally, as one would expect, the OPNAV-only OD solution demonstrated substantially better agreement with the OPNAV observables. This may be seen by the smaller measurement residuals for the OPNAV-only OD solution (blue dots) than for the DSN-based OD solution (red dots) in Fig.~\ref{fig:DistPlanetsResidual}.

\subsection{Scenario 2: OPNAV-Only OD with Earth, Moon, and Distant Planets}

\begin{figure}[b!]
    \centering
    \includegraphics[width=1\linewidth]{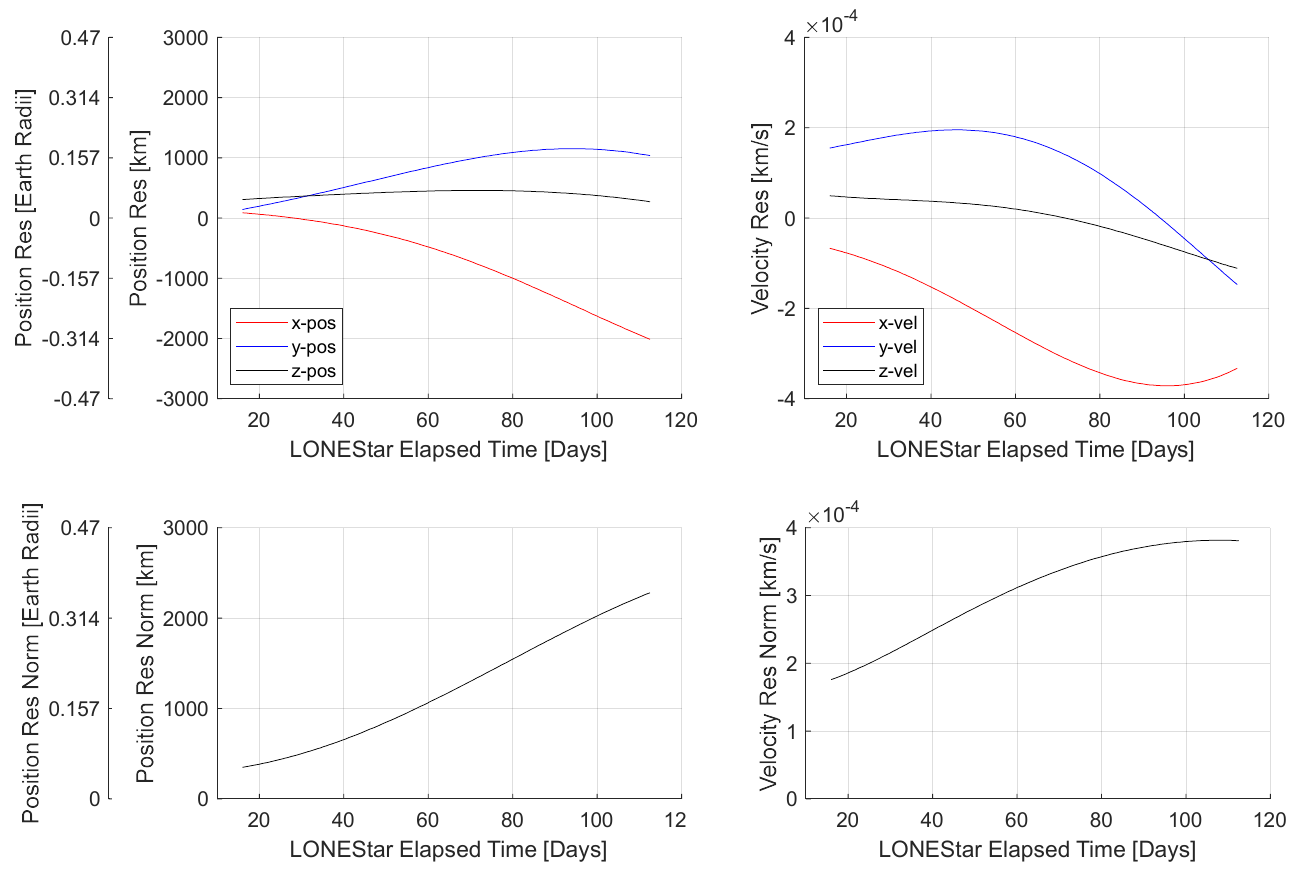}
    \caption{Orbit determination was completed with all OPNAV measurements, including measurements to distant planets, the Earth, and the Moon. The state residuals compared to the DSN solution are shown over the course of the imaging campaign.}
    \label{fig:AllObsOD}
\end{figure}

\begin{figure}[b!]
    \centering
    \includegraphics[width=0.8\linewidth]{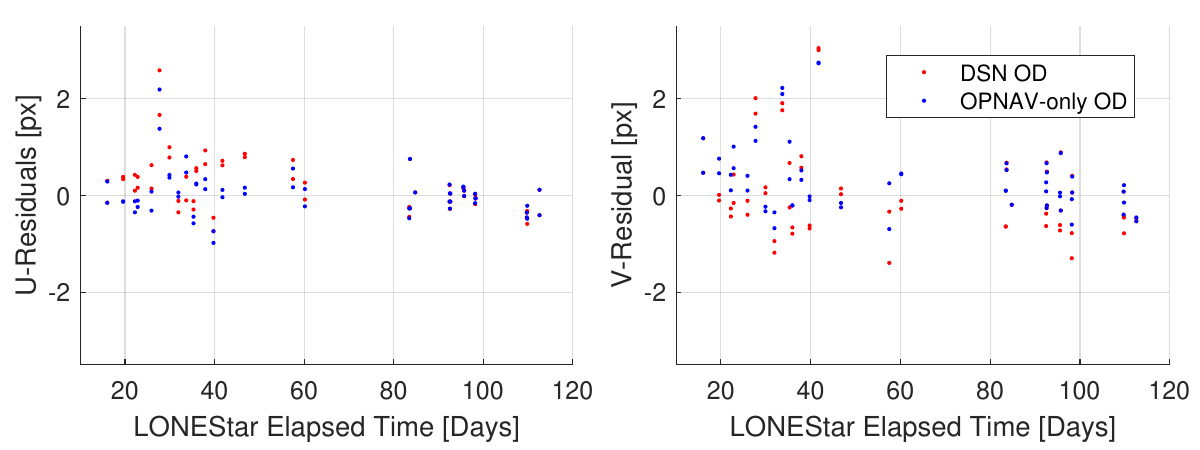}
    \caption{All OPNAV measurement residuals from the batch estimation compared to DSN residuals.}
    \label{fig:AllObsResidual}
\end{figure}

The full OPNAV-only OD solution for LF included all the measurements from Scenario 1, along with an additional 21 Earth observations and 21 Moon observations. In this case, the initial guess for the batch estimation was taken to be the OD solution from scenario 1. As before, this OD result was compared with the JPL-provided OD solution based on DSN observables, with residuals as shown in Fig.~\ref{fig:AllObsOD}. With the inclusion of Earth and Moon observations, which were much closer to LF than the distant planets, the LF position residual drops to about 0.075-0.45 Earth radii over the 97-day period. These residuals are on the order of the uncertainty in the DSN-based OD solution, which makes it difficult to say which one is better. In this case, the measurement residuals for the OPNAV-only OD solution (blue dots) and the DSN-based OD solution (red dots) are similar in Fig.~\ref{fig:AllObsResidual}, with the OPNAV residuals having a slightly smaller standard deviation.

\clearpage
\section{Conclusion}

Lunar Flashlight (LF) was a JPL-led mission launched in December 2022 that left the Earth-Moon system after an Earth flyby on 2023-MAY-17. Subsequently, the LF Optical Navigation Experiment with a Star tracker (LONEStar) investigation was performed as part of a Georgia Tech executed extended mission. This manuscript summarizes the results of LONEStar.

LONEStar successfully demonstrated the heliocentric localization of a spacecraft by triangulation with observations of distant planets in three different experiments. First, instantaneous triangulation was demonstrated using simultaneous observations of two planets (Mercury and Mars) in a single image. Second, when two planets were not close enough to be seen in one image, LONEStar demonstrated triangulation with a pair of sequential OPNAV images taken in close succession. This was accomplished with images of Jupiter and Saturn. Third, LONEStar demonstrated localization by dynamic triangulation for situations when sequential images are separated by long periods of time (not in close succession), which was again shown with observations of Jupiter and Saturn. LF localization by triangulation to distant planets was found to provide instantaneous position estimates with errors on the order of about 10-30 Earth radii (neglecting the Mercury-Mars cases with especially poor lighting).

LONEStar also successfully demonstrated OPNAV with images of the Earth and Moon. Line-of-sight (LOS) observables of the Moon were found through normalized cross-correlation with a template, while LOS observations of the Earth were found through horizon-based techniques. As with the distant planets, triangulation was used to provide instantaneous position estimates for LF. However, since the Earth and Moon are much closer than the distant planets, the instantaneous localization errors were usually less than about 1-2 Earth radii.

Finally, the instantaneous position estimates may be considerably improved by processing OPNAV observations directly within a batch orbit determination pipeline. The intent of this LONEStar study was to investigate OD using only OPNAV measurements (i.e., completely independent of DSN). Using OPNAV-only initial guesses of the trajectory (e.g., via dynamic triangulation), the batch OD solution showed residuals of about 9-11 Earth radii when including OPNAV measurements from the distant planets (Mercury, Mars, Jupiter, Saturn). When Earth and Moon OPNAV measurements were added, the residuals dropped to less than 0.35 Earth radii. This suggests that navigation results suitable for many mission profiles may indeed be achieved in heliocentric space using only OPNAV measurements.

\backmatter

\bmhead{Acknowledgements}

The authors thank the entire Lunar Flashlight (LF) team for making this mission possible, especially the student-led LF operations team at Georgia Tech. The authors also thank Stuart Demcak for assistance with the JPL-produced reference trajectory, Shyam Bhaskaran for helpful feedback on this manuscript, and S\'{e}bastien Henry for insightful discussions on triangulation techniques. This work made use of the Caltech/NASA Jet Propulsion Laboratory TBALL software.
 
The LONEStar investigation was made possible through internal funding from Georgia Institute of Technology and the Georgia Tech Research Institute. Contributing algorithms were partially developed under NASA awards 80NSSC22M0151 and 80NSSC23K1229. The authors also acknowledge Caltech/NASA Jet Propulsion Laboratory (JPL) award 80NM0018D004 for integration, testing, and operation of the LF spacecraft at Georgia Tech.




\end{document}